%% file: arxiv_google_main.tex
\pdfoutput=1
\documentclass[11pt, a4paper, logo, copyright]{arxiv_google_googlecloud}

\pdftrailerid{redacted}

\makeatletter
\renewcommand\bibentry[1]{\nocitep{#1}{\frenchspacing\@nameuse{BR@r@#1\@extra@b@citeb}}}
\makeatother

\usepackage[authoryear, sort&compress, round]{natbib}

\usepackage[utf8]{inputenc}
\usepackage[T1]{fontenc}
\usepackage{hyperref}
\usepackage{url}
\usepackage{booktabs}
\usepackage{amsfonts}
\usepackage{nicefrac}
\usepackage{microtype}
\usepackage{xcolor}
\usepackage{graphicx}
\usepackage{arxiv_google_algorithm}
\usepackage{arxiv_google_algorithmic}
\usepackage{amsmath}
\usepackage{multicol}
\usepackage{multirow}
\usepackage{enumitem}
\usepackage{float}
\usepackage{tcolorbox}
\tcbuselibrary{breakable}
\usepackage{wrapfig}
\usepackage{titletoc}
\usepackage{placeins}   %
\usepackage{xspace}
\usepackage{pifont}     %
\usepackage{cleveref} %

\crefname{appendix}{Appendix}{Appendices}
\Crefname{appendix}{Appendix}{Appendices}

\input{math_commands.tex}

\graphicspath{{./}}

\makeatletter
\renewcommand\paragraph{\@startsection{paragraph}{4}{\z@}{1.2ex \@plus .5ex \@minus .2ex}{-1em}{\normalfont\normalsize\bfseries}}
\makeatother
\newcommand{\fwscale}{1.1}   %

\newcommand{\bench}{\textsc{WhatIf-Bench}\xspace}
\providecommand{\cmark}{\textcolor{green!60!black}{\ding{51}}}
\providecommand{\xmark}{\textcolor{red!70!black}{\ding{55}}}
\providecommand{\Xb}{\ensuremath{\mathbf{X}}}
\providecommand{\Zb}{\ensuremath{\mathbf{Z}}}
\providecommand{\zb}{\ensuremath{\mathbf{z}}}

\definecolor{added}{rgb}{0,0.6,0}
\definecolor{deleted}{rgb}{1,0,0}

\title{\bench: Benchmarking Intervention-Aware Time Series Forecasting}

\correspondingauthor{Arjun Ashok \href{mailto:arjun.ashok.psg@gmail.com}{<arjun.ashok.psg@gmail.com>}, Mihir Parmar \href{mailto:mihirparmar@google.com}{<mihirparmar@google.com>} and Palash Goyal \href{mailto:palashgoyal@google.com}{<palashgoyal@google.com>}}

\renewcommand{\today}{}

\author[1, 2, 3]{\fontsize{10.0pt}{10.0pt}\selectfont Arjun Ashok}
\author[1]{\fontsize{10.0pt}{10.0pt}\selectfont Mihir Parmar}
\author[1]{\fontsize{10.0pt}{10.0pt}\selectfont Palash Goyal}
\author[1]{\fontsize{10.0pt}{10.0pt}\selectfont Bhavana Dalvi Mishra}
\author[1]{\fontsize{10.0pt}{10.0pt}\selectfont Chun-Liang Li}
\author[2, 3]{\fontsize{10.0pt}{10.0pt}\selectfont Irina Rish}
\author[2, 4]{\fontsize{10.0pt}{10.0pt}\selectfont Alexandre Drouin}
\author[1]{\fontsize{10.0pt}{10.0pt}\selectfont Jinsung Yoon}
\author[1]{\fontsize{10.0pt}{10.0pt}\selectfont Tomas Pfister}
\affil[1]{\fontsize{9.0pt}{9.0pt}\selectfont Google}
\affil[2]{\fontsize{9.0pt}{9.0pt}\selectfont Mila - Quebec AI Institute}
\affil[3]{\fontsize{9.0pt}{9.0pt}\selectfont Université de Montréal}
\affil[4]{\fontsize{9.0pt}{9.0pt}\selectfont Université Laval}

\begin{abstract}
\input{inputs/abstract.tex}

\par\vspace{4pt}\noindent\hfill
\href{https://huggingface.co/spaces/arjunashok/WhatIf-Bench}{\raisebox{-0.25em}{\includegraphics[height=1.25em]{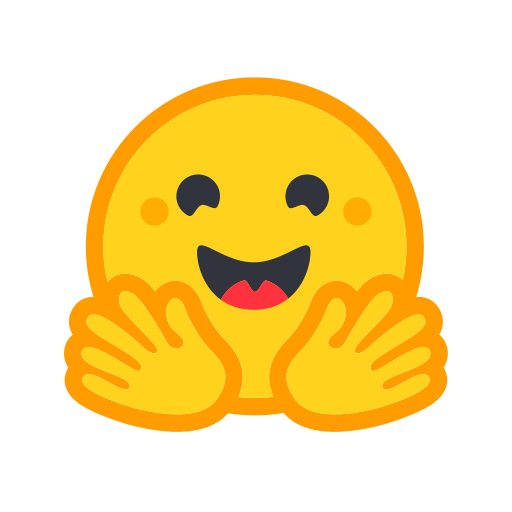}}\,\textbf{Leaderboard}}\qquad
\href{https://github.com/ashok-arjun/whatif-bench}{\raisebox{-0.2em}{\includegraphics[height=1.05em]{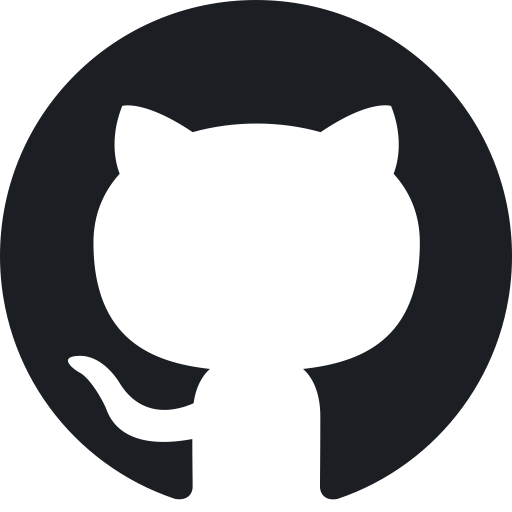}}\,\,\textbf{Code}}%
\hfill\null\vspace{-16pt}
\end{abstract}

\begin{document}

\maketitle

\input{inputs/intro_v2.tex}

\input{inputs/new_related_work.tex}

\input{inputs/benchmark_v2.tex}

\input{inputs/results_v2.tex}

\input{inputs/discussion_and_future_work.tex}

\bibliographystyle{abbrvnat}
\bibliography{whatifbench}

\appendix
\newpage

\crefalias{section}{appendix}
\crefalias{subsection}{appendix}

\titlecontents{section}[1.5em]
  {\vspace{0.5em}\bfseries\color{blue}}
  {\contentslabel{1.5em}}
  {\hspace*{-1.5em}}
  {\titlerule*[0.5pc]{.}\contentspage}
\titlecontents{subsection}[3.8em]
  {\normalfont\color{blue}}
  {\contentslabel{2.3em}}
  {\hspace*{-2.3em}}
  {\titlerule*[0.5pc]{.}\contentspage}
\vspace{2em}
\noindent{\Large\textbf{\textsf{Table of Contents for Appendix}}}\par
\vspace{1em}
\startcontents[appendix]
\printcontents[appendix]{}{1}{\setcounter{tocdepth}{2}}
\vspace{2em}
\clearpage

\input{inputs/appendix_benchmark.tex}

\input{inputs/appendix_v2}

\clearpage
\input{inputs/appendix_task_examples_v2}

\end{document}

%% file: math_commands.tex
\usepackage{amsmath,amsfonts,bm}

\def\eqref#1{equation~\ref{#1}}

\def\1{\bm{1}}

\DeclareMathAlphabet{\mathsfit}{\encodingdefault}{\sfdefault}{m}{sl}
\SetMathAlphabet{\mathsfit}{bold}{\encodingdefault}{\sfdefault}{bx}{n}

%% file: inputs/abstract.tex
\providecommand{\bench}{\textsc{WhatIfBench}}

Recent time series foundation models (TSFMs) can consume covariates in-context, extending zero-shot forecasting to problems driven by external factors.
This enables scenario analysis: forecasting how a target responds to \emph{events} that set covariates to chosen future values.
Whether they can do this reliably is an open question.
We introduce \bench{}, a benchmark measuring how well forecasters use covariates to forecast a target's response to future events.
\bench{} is \emph{semi-synthetic}: time series come from mechanistic simulators of data centers, power grids, road traffic, and water networks, with controlled covariate-target relationships grounded in each system.
Each task presents a history of events intervening on covariates, then asks to forecast a future event that either replays seen values (\emph{Replay}) or sets them to unseen ones (\emph{Novel}).
Varying the number of covariates, historical events, and event length yields 37,665 tasks.
Evaluating TSFMs, frontier LLMs, and simple regression baselines, we find the LLMs reach the highest skill (improvement over a naive baseline), ahead of the best TSFMs.
Every model forecasts well on Replay futures ($48$ to $85\%$ skill) but drops to $-7$ to $50\%$ on Novel futures, with the regression baselines falling the most.
Skill on Novel futures also falls as events set more covariates jointly.
More historical events raise skill on Novel futures across the board, but large differences remain.
\bench{} reveals generalization to Novel futures as the core challenge, and offers a testbed toward arbitrary scenario analysis.

%% file: inputs/intro_v2.tex
\section{Introduction}
\label{sec:intro}

\providecommand{\fwscale}{1}
\providecommand{\fw}[1]{\raisebox{-0.06in}[0.13in][0.05in]{\scalebox{\fwscale}{\includegraphics{figures/font_words/word_#1.pdf}}}}
\begin{wrapfigure}{r}{0.55\textwidth}
  \vspace{-\intextsep}
  \vspace{-2\baselineskip} %
  \centering
  \includegraphics[width=0.55\textwidth]{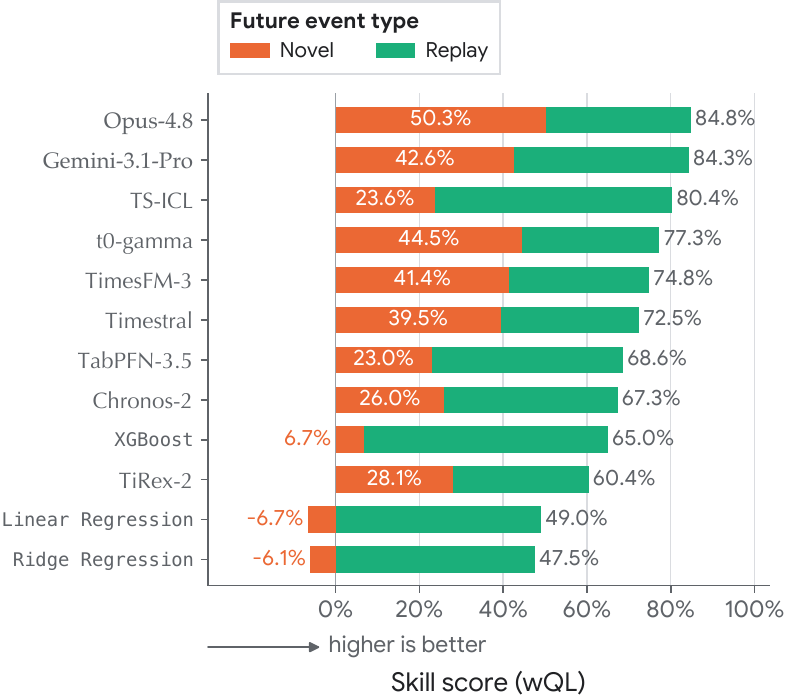}
  \caption{\small Performance of \fw{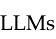}, \fw{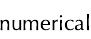} \fw{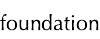} \fw{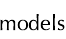}, and \fw{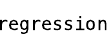} \fw{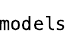} on \bench{} tasks.
  Every model's skill score drops sharply from Replay to Novel futures.}
  \label{fig:replay-vs-novel}
  \vspace{-3\intextsep}
\end{wrapfigure}
Forecasting guides decisions across science, engineering, and commerce~\citep{hyndman2018forecasting}, and models have advanced from classical methods and deep networks~\citep{box2015time,expsmoothing,benidis2022deeplearning,salinas2020deepar,nie2023patchtst} to time series foundation models (TSFMs) that forecast unseen series zero-shot~\citep{ansari2024chronos,das2023decoder,woo2024unified,liang2024foundation}.
Most TSFMs are univariate, yet many real targets are shaped by external drivers such as weather, prices, and policies; recent TSFMs can instead consume such covariates \emph{in-context} and factor them into the forecast without task-specific training~\citep{chronos2,timesfm3,tsicl,tirex2,t0alpha}.

When the future covariates are known, this enables \emph{scenario analysis}: predicting how the target would evolve under different futures, which is central to planning~\citep{doan1984forecasting,waggoner1999conditional,banbura2015conditional}.
Existing benchmarks, however, evaluate only on the covariate values their datasets happen to contain~\citep{shchur2025fevbench,lago2021epftoolbox}, so they cannot tell whether a model forecasts accurately when covariates take values beyond that range. They also measure performance only in aggregate over the conditions each dataset provides, so they cannot reveal where a model systematically fails at using covariates.

\begin{figure*}[t]
  \centering
  \includegraphics[width=\linewidth]{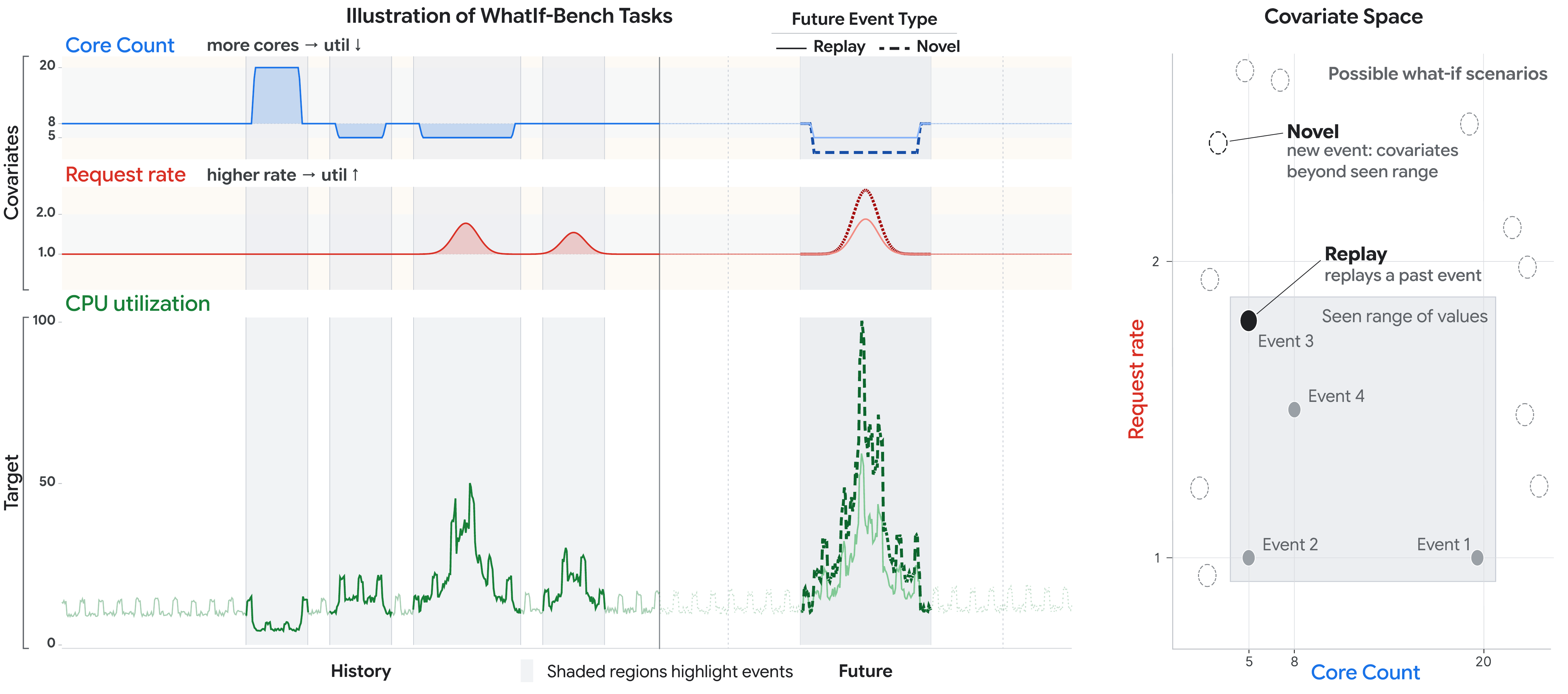}
  \caption{\textbf{Illustration of \bench{} tasks.} Left: a \bench{} task contains a history of events, in which covariates are set and the target responds, and asks a model to forecast the target under one further event. Right: the same events in covariate space. A Replay future repeats a past event (here, the third), so it lies inside the range of values the history has shown; a Novel future sets the covariates beyond that range, and is one of many possible what-if scenarios. Novel futures thus test whether a model extends what it learned to unseen values. \Cref{fig:replay-vs-novel} shows that current models do this poorly: their skill on Novel futures is far below their skill on Replay futures.}
  \label{fig:task}
  \vspace{-1.5em}
\end{figure*}

We present \bench{}, a \emph{semi-synthetic} benchmark that measures how well a forecaster uses in-context covariates to forecast a target's response to future events, including scenarios that set covariates to values never seen in the history.
Its target series come from mechanistic simulators of four domains - OpenDC (data centers)~\citep{iosup2017opendc}, GridLAB-D (power grids)~\citep{chassin2008gridlabd}, SUMO (road traffic)~\citep{lopez2018sumo}, and EPANET (water networks)~\citep{rossman2000epanet}, on which we impose controlled covariate-target relationships grounded in each system.
Because these relationships are known exactly, we can isolate the effect of any single covariate, so a model's error reflects how well it handles the covariate rather than irreducible noise.
Each task presents a history of \emph{events}, each setting covariates to chosen values and letting the target respond, and asks the model to forecast the target under one further event: a \emph{Replay} future that repeats a past event, or a \emph{Novel} future that sets covariates to unseen values, as illustrated in \Cref{fig:task}.
\bench{} is a controllable framework: we vary one axis at a time -- the future type, the number of covariates set jointly, the number and length of historical events, and the context and prediction length -- yielding $37{,}665$ tasks over $100$ target series, so each axis can be studied on its own.

We evaluate covariate-aware foundation models and frontier LLMs. The LLMs consume the series and its covariates as text, and have proven competitive at forecasting~\citep{gruver2023llmtime,requeima2024llmprocesses}. We also evaluate the simplest versions of regression baselines (XGBoost, Ridge Regression, and Linear Regression), which map the covariates at each step to the target.
We use the skill score as our main metric: a model's relative improvement over a naive baseline that repeats the last value.
First, on the full benchmark, the frontier LLMs lead: {Claude-Opus-4.8} and {Gemini-3.1-Pro} reach the top skill scores, $72.5\%$ and $70\%$, with the best covariate-aware TSFMs, {t0-gamma}, {TimesFM-3} and {TS-ICL}, close behind at $61$ to $65\%$, and a simple, untuned XGBoost baseline ($43\%$) below every foundation model.
Next, every model shares a failure mode: all forecast well on \emph{Replay} futures ($48$--$85\%$ skill) but drop to between $-7\%$ and $50\%$ on \emph{Novel} ones (\Cref{fig:replay-vs-novel}). The regression baselines collapse the most: Linear Regression has a skill of $49\%$ on Replay futures but $-7\%$ on Novel futures. Therefore, despite their own drop, the LLMs and foundation models keep a large advantage over the regression baselines on Novel futures.
Further, the number of covariates has no clear effect on Replay futures, but skill on Novel futures also falls as events set more covariates jointly.
Finally, a longer history of events helps: models improve their skill, especially on Novel futures, though large differences between models remain.
Together, these results show that generalizing to Novel futures is the core challenge, and the further axes of \bench{} show how it depends on the number of covariates and on the history available.
\bench{} provides a testbed toward models that support arbitrary scenario analysis.

%% file: inputs/new_related_work.tex
\begin{table*}[t]
\centering
\small
\resizebox{\textwidth}{!}{%
\begin{tabular}{llccccc}
\toprule
\textbf{Benchmark} & \textbf{Data source} & \textbf{\shortstack{Includes\\covariates}} & \textbf{\shortstack{Covariate effect\\known exactly}} & \textbf{\shortstack{Evaluates arbitrary\\what-if scenarios}} & \textbf{\shortstack{Systematic difficulty\\(\# covariates,\\historical evidence)}} & \textbf{\shortstack{Tasks with\\covariates}} \\
\midrule
Monash \citep{godahewa2021monash}       & Real-world      & \xmark & \xmark & \xmark & \xmark & 0 \\
GIFT-Eval \citep{aksu2024gifteval}     & Real-world      & \xmark & \xmark & \xmark & \xmark & 0 \\
TSFM-Bench \citep{li2024tsfmbench}     & Real-world      & \xmark & \xmark & \xmark & \xmark & 0 \\
TIME \citep{qiao2026time}              & Real-world      & \xmark & \xmark & \xmark & \xmark & 0 \\
{fev-bench} \citep{shchur2025fevbench} & Real-world & \cmark & \xmark & \xmark & \xmark & 46 \\
\cite{berthelier2026covrelationships} & Semi-synthetic (toy relations) & \cmark & \cmark & \xmark & \xmark & --$^\dagger$ \\
\textbf{\bench{} (Ours)}               & \textbf{Semi-synthetic (simulator-grounded)} & \cmark & \cmark & \cmark & \cmark & \textbf{37,665} \\
\bottomrule
\multicolumn{7}{l}{$^\dagger$Number of tasks not reported.} \\
\end{tabular}
}
\caption{\textbf{Comparison of \bench{} with existing forecasting benchmarks.} \bench{} supports evaluation on arbitrary what-if scenarios and lets difficulty be varied systematically, from few to many covariates and from few to many past observations of their effects, exposing differences between models that existing benchmarks cannot show.}
\label{tab:benchmark-comparison}
\end{table*}

\section{Related Work}
\label{sec:related}

Below, we position \bench{} with respect to existing forecasting benchmarks and related work on causal inference. Background on the models we evaluate is given in \Cref{app:extended_related_work}.

\subsection{Time Series Forecasting Benchmarks}
\label{sec:related-bench}

Early benchmarks such as the M-competitions and the Monash archive standardized evaluation but largely measure univariate accuracy~\citep{makridakis2000m3,makridakis2018m4,godahewa2021monash,alexandrov2020gluonts}, while more recent ones target TSFMs with large, leakage-free test sets, such as GIFT-Eval~\citep{aksu2024gifteval}, TSFM-Bench~\citep{li2024tsfmbench}, and TIME~\citep{qiao2026time}. All of them, however, evaluate univariate forecasting and do not contain covariates.
fev-bench~\citep{shchur2025fevbench} includes tasks with real-world covariates drawn from datasets such as M5~\citep{makridakis2022m5}, the EPF toolbox~\citep{lago2021epftoolbox}, Rohlik Sales~\citep{rohlik2025kaggle}, and Walmart~\citep{walmart2014kaggle}.
It evaluates only on covariate values that appear in the data, so it cannot tell whether a model forecasts accurately at unseen values. It also measures performance only under the conditions each dataset happens to provide, so it cannot reveal where a model systematically fails.
It evaluates only on the covariate values its datasets happen to contain, so it cannot tell whether a model forecasts accurately when covariates take values beyond that range. It also measures performance only in aggregate over the conditions each dataset provides, so it cannot reveal where a model systematically fails at using covariates.
\bench{} instead controls these conditions directly: it sets covariates to unseen values to test extrapolation, and varies one axis at a time to pinpoint where a model fails.
\citet{xu2024iatsf} also benchmark forecasting under interventions; they study forecasting under textual events, rather than as numeric covariates as we do; this places it in the different setting of multimodal forecasting, as in \citet{williams2025contextiskey}, where covariate values and their effects are not tested.
Closest to our work, \citet{berthelier2026covrelationships} probe whether two foundation models capture very simple covariate-target relationships - identity, sum, aggregate, and quadratic, and find that the two behave differently even on such simple relationships.
\bench{}, by contrast, is large-scale and diverse (\Cref{tab:benchmark-comparison}): it imposes many covariate-target relationships grounded in mechanistic simulators, and varies axes such as the future type (\emph{Replay} or \emph{Novel}), the number of covariates, and the amount of in-context evidence.
Across these conditions, we evaluate a broad range of numerical foundation models and LLMs.

\subsection{Causality and Interventional Estimation}
\label{sec:related-causal}

Measuring whether a forecast responds correctly to a covariate is an interventional question: it requires the target under new, chosen covariate settings, which real data seldom provides~\citep{holland1986statistics,imbens2015causal}.
The causal-inference community has long worked around this by evaluating on simulators and semi-synthetic data whose data-generating process is known, so the effect of any intervention can be computed exactly~\citep{hill2011bayesian,dorie2019automated}; \bench{} borrows this methodology, but applies it to forecasting rather than treatment-effect estimation.
A related thread predicts outcomes under time-varying treatments~\citep{robins2000msm}, such as Counterfactual Recurrent Networks~\citep{bica2020crn}, which forecast counterfactual trajectories and score them by forecast error on simulated environments. These methods are trained on each system and rely on identification assumptions. Similarly, DoTime~\citep{thumm2026dotime} generates fully synthetic causal models with interventions to benchmark treatment-effect estimation. We instead evaluate zero-shot forecasters, which must learn the covariate effects in-context from the history alone, with no training on the system.

%% file: inputs/benchmark_v2.tex
\providecommand{\bench}{\textsc{WhatIfBench}}
\providecommand{\Xb}{\ensuremath{\mathbf{X}}}
\providecommand{\Zb}{\ensuremath{\mathbf{Z}}}
\providecommand{\zb}{\ensuremath{\mathbf{z}}}

\section{\bench{}}
\label{sec:benchmark}

\subsection{Problem setting}
\label{sec:benchmark-problem}

Let $\Xb_H=[X_1,\dots,X_t]$ be the observed history of a scalar target, each $X_\tau\in\mathbb{R}$. Assume the target is driven by $K$ exogenous, numerical covariates, observed over the same period as $\Zb_H=[\zb_1,\dots,\zb_t]$, each $\zb_\tau\in\mathbb{R}^K$.  %
Canonically, forecasting concerns predicting the target's evolution $\Xb_F=[X_{t+1},\dots,X_T]$ from this history alone, by estimating $P(\Xb_F\mid\Xb_H,\Zb_H)$. When the covariates' future values $\Zb_F=[\zb_{t+1},\dots,\zb_T]$ are themselves known, a forecaster can condition on them as well, to instead estimate $P(\Xb_F\mid\Xb_H,\Zb_H,\Zb_F)$. 
The future covariates $\Zb_F$ allow one to express various future \textit{what-if scenarios}, and produce a conditional forecast of the target under each scenario.
This can reveal which drivers move the target and how strongly, and supports planning against many possible futures rather than the single one implied by holding covariates at their most recent values~\citep{brodersen2015causalimpact}.

The core challenge is to learn the covariate-target relationships the history reveals, separating the target's own variation from covariate-driven changes, and then apply them at the arbitrary values $\Zb_F$ the scenario specifies.
Since a scenario may set $\Zb_F$ to values the history has shown, or to values beyond anything it contains, the relationships must be learned well enough to extrapolate: to apply at covariate values never seen, and predict the target at levels it never reached.
We call this setting \emph{intervention-aware forecasting}, as it requires models to learn from past ``interventions'' in the system - from the past changes in the covariates that drive the target, and to respond to the future interventions a scenario specifies \footnote{We use ``intervention'' to mean setting the values of the covariates, which are among the parents of the target, i.e.\ $\mathrm{do}(\Zb=\zb)$~\citep{pearl2009causality}. The target is never intervened on.}.
We operate in the \textit{in-context learning} setting, where these relationships must be learned \textit{in-context} from the dynamics seen in the history, across the segments in which the covariates take on different values~\citep{shchur2025fevbench}. 

\subsection{Semi-synthetic design}
\label{sec:benchmark-design}

\paragraph{Motivation.}
Evaluating intervention-aware forecasting requires assessing whether a model forecasts the target accurately under different settings of the future covariates. Real-world data, however, gives only one realization of the covariates and target, and A/B tests that change covariates to observe their effects are usually infeasible~\citep{holland1986statistics,imbens2015causal}. We instead need an environment where we control the covariates, record their effects on the target, and test how sensitive a model is to changing them in the future.
Research in causality has long addressed this with \emph{semi-synthetic} data, where a known, generated process supplies exact counterfactuals~\citep{hill2011bayesian,dorie2019automated}.
Building on this principle, we introduce \bench{}, a benchmark for intervention-aware forecasting that combines realistic time series from mechanistic simulators of real-world systems with controlled covariate-target relationships whose effects are known exactly.

\begin{figure*}[t]
  \centering
  \includegraphics[width=\linewidth]{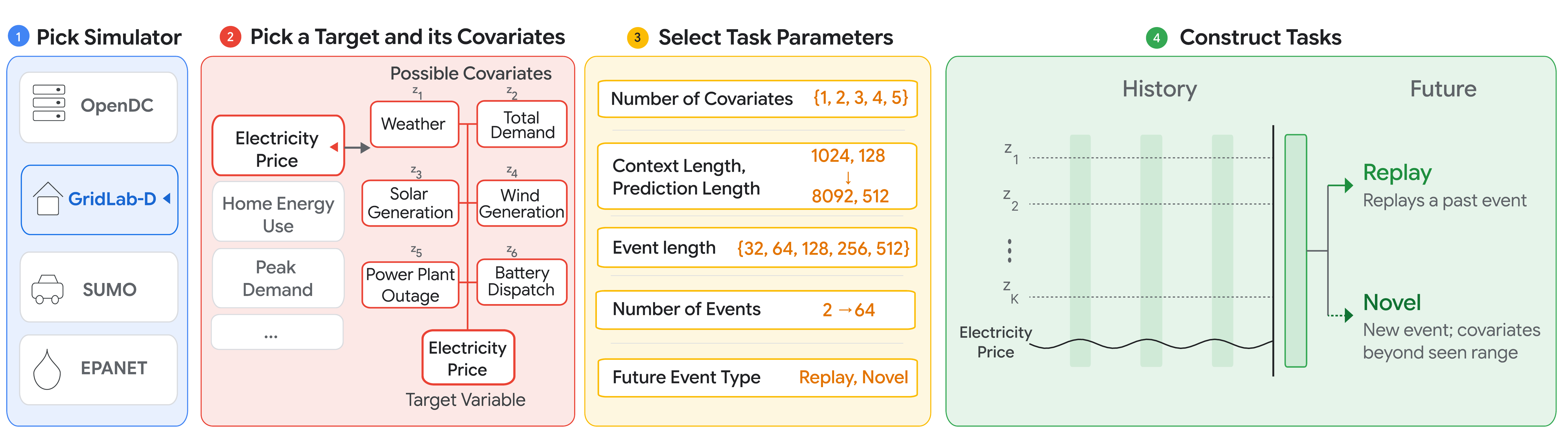}
  \caption{\textbf{The \bench{} construction recipe.}: We \textbf{(1)}~pick a simulator and a base setting; \textbf{(2)}~choose a target and the covariates that drive it; \textbf{(3)}~select the parameters of the task; finally \textbf{(4)}~construct the events in the history, to produce two futures over the same history. This yields $2$ tasks. We generate the benchmark by repeating this process for every target across all simulators, and over all axes: number of covariates, number of events, event length, and context length.}
  \label{fig:framework}
\end{figure*}

\paragraph{The simulators.}
We use four established, open-source simulators from distinct domains, each developed by domain experts and widely used in research and practice to study and plan real-world systems:
\textbf{OpenDC}~\citep{iosup2017opendc,mastenbroek2021opendc,andreadis2022capelin} models a data center's resource usage for capacity planning; \textbf{GridLAB-D}~\citep{chassin2008gridlabd,schneider2010cvr} models a power-distribution grid and the buildings on it; \textbf{SUMO}~\citep{lopez2018sumo,lu2023speedlimits} models road traffic vehicle by vehicle; and \textbf{EPANET}~\citep{rossman2000epanet,kepa2025hydraulic} models pressurized water-distribution networks.
\paragraph{Targets, covariates, and their relationships.}
We run each simulator on a fixed \emph{base setting}: a real configuration of the system together with the recorded input that drives it -- for example, a data center's setup and the demand placed on it. This produces time series of the system's variables, such as the CPU utilization of each host, as they evolve under the simulator's dynamics.
From each run, a system variable serves as the target and covariates drawn from the simulator's adjustable properties drive it. On top of these initial targets and their associated covariates, we impose controlled relationships that govern how changes in the covariates affect the target, grounded in each system, so their effect is known exactly at any value they take. This gives us a semi-synthetic environment of realistic time series targets and covariates with known, exactly reproducible effects.
The rationale behind this design, the target variables we extract from each simulator, and the relationships we impose for how changes in the covariates affect the target are given in \Cref{app:benchmark}. %

\definecolor{covcore}{HTML}{1A73E8}   %
\definecolor{covreq}{HTML}{D93025}     %
\definecolor{covtgt}{HTML}{188038}     %
\providecommand{\solidkey}{\raisebox{2pt}{\rule{0.35cm}{0.9pt}}}
\providecommand{\dashedkey}{\raisebox{2pt}{\rule{1.5pt}{0.9pt}\hspace{1.5pt}\rule{1.5pt}{0.9pt}\hspace{1.5pt}\rule{1.5pt}{0.9pt}}}

\subsection{Events}
\label{sec:benchmark-construction}

\paragraph{The building block.}

An \textit{event} is an intervention held over a bounded window of time, acting on one or more covariates at once. Within the window the covariates are moved away from their base setting, and the target responds for as long as the covariates are held there.
Such a setup reflects how change happens in real systems, where discrete interventions move the drivers over a bounded window and the target responds, a structure long studied as interrupted time series and event studies~\citep{brodersen2015causalimpact,box1975intervention}.

\paragraph{Event design.}

An event specifies two properties for each covariate it acts on: its \textit{magnitude} - how far to move it, and its \textit{shape} - how it moves across the window.
For the magnitude, we apply a \textit{multiplicative factor} to the base value, which scales the covariate and moves the target by the amount the imposed relationship prescribes. This relationship is fixed for each covariate-target pair and defined at every factor value, but need not be proportional: it may be inverse, sublinear, or piecewise, so a factor of two on the covariate need not mean a factor of two on the target. Within the factors we use, each relationship moves the target in one direction and never reverses, so it can be learned from the history.
For the shape, we use two: a step, which holds the new value constant throughout, and a gaussian bell, which rises to it and decays back. Each covariate in our benchmark is assigned one.
Additional details are given in \Cref{app:simulators}.

\subsection{Task structure and axes}
\label{sec:benchmark-axes}

\paragraph{From events to tasks.}
A \bench{} task simply places a sequence of events in the history, each leaving its effect on the target, and asks the model to forecast the target's response to one further event in the future. We build the tasks of \bench{} by varying five axes of this structure, described next (illustrated in \Cref{fig:framework}).

\paragraph{Replay and Novel Futures.}
The first axis is the future event: it is designed to evaluate whether a model can apply the relationships learned from history to \textit{arbitrary} future covariate values. We generate tasks with two futures over the same history, differing only in the future covariate values. We call the choice between them the \emph{future event type}:
\begin{itemize}[noitemsep, leftmargin=*]
  \item \textbf{Replay}. The future event takes on the covariate values of a past event, so the covariates take values they have already taken in the history.
  \item \textbf{Novel}. The future event is unseen: the covariates are driven to values beyond the range seen in the history.
\end{itemize} 
In both cases, the model must forecast the target's intrinsic dynamics (its trend, seasonality, and level) and, on top of that, the covariate effect at the specified values.
The two futures give complementary evaluation sets: Replay measures a model's basic covariate-conditioned forecasting, while a model that has learned the relationships \textit{robustly} should do equally well on Novel. This is closely tied to the principle of invariance in causality~\citep{peters2016causal,rojas2018invariant}: a model that has captured the true mechanism linking the covariates to the target should transfer to new covariate values.\footnote{Novel thus tests this consequence of learning the mechanism. Doing well on Novel does not in turn guarantee that the mechanism has been learned, since Novel covers a finite set of covariate shifts; we use it to measure how well this transfer holds in practice.} The degradation from Replay to Novel indicates whether a model is reliable for arbitrary scenario analysis. \Cref{fig:task} illustrates the structure of \bench{} tasks.

\paragraph{Covariate settings.}
The second axis is the number of covariates in a task $K\in\mathcal{K}=\{1,\dots,5\}$, which sets how many covariates every event acts on at once. Across the events in the history, these covariates take a range of values, from which a model can learn how each acts on the target and how they act together. Larger $K$ means more relationships to learn, and therefore harder tasks.

\begin{wraptable}[18]{r}{0.52\linewidth}
\centering
\small
\setlength{\tabcolsep}{4pt}
\resizebox{\linewidth}{!}{%
\begin{tabular}{llc}
\toprule
\multicolumn{3}{l}{\emph{Benchmark composition}} \\
\midrule
Simulator & Domain & \#\,Targets \\
\midrule
OpenDC~\citep{iosup2017opendc}    & Data center   & $22$ \\
GridLAB-D~\citep{chassin2008gridlabd} & Power grid    & $11$ \\
SUMO~\citep{lopez2018sumo}      & Road traffic  & $30$ \\
EPANET~\citep{rossman2000epanet}    & Water network & $37$ \\
\midrule
\textbf{Total} & & $\mathbf{100}$ \\
\midrule
\multicolumn{3}{l}{\emph{Design axes}} \\
\midrule
Axis & Values & Count \\
\midrule
Targets              & across $4$ domains                        & $100$ \\
Covariates           & $1,2,3,4,5$                               & $5$   \\
Context / prediction length           & $1$k/$128$, $2$k/$256$, $4$k/$384$, ${\sim}8$k/$512$--$864$ & $4$   \\
Event length         & $32,64,128,256,512$                       & $5$   \\
\# events            & $1$--$64$ (fewest, most)                  & $2$   \\
Futures              & Replay, Novel                             & $2$   \\
\midrule
\multicolumn{2}{l}{\textbf{Tasks Generated}} & $\mathbf{37{,}665}$ \\
\bottomrule
\end{tabular}%
}
\caption{\textbf{The \bench{} instantiation.} The number of series contributed by the four simulators, and the axes we vary over them to build the tasks. The exact context and prediction lengths per simulator are given in \Cref{tab:ctx-pred-pairs}.} 
\label{tab:instantiation}
\end{wraptable}
\paragraph{Context and prediction length.}
The third axis is the context and prediction length: events sit within a history of fixed length, the \emph{context length} -- and the model forecasts a fixed horizon beyond it, the \emph{prediction length}. We use four pairs, $c\in\mathcal{C}=\{C_1,\dots,C_4\}$, from the smallest $C_1$ (context $1024$, prediction $128$) to the largest $C_4$ (context about $8{,}000$, prediction $512$ to $864$, depending on the simulator; details in App. \Cref{tab:ctx-pred-pairs}). The context sets the budget the events must fit within; for the longest events, which would not fit at the smallest contexts, we extend the context slightly (\Cref{app:task-layout}).

\paragraph{Event length.}
The fourth axis is the event length $\ell\in\mathcal{L}=\{32,64,128,256,512\}$, which sets how long each event holds its covariates in place, spanning brief interventions to sustained ones. A short event gives only a brief glimpse of a relationship, whereas a long one lets the model see it settle and hold, but takes up more of the history and so leaves room for fewer events.

\paragraph{Number of events.}
The fifth and final axis is the number of events $E$, which controls how often the covariates are set across the history: more events expose their relationships at more values, which can let a model fit them more robustly. How many fit is bounded by the context and event lengths -- at a context of $1024$, only a few long events fit, but far more short ones do. For each event length and context/prediction pair, we build tasks at two settings: the fewest events (one or two) and the most the context allows, up to $64$.
We design \bench{} around real-world regimes rather than guaranteed identifiability: since limited-evidence tasks (e.g., many covariates, few events) do occur in practice, we include them and expect a reliable forecaster to respond with calibrated uncertainty rather than overconfidence; we discuss this design choice further in \Cref{app:design}.

\paragraph{Instantiating \bench{}.}

To recap the recipe (\Cref{fig:framework}): we \textbf{(1)}~pick a simulator and a base setting; \textbf{(2)}~choose a target variable and the covariates that drive it; \textbf{(3)}~select the task's parameters; and \textbf{(4)}~construct the events in the history, producing two futures over the same history.
We repeat this recipe over all targets and every combination of the axes above, giving $40{,}000$ tasks in all (\Cref{tab:instantiation}).
Out of these, we keep only tasks whose covariates have a non-negligible effect in every event, leaving $37{,}665$ of the $40{,}000$ tasks. 
Additional details on the benchmark and the various design choices are in \Cref{app:simulators}. Example tasks are provided in \Cref{app:task-examples}.

\subsection{Metrics and aggregation}
\label{sec:benchmark-aggregation}

\paragraph{Metrics}
We score each task with standard forecasting metrics. Our primary metric is the weighted quantile loss (wQL), which evaluates probabilistic forecasts across a set of quantile levels; for point-forecast accuracy we also report the weighted absolute percentage error (WAPE) on the median forecast (definitions in \Cref{app:metrics}).

\paragraph{Aggregating performance across tasks}
Following fev-bench~\citep{shchur2025fevbench}, we summarize the per-task errors (wQL or WAPE) along two dimensions. The primary one is the \emph{skill score}, a model's relative reduction in error against a fixed naive baseline that repeats the last observed value. As a secondary measure, we use the \emph{average win rate}: the fraction of tasks on which a model has a lower error than another model, averaged over all other models in the set (higher is better for both). We use it to compare models directly against one another. We report win rates in \Cref{app:win-rates} and give formal definitions of both in \Cref{app:metrics}.

\subsection{Models and Evaluation Protocol}
\label{sec:benchmark-models}

\paragraph{Numerical forecasters.}
Our main candidates for evaluation are foundation models that can forecast time series while consuming past and future covariates in-context. Six such models exist, and we evaluate all of them: Chronos-2~\citep{chronos2}, t0-gamma~\citep{t0alpha}, Timestral~\citep{timestral}, TiRex-2~\citep{tirex2}, TS-ICL~\citep{tsicl}, and TimesFM-3.0~\citep{timesfm3}. We also evaluate TabPFN-3.5~\citep{tabpfn35}, a tabular foundation model, which we run for time series forecasting through the TabPFN-TS harness~\citep{tabpfnts}. These models output nine quantile levels from $0.1$ to $0.9$ directly.

\paragraph{Large language models.}
We also evaluate frontier LLMs such as Gemini-3.1-Pro~\citep{gemini31pro} and Claude-Opus-4.8~\citep{claudeopus48}, prompted directly with the target and covariates as text. We run each at two reasoning efforts, Medium and High, to measure the effect of reasoning. Unlike the foundation models, the LLMs are prompted for the median rather than for quantiles: we find that prompting them for many quantiles exceeds the output length and causes formatting failures, so we score their wQL on the median alone, at the $0.5$ level, which disadvantages them under probabilistic metrics. Implementation details on all models are given in \cref{app:experimental-setup}.

\paragraph{Baselines.}
We include two kinds of baselines. The naive baseline copies the last observed value and does not use covariates. The regression baselines - Linear Regression, Ridge Regression, and XGBoost - use both past and future covariates, in their simplest form: we fit each one on a task's history with one training row per time step, predicting the target from the covariates at that step only, and then predict each future step separately from its known covariates. Our problem setting does not assume a separate training or validation set for each task, so these baselines cannot be tuned on held-out data. We use the simplest versions to understand how far the covariates alone take a forecaster; variants with minor tuning on the test set are explored in \Cref{app:tuned-regression}. Like the LLMs, they only provide point forecasts.
We use the naive baseline as the reference for the skill score, since it gives the same forecast on Replay and Novel futures over a shared history, which makes it a consistent reference to measure skill against. We also explore using the covariate-aware Linear Regression as the reference in \Cref{app:linear-baseline}.

%% file: inputs/results_v2.tex
\providecommand{\bench}{\textsc{What-IfBench}}
\definecolor{llmhue}{HTML}{D93025}
\definecolor{tsfmhue}{HTML}{1A73E8}
\definecolor{basehue}{HTML}{5F6368}
\definecolor{llmpale}{HTML}{F4C1BE}   %
\definecolor{tsfmpale}{HTML}{BAD5F8}  %
\definecolor{basepale}{HTML}{CFD0D2}  %
\providecommand{\base}[1]{\textcolor{basehue}{#1}}
\providecommand{\swatch}[1]{{\setlength{\fboxsep}{0pt}%
  \colorbox{#1}{\rule{0.9em}{0pt}\rule{0pt}{0.65em}}}}
\providecommand{\takeaway}[1]{%
  \par\medskip\noindent
  \colorbox{black!8}{%
    \parbox{\dimexpr\linewidth-2\fboxsep\relax}{\small\textbf{Takeaway.} #1}}%
  \par\medskip}

\input{tables/table_v10_summary_paper.tex}

\vspace{-0.3cm}
\section{Results}
\label{sec:results}   %

\subsection{Performance on \bench{}}
\label{subsec:performance-main}

We present results on \bench{} in \Cref{tab:leaderboard}.
The regression baselines, in their simplest form, already obtain substantial skill over the naive baseline: XGBoost reaches $42.9\%$, and even Ridge and Linear Regression reach $25.4\%$ and $26.2\%$.
Among the foundation models, t0-gamma ($64.5\%$), TimesFM-3 ($61.6\%$) and TS-ICL ($61.3\%$) obtain the best skill scores, about $19$ to $22$ points above XGBoost.
Timestral ($59.2\%$) follows, then TabPFN-3.5 and Chronos-2 ($50.8\%$ each), about $8$ points above XGBoost, followed by TiRex-2 ($46.7\%$), about $4$ points above it.
Notably, the LLMs obtain the highest skill scores on the benchmark, with Claude-Opus-4.8 (High) at $72.5\%$ followed by Gemini-3.1-Pro (High) at $70\%$.
Reasoning helps here: moving from Medium to High effort raises skill from $70.3\%$ to $72.5\%$ for Claude and from $58.2\%$ to $70\%$ for Gemini.
On inference time, the foundation models are the fastest: most run in under half a second per series, and TabPFN-3.5 is the slowest at $21$s. The LLMs are far slower, at around $300$ seconds per series, over an order of magnitude beyond TabPFN-3.5, and almost $4$ orders of magnitude beyond the rest.
\takeaway{Frontier LLMs score highest. Among the foundation models, t0-gamma leads, followed by TimesFM-3 and TS-ICL, and all of them are ahead of a simple, untuned XGBoost baseline that uses the covariates. The LLMs are, however, about $4$ orders of magnitude slower at inference.}

\begin{figure}[t]
  \centering
  \includegraphics[width=\linewidth]{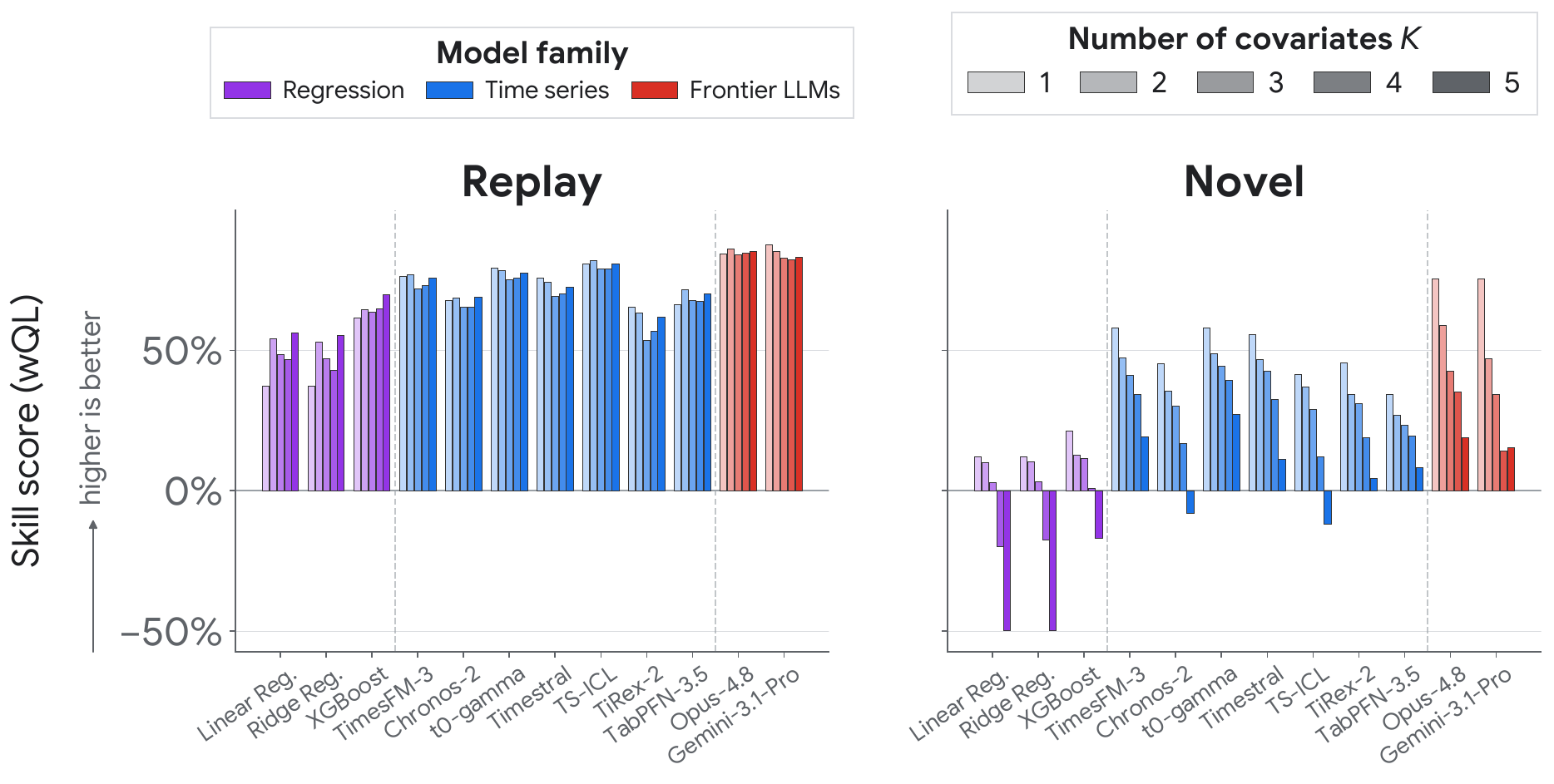}
  \caption{Skill scores on tasks with different number of covariates on \bench{} on Replay (left) and Novel (right) future types respectively.}
  \label{fig:v10-by-k-wql}
  \vspace{-1.2em}  %
\end{figure}

\subsection{Performance on Replay vs Novel tasks}

\Cref{fig:replay-vs-novel}, in the introduction, compares model performance on the Replay and Novel tasks in \bench{}.
On Replay tasks, every model achieves a high skill score, indicating that the models forecast well when the future event repeats covariate patterns already seen in the history. 
The LLMs are strongest here, with Claude Opus 4.8 (High) at $84.8\%$ and Gemini 3.1 Pro (High) at $84.3\%$, followed by TS-ICL ($80.4\%$) and t0-gamma ($77.3\%$) among the foundation models. The regression baselines are competitive on Replay: XGBoost reaches $65.0\%$.
On Novel futures, however, the skill of every model drops sharply, and the LLMs fall just as much: Claude Opus 4.8 drops from $84.8\%$ to $50.3\%$ and Gemini 3.1 Pro from $84.3\%$ to $42.6\%$, while TS-ICL falls furthest among the foundation models, from $80.4\%$ to $23.6\%$.
This degradation appears for every model, indicating the difficulty of extrapolating to unseen values, and points to a lack of robustness.
The regression baselines fall the most: XGBoost drops from $65.0\%$ to $6.7\%$ and Linear Regression from $49.0\%$ to $-6.7\%$, worse than the naive baseline.
Novel futures are thus where the LLMs and foundation models have their largest advantage over the regression baselines: TimesFM-3 is only $10$ points above XGBoost on Replay but $35$ points above it on Novel.
\takeaway{Every model forecasts well on Replay, but skill drops sharply on Novel. The regression baselines fall the most, so Novel is where the LLMs and foundation models have their largest advantage over simple models. Yet even the LLMs and foundation models drop substantially, so strong Replay performance does not carry over to Novel futures. Closing this gap is the core challenge \bench{} exposes.}

\begin{figure}
    \centering
    \includegraphics[width=\linewidth]{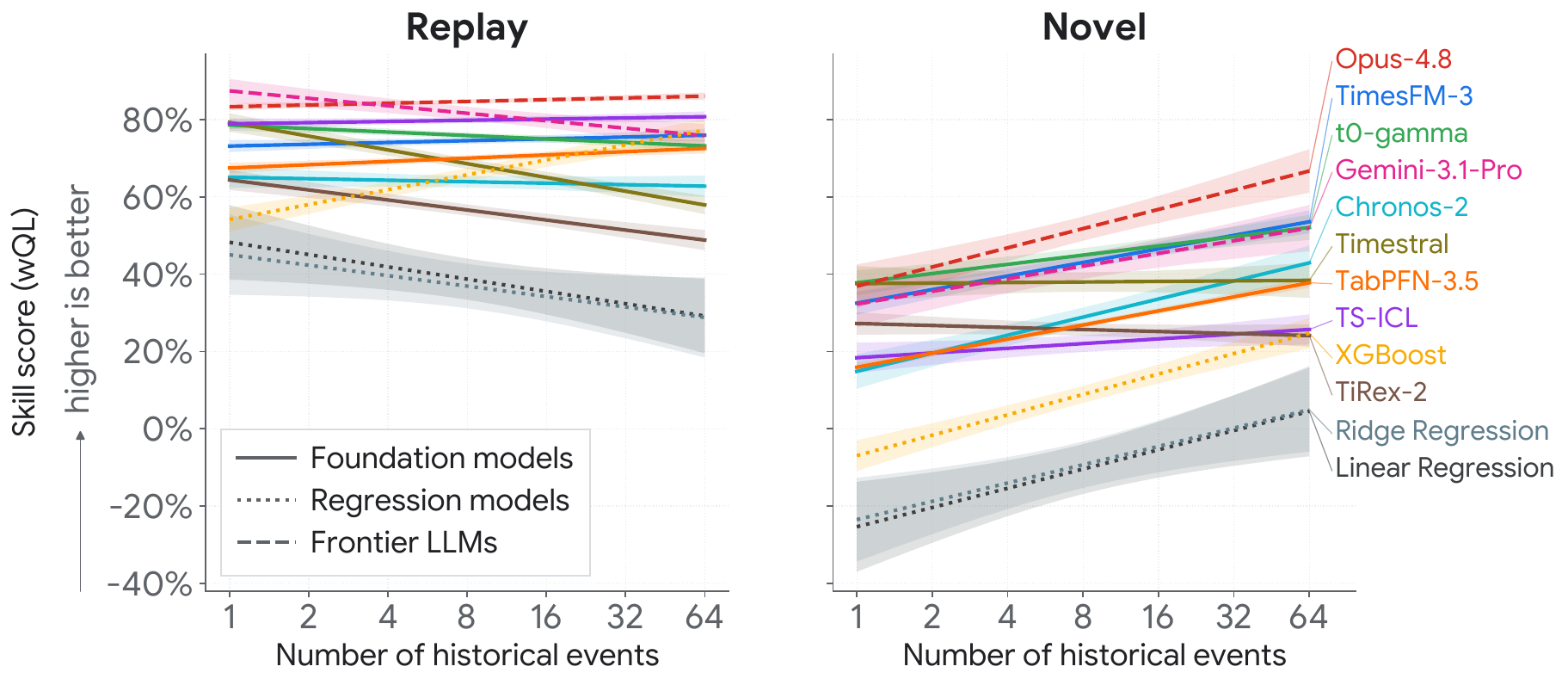}
    \caption{Skill scores with different number of historical events on \bench{}, on tasks with Replay (left) and Novel (right) futures respectively. Lines are least-squares trends of skill on $\log_2$ of the number of events; they summarise the direction of change.}
    \label{fig:v10-kevents-pair-wql}
    \vspace{-1.2em}  %
\end{figure}

\subsection{Effect of the number of covariates}

\Cref{fig:v10-by-k-wql} shows how skill varies with the number of covariates $K$, for Replay and Novel futures.
On Replay futures, skill varies marginally as the number of covariates grows, for both foundation models and LLMs, which indicates that adding covariates does not make the Replay task harder.
However, on Novel futures, skill decreases consistently as the number of covariates increases. Claude Opus 4.8 (High) and Gemini 3.1 Pro (High) both reach $75\%$ at one covariate, far ahead of the other models. But their performance falls steeply, to $19\%$ and $15\%$ respectively at five covariates. By five covariates the best foundation models catch up with them: t0-gamma is strongest at $27\%$, ahead of both LLMs, and TimesFM-3 ($19\%$) is level with Claude Opus 4.8 (High). This drop also reshuffles the ranking: TS-ICL, for instance, is among the best models on Replay at every covariate count (around $80\%$), yet drops to $-12\%$ at five covariates on Novel.
The regression baselines end lowest at five covariates: Linear Regression drops from $12.1\%$ to $-49.8\%$ and Ridge Regression from $12.2\%$ to $-49.9\%$, so even a linear model loses $62$ points on Novel.
Since the same models handle five covariates as well as one on Replay, the difficulty lies not in the number of covariates but in extrapolating their joint effect to unseen values, so strong Replay performance does not carry over to the Novel futures that scenario analysis requires.
\takeaway{The number of covariates has no clear effect on Replay, but on Novel skill degrades progressively for every model as more are set together, including the regression baselines. The LLMs lead at one covariate, but by five covariates, t0-gamma is ahead of the best LLM and TimesFM-3 is on par with it. This consistent drop points again to Novel futures as the core challenge, across all covariate settings.}

\vspace{-0.3cm}
\subsection{Effect of the number of historical events}

\Cref{fig:v10-kevents-pair-wql} shows skill as a function of the number of historical events in the context, separately for Replay and Novel futures. On Replay futures, more history brings only marginal gains: skill stays roughly flat for most models. XGBoost is the one model that clearly gains, by about $25$ points.
This shows that Replay futures can be predicted with very little history, as little as a single event. However, on Novel futures the trend is far stronger: skill rises sharply as the number of historical events grows. Both the LLMs and the foundation models improve substantially: Claude Opus 4.8 (High) more than doubles from $24\%$ to $58\%$ and Gemini 3.1 Pro (High) from $19\%$ to $42\%$, while Chronos-2 rises the most among the foundation models, from $6\%$ to $35\%$.
The regression baselines gain as well, Linear Regression from $-27\%$ at one event to $9\%$ at $64$, but they remain well below the best models.
This indicates most models can use a longer history to forecast Novel futures more accurately.
We find the same pattern with event length and context length (\Cref{app:axis-ablations}): on Novel futures, skill improves in both cases whenever the context can hold more past events - i.e., both when events are shorter or the context is longer.
Even with the most evidence, however, large differences between models remain: more history helps nearly every model, but models differ widely in how effectively they use it. We analyze this further in \Cref{app:axis-ablations}.
\takeaway{Replay needs very little history, while on Novel most models improve as they see more past events, including the regression baselines. Even with the most events, however, Novel futures stay well below Replay, and models differ widely in how well they use the history.}

%% file: tables/table_v10_summary_paper.tex
\begin{table}[t]
\centering
\small
\renewcommand{\arraystretch}{0.96}
\vspace{-1.5mm}
\begin{tabular*}{\textwidth}{@{\extracolsep{\fill}}l ll c@{}}
\toprule
\textbf{Model} & \textbf{wQL Skill} & \textbf{WAPE Skill} & \textbf{Inf. time / series (s)} \\
\midrule
\multicolumn{4}{@{}l}{\textbf{\textit{Frontier LLMs}}} \\
\quad Claude-Opus-4.8 (High) & \makebox[3.4em][r]{\textbf{72.5\%}}\,\textcolor{blue!55!black}{\rule{14.50mm}{1.3ex}} & \makebox[3.4em][r]{\textbf{72.5\%}}\,\textcolor{black!35}{\rule{14.50mm}{1.3ex}} & 270.13 \\
\quad Claude-Opus-4.8 (Medium) & \makebox[3.4em][r]{\underline{70.3\%}}\,\textcolor{blue!55!black}{\rule{14.06mm}{1.3ex}} & \makebox[3.4em][r]{\underline{70.3\%}}\,\textcolor{black!35}{\rule{14.06mm}{1.3ex}} & 206.29 \\
\quad Gemini-3.1-Pro (High) & \makebox[3.4em][r]{70.0\%}\,\textcolor{blue!55!black}{\rule{14.00mm}{1.3ex}} & \makebox[3.4em][r]{70.0\%}\,\textcolor{black!35}{\rule{14.00mm}{1.3ex}} & 355.28 \\
\quad Gemini-3.1-Pro (Medium) & \makebox[3.4em][r]{58.2\%}\,\textcolor{blue!55!black}{\rule{11.64mm}{1.3ex}} & \makebox[3.4em][r]{58.2\%}\,\textcolor{black!35}{\rule{11.64mm}{1.3ex}} & 97.35 \\
\addlinespace[0.5ex]
\multicolumn{4}{@{}l}{\textbf{\textit{Numerical Foundation Models}}} \\
\quad t0-gamma & \makebox[3.4em][r]{64.5\%}\,\textcolor{blue!55!black}{\rule{12.90mm}{1.3ex}} & \makebox[3.4em][r]{58.4\%}\,\textcolor{black!35}{\rule{11.68mm}{1.3ex}} & 0.043 \\
\quad TimesFM-3 & \makebox[3.4em][r]{61.6\%}\,\textcolor{blue!55!black}{\rule{12.32mm}{1.3ex}} & \makebox[3.4em][r]{54.5\%}\,\textcolor{black!35}{\rule{10.90mm}{1.3ex}} & 0.419 \\
\quad TS-ICL & \makebox[3.4em][r]{61.3\%}\,\textcolor{blue!55!black}{\rule{12.26mm}{1.3ex}} & \makebox[3.4em][r]{55.3\%}\,\textcolor{black!35}{\rule{11.06mm}{1.3ex}} & 0.083 \\
\quad Timestral & \makebox[3.4em][r]{59.2\%}\,\textcolor{blue!55!black}{\rule{11.84mm}{1.3ex}} & \makebox[3.4em][r]{52.7\%}\,\textcolor{black!35}{\rule{10.54mm}{1.3ex}} & 0.048 \\
\quad TabPFN-3.5 & \makebox[3.4em][r]{50.8\%}\,\textcolor{blue!55!black}{\rule{10.16mm}{1.3ex}} & \makebox[3.4em][r]{42.3\%}\,\textcolor{black!35}{\rule{8.46mm}{1.3ex}} & 21.26 \\
\quad Chronos-2 & \makebox[3.4em][r]{50.8\%}\,\textcolor{blue!55!black}{\rule{10.16mm}{1.3ex}} & \makebox[3.4em][r]{42.0\%}\,\textcolor{black!35}{\rule{8.40mm}{1.3ex}} & 0.067 \\
\quad TiRex-2 & \makebox[3.4em][r]{46.7\%}\,\textcolor{blue!55!black}{\rule{9.34mm}{1.3ex}} & \makebox[3.4em][r]{36.5\%}\,\textcolor{black!35}{\rule{7.30mm}{1.3ex}} & 0.079 \\
\addlinespace[0.5ex]
\multicolumn{4}{@{}l}{\textbf{\textit{Regression Models$^\dagger$}}} \\
\quad XGBoost & \makebox[3.4em][r]{42.9\%}\,\textcolor{blue!55!black}{\rule{8.58mm}{1.3ex}} & \makebox[3.4em][r]{42.9\%}\,\textcolor{black!35}{\rule{8.58mm}{1.3ex}} & 0.068 \\
\quad Linear Regression & \makebox[3.4em][r]{26.2\%}\,\textcolor{blue!55!black}{\rule{5.24mm}{1.3ex}} & \makebox[3.4em][r]{26.2\%}\,\textcolor{black!35}{\rule{5.24mm}{1.3ex}} & 0.002 \\
\quad Ridge Regression & \makebox[3.4em][r]{25.4\%}\,\textcolor{blue!55!black}{\rule{5.08mm}{1.3ex}} & \makebox[3.4em][r]{25.4\%}\,\textcolor{black!35}{\rule{5.08mm}{1.3ex}} & 0.002 \\
\addlinespace[0.5ex]
\multicolumn{4}{@{}l}{\textbf{\textit{Baseline}}} \\
\quad Naive* & \makebox[3.4em][r]{0.0\%}\,\textcolor{blue!55!black}{\rule{0.00mm}{1.3ex}} & \makebox[3.4em][r]{0.0\%}\,\textcolor{black!35}{\rule{0.00mm}{1.3ex}} & 0.072 \\
\bottomrule
\end{tabular*}
\caption{Performance of models on \bench{}, measured as skill score under both wQL (probabilistic) and WAPE (point) metrics, alongside inference latency. The best score per column is \textbf{bolded} and the second-best is \underline{underlined}. Naive* represents a naive baseline which does not consume covariates. $^\dagger$The simplest, untuned versions of the regression models; tuned variants are in \Cref{app:tuned-regression}. Win rates are provided in \Cref{app:win-rates}.}
\label{tab:leaderboard}
\vspace{-2mm}
\end{table}

%% file: inputs/discussion_and_future_work.tex
\providecommand{\bench}{\textsc{WhatIfBench}}

\section{Discussion and Future Work}
\label{sec:discussion}

We propose \bench{}, a semi-synthetic benchmark for intervention-aware forecasting: its target series come from mechanistic simulators of real systems, on which we impose controlled effects that determine how the target responds to changes in the covariates.
Each task presents a history of events and asks a model to forecast the target's response to a further event, either a Replay event that repeats values seen in the history or a Novel event that sets them beyond that range; \bench{} varies the number of covariates, the number of events, the event length, and the context length.
Across these conditions, the frontier LLMs reach the highest skill scores overall, with the best foundation models close behind.
Yet every model forecasts well on Replay futures but sees its skill drop sharply on Novel futures, and the regression baselines collapse the most. Despite their own drop, the LLMs and foundation models thus keep their clearest advantage over simple models on Novel futures.
On Novel futures, skill also decreases as more covariates are set jointly, while more past events help.
These results point to generalizing to Novel futures as the central open problem, and open new directions for building models that forecast reliably under novel scenarios.

We discuss the limitations of \bench{} and future work in \Cref{app:future-work}.
Addressing the limitations of \bench{} opens natural extensions: richer system dynamics, and larger covariate spaces where many drivers are irrelevant, confounded, or misleading.
Beyond these, the covariate-aware forecasting ability of frontier LLMs, despite their far slower inference, opens up more flexible what-if analysis, letting users explore queries that are not directly expressible through covariates, such as those posed in natural language, which could bring scenario analysis to many more users.
Finally, beyond evaluating models, training them in such simulation environments could yield forecasters that learn a system in-context and reason over its possible futures - a step toward \emph{world models} for planning and decision-making~\citep{ha2018worldmodels,hafner2023dreamerv3,lecun2022jepa}.

%% file: inputs/appendix_benchmark.tex
\section{Extended Related Work}
\label{app:extended_related_work}

\subsection{Time Series Foundation Models}
\label{sec:related-tsfm}

Time series forecasting has moved from classical methods~\citep{expsmoothing,box2015time,hyndman2018forecasting} and deep networks~\citep{salinas2020deepar,benidis2022deeplearning,lim2021temporal,zhou2021informer,wu2021autoformer,nie2023patchtst,drouin2022tactis,ashok2024tactis} to time series foundation models (TSFMs), pretrained once on large corpora to forecast unseen series zero-shot, often outperforming dataset-specific models~\citep{rasul2023lag,goswami2024moment,woo2024unified,ansari2024chronos}.
These models were long univariate; using the external covariates many tasks depend on required task-specific fine-tuning~\citep{shchur2025fevbench}.
Some foundation models can use covariates only after such adaptation, such as Moirai~\citep{woo2024unified} and TTM~\citep{ekambaram2024tiny}.
The most recent TSFMs instead consume covariates in-context, inferring how they affect the target from the examples alone and factoring this into the forecast without any training~\citep{chronos2,timesfm3,tsicl,tirex2,t0alpha}.
We focus on this zero-shot setting, as in fev-bench~\citep{shchur2025fevbench}.
\bench{} evaluates these covariate-aware TSFMs for arbitrary scenario analysis: whether they can forecast the target accurately when covariates are set to unseen future values.

\subsection{Large Language Models for Forecasting}
\label{sec:related-llm}

Large language models (LLMs), though not built for forecasting, have also been repurposed for it: prompted with a numerical time series as text, a pretrained LLM can continue it zero-shot~\citep{gruver2023llmtime,requeima2024llmprocesses}, and later work uses LLMs as strong forecasting baselines~\citep{williams2025contextiskey,ashok2026beyond}.
Their performance improves with fine-tuning or adaptation for time series, as in Time-LLM~\citep{jin2023timellm} and UniTime~\citep{liu2024unitime}, and they have been explored for multimodal forecasting, using textual context such as news~\citep{williams2025contextiskey,liu2024timemmd,lee2025timecap,wang2024newsforecast}.
Our focus is on zero-shot forecasting, so we prompt LLMs directly rather than adapting them, as in a few recent forecasting studies~\citep{williams2025contextiskey,ashok2026beyond}.
We evaluate frontier LLMs in \bench{}, and find they are in fact the strongest models we test, outperforming covariate-aware TSFMs both on tasks with repeated future covariate values and on tasks requiring extrapolation to unseen values.

\section{Benchmark details}
\label{app:benchmark}

\subsection{Rationale behind semi-synthetic setup}
\label{app:rationale}

Intervention-aware forecasting needs an environment where the target's response to a covariate is known exactly at any value the covariate takes, including values never observed in the history. Real data does not meet this need: it gives one realization of the covariates and targets, and running controlled A/B tests to see how the target moves is usually infeasible~\citep{holland1986statistics,imbens2015causal}. This is what motivates a semi-synthetic setup, where a known, generated process supplies exact counterfactuals~\citep{hill2011bayesian,dorie2019automated}.

The simulators alone do not meet this need either. They are built mainly to generate realistic dynamics for a domain, and they do this well: running one on a base setting gives a target series that is realistic and forecastable on its own. Interventions can be performed on them, but not the kind we need. To construct our tasks we would have to re-run the simulator for every event and every covariate value, including the novel values that fall outside the range a simulator is calibrated for, and we found the simulators were not robust to this. Interventions in OpenDC often shift the system into a new regime, producing a change that is not forecastable from the history; in GridLAB-D most interventions required several attempts to run and some failed outright; in SUMO the covariates we wanted moved the target reliably only at large factors, too coarse to span the range that novel extrapolation needs; and across simulators, setting several covariates at once rarely produced a clear, separable effect on the target.

A further problem is noise. The simulators add stochastic variation to the target, so the ground-truth series carries noise unrelated to the covariates. This makes evaluation unreliable: even when a model uses the covariates correctly, the noise in the ground truth inflates the error and drives metrics down, so scores reflect the noise rather than the model's grasp of the covariate--target relationships. Imposing the relationships ourselves gives a clean ground truth, on which such scores are meaningful.

We therefore keep the simulators for what they are good at, the realistic baseline series, and impose the covariate--target relationships ourselves. %
Each relationship is derived from the simulator's documented model and reduced to a smooth, fixed form that holds at every factor value: faithful to the system under simplifying assumptions, but deliberately simple so it stays reliable and exactly reproducible, not an attempt to reproduce the simulator's exact dynamics. Under other operating regimes some effects could take a different sign or shape (for instance, a network quantity's response to added demand depends on where the bottleneck is), but we do not consider this a limitation for benchmarking: \bench{} does not test physics, only whether a model learns a covariate--target relationship from its context. Every foundation model sees only numbers, and the LLMs see dummy variable names (\Cref{app:experimental-setup}), so no model can tell one form of a relationship from another except through the data, and results do not depend on whether a relationship matches a practitioner's expectation. 
We further document every relationship used in the benchmark (\Cref{app:simulators}).

This setup has limitations. The relationships are hand-engineered, and building them for each simulator took manual effort. We see domain experts building bespoke, faithful benchmarks for each use case as the ideal, and leave this to future work. We view \bench{} as complementary to benchmarks such as fev-bench~\citep{shchur2025fevbench}, and recommend using it alongside them: our axes are designed to cover several regimes of real-world data as proxies, and to test regimes that are only reachable with semi-synthetic data, such as novel interventions and events with several covariates.

\subsection{Verification}

For a task to be useful, each event in it must leave an effect on the target that is large enough to measure and clearly attributable to the covariates. We keep a covariate--target pair only if the covariate moves the target by a measurable amount and in a consistent direction, so the change can be attributed to the covariate rather than the target's own variation. Some relationships are not monotone over all factors (for example, SUMO flow is piecewise), but the factors we use reach only one branch, so each target moves the same way in every event.Concretely, we apply a $10\%$ criterion: every event must move the target at least $10\%$ away from its intervention-free baseline over the event window, measured on average across the window rather than at each timestep. This threshold keeps the effect above the level at which the metrics could measure it reliably. We check every event in a task, and drop the whole task if any single event fails the criterion. One alternative would be to retune the equations or factors to force the effect through, but we instead drop the failing tasks, since only a small fraction fall short: this leaves $37{,}665$ of the $40{,}000$ generated tasks. 

\subsection{Task layout}
\label{app:task-layout}

Within a task's history, the events do not fill the whole context. We lay out each history as an observational prefix, followed by the events, each separated from the next by a gap, and a small buffer at the end. The layout is:
\begin{itemize}
  \item \textbf{Observational prefix.} The first $40\%$ of the context holds no events: the covariates sit at their base setting and the target evolves on its own. This lets a model see the target's intrinsic dynamics, apart from any intervention, so it can separate them from the covariate-driven changes later.
  \item \textbf{Events.} Each event is exactly the event length $\ell$ long, and the events are placed one after another through the rest of the history.
  \item \textbf{Gaps.} Every event is followed by a gap of at least $32$ timesteps, so that one event never runs into the next and the target can settle back toward its baseline between events.
  \item \textbf{End buffer.} The last $2\%$ of the context is left clear, keeping the final event away from the forecast boundary.
\end{itemize}
The observational prefix, events, and gaps together fill the context, and the split depends on how many events a task has. At the most events a context and event length allow, the layout reduces to the $40\%$ prefix above; with fewer events, the extra room is added to the prefix, so the events stay anchored toward the recent end of the history. The exact prefix length is therefore set per task, from its context length, event length, and number of events. If the events and gaps do not fit in the nominal context, we extend the context instead of dropping events (\Cref{tab:ctx-pred-pairs}). This layout is a design choice specific to \bench{}, made mainly for clear benchmarking and evaluation; we keep these parameters fixed and instead vary the main axes described in \Cref{sec:benchmark-construction}.

\Cref{tab:ctx-pred-pairs} gives the nominal context and prediction lengths of the four pairs $C_1$--$C_4$ for each simulator. The three smallest pairs are shared; only the largest, $C_4$, varies slightly across simulators. For the longest events (length $512$), we extend the context so that the events fit (to $1{,}876$ at $C_1$, and to $7{,}504$ at $C_4$ on OpenDC).

\begin{table}[h]
\centering
\small
\begin{tabular}{lccc}
\toprule
 & \multicolumn{3}{c}{Context / prediction length} \\
\cmidrule(lr){2-4}
Pair & EPANET, GridLAB-D & OpenDC & SUMO \\
\midrule
$C_1$ & $1024$ / $128$ & $1024$ / $128$ & $1024$ / $128$ \\
$C_2$ & $2048$ / $256$ & $2048$ / $256$ & $2048$ / $256$ \\
$C_3$ & $4096$ / $384$ & $4096$ / $384$ & $4096$ / $384$ \\
$C_4$ & $8192$ / $512$ & $7328$ / $864$ & $7776$ / $864$ \\
\bottomrule
\end{tabular}
\caption{Nominal context and prediction lengths (in timesteps) of the four pairs $C_1$--$C_4$, per simulator.}
\label{tab:ctx-pred-pairs}
\end{table}

\subsection{Task generation procedure}

We generate the benchmark by repeating one procedure over every point of the axes grid. We iterate over each simulator, each of its targets, and each setting of the axes: the number of covariates $K$, the context/prediction pair, the event length, and the number of events (\Cref{tab:instantiation}). For a single such setting, we build a task as follows.
\begin{enumerate}
  \item \textbf{Take the baseline series.} Use the intervention-free target series, produced by running the simulator once on its base setting, and lay out the history with the observational prefix, events, and gaps described above.
  \item \textbf{Pick the covariates.} For a task with $K$ covariates, select $K$ of the target's covariates, including its primary one, to act together in each event.
  \item \textbf{Pick the factors.} For each event in the history, draw a multiplicative factor for each active covariate from a fixed pool of factors (given per simulator in \Cref{app:simulators}). Across the events, the factors take a spread of values, so the covariate--target relationship is seen at several points.
  \item \textbf{Apply the events.} At each event, scale the active covariates by their factors over the event window, following the assigned shape, and move the target by the amount the imposed covariate--target relationship prescribes. When several covariates are active, combine their effects using the rules in \Cref{app:simulators}.
  \item \textbf{Build the two futures.} Over the same history, form two future events. The Replay future reuses the covariate values of a past event; the Novel future pushes them beyond the range seen in the history. The Novel factor is set relative to the most extreme factor in the history: the largest one plus an offset drawn uniformly from a fixed range (or, for a covariate with an inverse relationship, the smallest one scaled by a factor drawn from a fixed range), so the extrapolation distance beyond the seen range does not depend on the number of events. This yields two tasks that share a history and differ only in the future.
  \item \textbf{Verify and keep.} Check the $10\%$ criterion on every event, and keep the two tasks only if all events pass; otherwise drop them.
\end{enumerate}
Repeating this over all targets and axis settings produces the full benchmark of $37{,}665$ verified tasks.
\paragraph{Generation parameters.}
For completeness, we give the values fixed in the generation code.
\begin{itemize}
  \item \textbf{Covariate selection.} Each simulator fixes an ordered list of covariates. A task with $K$ covariates uses the target's primary covariate followed by the first $K-1$ other covariates in that order, so the covariate set is deterministic given the target and $K$.
  \item \textbf{Bell shape.} Over an event window of length $\ell$ with timesteps $t=0,\dots,\ell-1$, a bell-shaped covariate with peak factor $f$ takes the factor $1+(f-1)\exp\!\big(-\tfrac{1}{2}\big((t-t_m)/\sigma\big)^2\big)$, with $t_m=(\ell-1)/2$ and $\sigma=\ell/6$. It rises from the base value to $f$ at the window midpoint and decays back.
  \item \textbf{History factors.} Each event draws a multiplicative factor per active covariate from a fixed pool: $\{0.1, 0.25, 0.5, 2, 4, 6, 8\}$ for OpenDC, GridLAB-D, and SUMO, and $\{2, 4, 6, 8\}$ for EPANET, whose covariates are only ever raised. Within an event, all active covariates push the target in one direction, chosen at random per event: a covariate with a direct relationship draws from the pool values above $1$ and one with an inverse relationship from the values below $1$, or the reverse for the opposite direction. This keeps the covariates' effects from cancelling each other within an event.
  \item \textbf{Novel factors.} For a covariate with a direct relationship, the Novel factor is the largest factor it took in the history plus an offset drawn uniformly from $[5,12]$. For an inverse relationship, it is the smallest history factor times a value drawn uniformly from $[0.1,0.5]$.
\end{itemize}

\subsection{Design choices: identifiability and evaluation}
\label{app:design}

\paragraph{Identifiability.}
We do not constrain the number or placement of events to guarantee that every covariate--target relationship is identifiable from a task's history. The evidence a task offers may be insufficient to learn them -- most clearly when many covariates are set but few events occur, but in other cases too, since it also depends on where each event acts on the target (for instance, an event applied at an already-high level can confound its effect). We treat the available evidence as a controlled axis: the number of historical events sets how much a model has to learn from, ranging from evidence-poor tasks (few events, many covariates) to evidence-rich ones. We report these regimes separately, and practitioners can filter to the high-event regime to focus on near-identifiable tasks.

Including tasks with limited historical evidence is deliberate and, we argue, principled, for three reasons. First, they are the norm in real scenario analysis, where a useful forecast is still expected under limited evidence. Second, our metric rewards the correct response to this: wQL is a proper scoring rule (\Cref{app:metrics}), so a forecaster that widens its predictive distribution under weak evidence is scored well, while an overconfident one is penalized -- especially on Novel futures.\footnote{This applies to the foundation models, which emit quantiles; the LLMs return only a median point forecast and so cannot express this uncertainty through their intervals, as noted in \Cref{app:experimental-setup}.} Third, the skill score measures improvement over a fixed baseline on the same task, so tasks that no forecaster can resolve do not distort the aggregate. The task also remains well-posed throughout: the simulator defines one true target trajectory, so limited identifiability reflects limited evidence, not an undefined target -- as in real forecasting. With enough events, every relationship can in principle be learned: it is exact, and it does not change direction over the factors we use. We leave designs that target identifiability directly -- for example, restricting to identifiable tasks to test identification of every relationship, or scoring calibrated abstention on tasks that are unidentifiable -- to future work.

\subsection{Simulator Details}
\label{app:simulators}

For each simulator we give the base setting we run it on, the variables we take as
targets and the properties we use as covariates, the shape assigned to each
covariate, and the covariate--target relationships we impose. Throughout, $y$
denotes the intervention-free baseline value of a target at a timestep, produced by
running the simulator, and $f$ denotes the multiplicative factor an event applies to
a covariate (so $f=1$ leaves the covariate at its base setting). A relationship maps
$(y,f)$ to the target value under the event.

The four simulators span distinct domains. \textbf{OpenDC} models a data center's resource usage for capacity planning -- for example, how CPU utilization and power draw respond as servers are added to a cluster. \textbf{GridLAB-D} models a power-distribution grid and the buildings on it, used to study how electricity demand responds to control -- for example, how a building's power use and indoor temperature respond when its thermostat set-point is raised. \textbf{SUMO} models road traffic vehicle by vehicle to evaluate traffic policies -- for example, how traffic flow responds when a road's speed limit is lowered. \textbf{EPANET} models pressurized water-distribution networks to study pump operation and demand growth -- for example, how network pressure responds to a surge in demand. All covariates are numerical, since the simulator properties they come from take numerical values.

\medskip\noindent
\colorbox{black!8}{\parbox{\dimexpr\linewidth-2\fboxsep\relax}{\small
\textbf{How the relationships are designed.} Every covariate--target relationship below follows the same rules:
\begin{itemize}[leftmargin=*, noitemsep, topsep=2pt]
  \item \textbf{Grounded in the simulator.} Each relationship is derived from the simulator's documented model of the system, and restated in closed form. These are the power and performance model of OpenDC, the residential load and power-flow model of GridLAB-D, the hydraulic model of EPANET, and the car-following, emission, and detector models of SUMO.
  \item \textbf{Direction from the system.} A covariate moves the target in the direction the real system would. For example, more cores lower the per-core utilization, and a higher clock speed raises the power draw.
  \item \textbf{Monotonic within the factors we use.} Each relationship is monotonic over the factors an event can draw, so the target always moves in one direction as the factor grows. Relationships that are piecewise in general, such as SUMO flow, stay on one branch.
  \item \textbf{Varied in form.} The forms include proportional, inverse, sublinear, faster-than-linear, exponential, and asymmetric relationships. A factor of two on a covariate need not mean a factor of two on the target.
  \item \textbf{Combined in a fixed way.} When an event sets several covariates at once, the secondary covariates act on the target either multiplicatively or additively, as given for each target.
  \item \textbf{Large enough to measure.} Every event moves the target by at least $10\%$ from its baseline, averaged over the event window. We drop any task in which an event does not.
\end{itemize}}}
\medskip

The rest of this section gives, for each simulator, the exact forms and constants, for reproducibility.

\subsubsection{OpenDC}
\label{app:sim-opendc}

\paragraph{Base setting.}
We run OpenDC on a single rack of four identical hosts (\texttt{host-0}--\texttt{host-3}),
each with $16$ cores at a core speed of $3000$ and $64\,$GB of memory, driven by the
Bitbrains workload trace of four virtual machines.\footnote{\url{https://github.com/atlarge-research/quantifying-api-design/tree/master/traces/bitbrains-small}}
Running it exports the resource usage of the hosts and virtual machines at a fixed
interval, which forms the baseline series.

\paragraph{Targets and covariates.}
We take $22$ target series and use six covariates, named as they appear in the
released benchmark:
\begin{itemize}
  \item \textbf{Host targets ($10$).} \texttt{cpu\_demand\_host-0},
  \texttt{cpu\_usage\_host-0}, \texttt{cpu\_utilization\_host-0},
  \texttt{cpu\_time\_active\_host-0}, \texttt{power\_draw\_host-0}, and the same five
  for \texttt{host-1}.
  \item \textbf{Virtual machine targets ($12$).} For each of the VMs
  \texttt{1019}, \texttt{1023} (running on \texttt{host-0}) and \texttt{1026}
  (running on \texttt{host-1}): \texttt{cpu\_demand\_<vm>},
  \texttt{cpu\_usage\_<vm>}, \texttt{cpu\_time\_active\_<vm>}, and
  \texttt{cpu\_time\_idle\_<vm>}, e.g.\ \texttt{cpu\_demand\_1019}.
  \item \textbf{Covariates ($6$).} \texttt{coreCount\_host-0},
  \texttt{coreSpeed\_host-0}, \texttt{memorySize\_host-0}, and the same three for
  \texttt{host-1}.
\end{itemize}
Each target has one \emph{primary} covariate that drives it most directly; the
pairing is given in \Cref{tab:opendc-primary}. The other covariates can also act on a
target as \emph{secondary} drivers when an event sets several covariates at once
(described below).

\paragraph{Covariate shapes.}
The \texttt{coreSpeed} covariates (\texttt{coreSpeed\_host-0},
\texttt{coreSpeed\_host-1}) are assigned the bell shape (clock speed could be ramped
up for a period and then brought back down). The \texttt{coreCount} and
\texttt{memorySize} covariates are assigned the step shape (a hardware
allocation is set to a level and held there for the duration of the event).
Each event draws its multiplicative factor for a covariate from the pool
$\{0.1, 0.25, 0.5, 2, 4, 6, 8\}$.

\begin{table}[t]
\centering
\small
\begin{tabular}{lll}
\toprule
Target & Primary covariate & Combination \\
\midrule
\texttt{cpu\_demand\_host-0}       & \texttt{coreSpeed\_host-0}  & multiplicative \\
\texttt{cpu\_usage\_host-0}        & \texttt{coreSpeed\_host-0}  & multiplicative \\
\texttt{power\_draw\_host-0}       & \texttt{coreSpeed\_host-0}  & additive \\
\texttt{cpu\_utilization\_host-0}  & \texttt{coreCount\_host-0}  & additive \\
\texttt{cpu\_time\_active\_host-0} & \texttt{coreCount\_host-0}  & multiplicative \\
\texttt{cpu\_demand\_1019}         & \texttt{coreSpeed\_host-0}  & multiplicative \\
\texttt{cpu\_usage\_1019}          & \texttt{coreSpeed\_host-0}  & multiplicative \\
\texttt{cpu\_time\_active\_1019}   & \texttt{coreCount\_host-0}  & multiplicative \\
\texttt{cpu\_time\_idle\_1019}     & \texttt{memorySize\_host-0} & multiplicative \\
\texttt{cpu\_demand\_1023}         & \texttt{coreSpeed\_host-0}  & multiplicative \\
\texttt{cpu\_usage\_1023}          & \texttt{coreSpeed\_host-0}  & multiplicative \\
\texttt{cpu\_time\_active\_1023}   & \texttt{coreCount\_host-0}  & multiplicative \\
\texttt{cpu\_time\_idle\_1023}     & \texttt{memorySize\_host-0} & multiplicative \\
\bottomrule
\end{tabular}
\caption{\textbf{OpenDC target--covariate pairing (\texttt{host-0} side).} Each target
is driven by one primary covariate; whether secondary covariates combine
multiplicatively or additively is set by the target's class. The \texttt{host-1}
targets and VM \texttt{1026} pair with the \texttt{host-1} covariates in the same
way.}
\label{tab:opendc-primary}
\end{table}

\paragraph{Primary relationships.}
Each primary relationship maps the baseline value $y$ and the primary covariate's
factor $f$ to the target under the event. The relationships below are derived from
OpenDC's documented power and performance model~\citep{mastenbroek2021opendc};\footnote{\url{https://opendc.org}}
we restate them in closed form and give a short description of the behavior each one
captures. We give the forms for \texttt{host-0} below; \texttt{host-1} uses the same
forms with slightly different constants, and the virtual machine targets use the same
forms with per-machine constants.
\begin{itemize}
  \item \textbf{\texttt{cpu\_demand\_host-0}, \texttt{cpu\_usage\_host-0} (from \texttt{coreSpeed\_host-0}).}
  Throughput scales close to linearly with clock frequency, with a small fixed
  overhead:
  \begin{align*}
    \texttt{cpu\_demand\_host-0}&: y' = f\,y + 0.05\,(f-1), \\
    \texttt{cpu\_usage\_host-0}&: y' = (0.98 f + 0.02)\,y + 0.15\,(f-1).
  \end{align*}
  \item \textbf{\texttt{power\_draw\_host-0} (from \texttt{coreSpeed\_host-0}).} A host
  draws a fixed idle power (about $105.5\,$W) plus a dynamic part that scales with
  frequency, with a small switching term:
  \begin{equation*}
    y' = 105.5 + f\,(y - 105.5) + 2.2\,\lvert f-1\rvert.
  \end{equation*}
  \item \textbf{\texttt{cpu\_utilization\_host-0} (from \texttt{coreCount\_host-0}).}
  The same work spread over more cores lowers per-core utilization inversely:
  \begin{equation*}
    y' = \tfrac{1}{f}\,y.
  \end{equation*}
  \item \textbf{\texttt{cpu\_time\_active\_host-0} (from \texttt{coreCount\_host-0}).}
  Adding cores raises the total active core-time sublinearly, reflecting imperfect
  parallel speedup:
  \begin{equation*}
    y' = \big(1 + 0.4\,(f-1)\big)\,y.
  \end{equation*}
  \item \textbf{\texttt{cpu\_time\_idle\_1019} (from \texttt{memorySize\_host-0}).}
  Adding memory slightly reduces idle waiting; shrinking memory raises it sharply as
  the machine begins to swap pages. The relationship is therefore asymmetric in $f$:
  \begin{equation*}
    y' =
    \begin{cases}
      \big(0.28 + \tfrac{0.72}{f}\big)\,y, & f \ge 1, \\[4pt]
      \big(1 + 3.2\,(1-f)\big)\,y + 12\,(1-f), & f < 1.
    \end{cases}
  \end{equation*}
\end{itemize}

\begin{table}[t]
\centering
\small
\setlength{\tabcolsep}{5pt}
\resizebox{\textwidth}{!}{%
\begin{tabular}{llll}
\toprule
Target class & Secondary covariate & Weight $w_c$ & Effect ($r$ or $o$) \\
\midrule
\multirow{5}{*}{CPU demand / usage (mult.)}
  & core count, same host   & $0.60$ & $r = f^2$ \\
  & memory size, same host ($f<1$)  & $0.40$ & $r = 10/f$ \\
  & core speed, other host ($f<1$)  & $0.35$ & $r = 5\,(2-f)$ \\
  & core count, other host  & $0.30$ & $r = (2-f)^3$ \\
  & memory size, other host & $0.20$ & $r = 1 + 3/f$ \\
\midrule
\multirow{5}{*}{Active CPU time (mult.)}
  & core speed, same host   & $0.80$ & $r = 1/f^3$ \\
  & memory size, same host ($f<1$)  & $0.50$ & $r = 500/f$ \\
  & core speed, other host ($f<1$)  & $0.35$ & $r = 400\,(2-f)$ \\
  & core count, other host  & $0.30$ & $r = 300\,(2-f)$ \\
  & memory size, other host & $0.20$ & $r = 200\,(2-f)$ \\
\midrule
\multirow{5}{*}{Idle CPU time (mult.)}
  & core speed, same host   & $0.60$ & $r = f^2$ if $f_m\ge1$, else $1/f^2$ \\
  & core count, same host   & $0.50$ & $r = 1/f^3$ \\
  & core speed, other host ($f<1$)  & $0.30$ & $r = f$ \\
  & core count, other host ($f<1$)  & $0.20$ & $r = f$ \\
  & memory size, other host ($f<1$) & $0.20$ & $r = f$ \\
\midrule
\multirow{5}{*}{Power draw (add.)}
  & core count, same host   & $0.70$ & $o = 50\,(f-1)$ if $f\ge1$, else $-50\,(1-f)$ \\
  & memory size, same host  & $0.50$ & $o = 150\,(f-1)$ if $f\ge1$, else $300\,(1-f)$ \\
  & core speed, other host ($f<1$)  & $0.35$ & $o = 250\,(1-f)$ \\
  & core count, other host  & $0.30$ & $o = 200\,(1-f)$ \\
  & memory size, other host & $0.20$ & $o = 100\,(1-f)$ \\
\midrule
\multirow{5}{*}{CPU utilization (add.)}
  & core speed, same host   & $0.70$ & $o = -0.05\,(f-1)$ if $f\ge1$, else $0.15\,(1-f)$ \\
  & memory size, same host ($f<1$)  & $0.55$ & $o = 2\,(1-f)$ \\
  & core speed, other host ($f<1$)  & $0.35$ & $o = 3\,(1-f)$ \\
  & core count, other host  & $0.30$ & $o = 2\,(1-f)$ \\
  & memory size, other host & $0.20$ & $o = 1\,(1-f)$ \\
\bottomrule
\end{tabular}%
}
\caption{\textbf{OpenDC secondary covariate effects (\texttt{host-0} targets).} For a
target, each secondary covariate contributes an effect scaled by its weight $w_c$ and
the joint dampening $s$. Multiplicative classes use the raw effect $r$; additive
classes use the offset $o$. Here $f$ is the secondary covariate's own factor, and
$f_m$ the memory-size factor. Effects marked ($f<1$) apply only when that covariate
is reduced. \texttt{host-1} targets use the same template with mirrored hosts.}
\label{tab:opendc-secondary}
\end{table}

\paragraph{Combining multiple covariates.}
When an event sets more than one covariate, the target starts from its primary
relationship and each additional active covariate applies a further, dampened effect.
A target's class fixes whether these effects combine multiplicatively
(\texttt{cpu\_demand}, \texttt{cpu\_usage}, \texttt{cpu\_time\_active},
\texttt{cpu\_time\_idle}) or additively (\texttt{power\_draw},
\texttt{cpu\_utilization}).
Let $n$ be the number of active secondary covariates in the event and set the joint
dampening
\begin{equation*}
  s =
  \begin{cases}
    1, & n \le 1, \\
    \tfrac{1}{\sqrt{n}}, & n > 1,
  \end{cases}
\end{equation*}
so that the more covariates move together, the less each one contributes. Each
secondary covariate $c$ also carries a fixed weight $w_c$ that orders it by
importance. For a multiplicative target with a secondary covariate whose own effect
is $r$, and an additive target with a secondary offset $o$, the running target value
is updated as
\begin{equation*}
  \text{multiplicative: } y \leftarrow y\,\big[\,1 + (s\,w_c)\,(r-1)\,\big],
  \qquad
  \text{additive: } y \leftarrow y + (s\,w_c)\,o .
\end{equation*}
The weights $w_c$ and the effects $r$ and $o$ are listed in
\Cref{tab:opendc-secondary}. Two patterns are worth noting. First, a covariate on the
\emph{same} host as the target acts in the opposite direction to the \emph{other}
host's covariate: reducing a remote host's resources pushes load onto the local host
and raises its demand, active time, and power, reflecting consolidation and
contention across the shared rack. Second, some effects only apply in one direction.
For example, memory acts only when it shrinks (when $f<1$, as swapping begins), and a
remote host's speed acts only when it is reduced. The final target value is clipped at
zero.

The direction and qualitative form of every effect follow the mechanisms above. The
magnitudes (the weights $w_c$ and the constants in $r$ and $o$) are chosen so that each
covariate leaves a clear, separable imprint on the target, rather than being derived
from the simulator.

\subsubsection{GridLAB-D}
\label{app:sim-gridlab}

\paragraph{Base setting.}
We run GridLAB-D 5.3.0 with its \texttt{climate}, \texttt{powerflow},
\texttt{residential}, and \texttt{tape} modules, driven by Seattle TMY3 weather data.
The topology is a single-house feeder: a \texttt{SWING} source node at $2400\,$V feeds
a $75\,$kVA center-tapped transformer, then a $120\,$V triplex node, then a single
house ($1550\,$ft$^2$, heating set-point $60\,^\circ$F, cooling set-point
$77\,^\circ$F, thermostat deadband $1.6\,^\circ$F) with a constant $10\,$kW ZIP load
(impedance/current/power fractions $0.3/0.5/0.2$). Outputs are recorded every
$60\,$seconds, which forms the baseline series.

\paragraph{Targets and covariates.}
We generate tasks for the $11$ target series below and use $21$ covariates, named as
they appear in the released benchmark. One target, \texttt{is\_COOL\_on}, fails the
$10\%$ verification criterion in every task and is dropped entirely, leaving $10$
GridLAB-D targets in the released benchmark:
\begin{itemize}
  \item \textbf{Power targets ($5$).} \texttt{actual\_power}, \texttt{total\_load},
  \texttt{hvac\_load}, \texttt{hvac\_power}, \texttt{fan\_power}.
  \item \textbf{Temperature targets ($2$).} \texttt{air\_temperature},
  \texttt{mass\_temperature}.
  \item \textbf{Demand target ($1$).} \texttt{heating\_demand}.
  \item \textbf{State targets ($3$).} \texttt{is\_AUX\_on}, \texttt{is\_HEAT\_on},
  \texttt{is\_COOL\_on}. These are on/off indicators; each is stored offset by $1$ and
  kept in the window $[1,2]$.
  \item \textbf{Covariates ($21$).} \texttt{base\_power}, \texttt{current\_pf},
  \texttt{power\_pf}, \texttt{impedance\_pf}, \texttt{duty\_cycle}, \texttt{period},
  \texttt{phase}, \texttt{heat\_fraction}, \texttt{multiplier},
  \texttt{voltage\_factor}, \texttt{breaker\_amps}, \texttt{power\_factor},
  \texttt{number\_of\_devices}, \texttt{thermostatic\_control\_range},
  \texttt{rate\_of\_cooling}, \texttt{rate\_of\_heating}, \texttt{temperature},
  \texttt{phi}, \texttt{demand\_rate}, \texttt{nominal\_power},
  \texttt{recovery\_duty\_cycle}.
\end{itemize}
Each target has one \emph{primary} covariate that drives it most directly; the pairing
is given in \Cref{tab:gridlab-primary}. Eleven of the covariates act as a primary
driver; the remaining ten only act as \emph{secondary} drivers when an event sets
several covariates at once (described below).

\paragraph{Covariate shapes.}
The covariates \texttt{temperature}, \texttt{base\_power}, \texttt{multiplier},
\texttt{voltage\_factor}, \texttt{duty\_cycle}, and \texttt{recovery\_duty\_cycle} are
assigned the bell shape (a setting is ramped up for a period and then brought back
down). The remaining covariates are assigned the step shape (a setting is moved to a
level and held there for the duration of the event).
Each event draws its multiplicative factor for a covariate from the pool
$\{0.1, 0.25, 0.5, 2, 4, 6, 8\}$.

\begin{table}[t]
\centering
\small
\begin{tabular}{lll}
\toprule
Target & Primary covariate & Combination \\
\midrule
\texttt{actual\_power}    & \texttt{base\_power}                 & multiplicative \\
\texttt{total\_load}      & \texttt{nominal\_power}              & multiplicative \\
\texttt{fan\_power}       & \texttt{power\_pf}                   & multiplicative \\
\texttt{hvac\_load}       & \texttt{voltage\_factor}            & multiplicative \\
\texttt{hvac\_power}      & \texttt{duty\_cycle}                & multiplicative \\
\texttt{heating\_demand}  & \texttt{rate\_of\_heating}          & multiplicative \\
\texttt{air\_temperature} & \texttt{heat\_fraction}             & additive \\
\texttt{mass\_temperature}& \texttt{temperature}                & additive \\
\texttt{is\_COOL\_on}     & \texttt{thermostatic\_control\_range}& additive \\
\texttt{is\_HEAT\_on}     & \texttt{recovery\_duty\_cycle}      & additive \\
\texttt{is\_AUX\_on}      & \texttt{period}                     & additive \\
\bottomrule
\end{tabular}
\caption{\textbf{GridLAB-D target--covariate pairing.} Each target is driven by one
primary covariate; whether secondary covariates combine multiplicatively or additively
is set by the target's class.}
\label{tab:gridlab-primary}
\end{table}

\paragraph{Primary relationships.}
Each primary relationship maps the baseline value $y$ and the primary covariate's
factor $f$ to the target under the event. The relationships below are derived from
GridLAB-D's documented residential and power-flow model~\citep{chassin2008gridlabd};\footnote{\url{https://www.gridlabd.org/}}
we restate them in closed form and give a short description of the behavior each one
captures. For brevity we write $g(f) = \operatorname{sign}(f-1)\,\lvert f-1\rvert^{0.4}$
for the damped shift used by several relationships.
\begin{itemize}
  \item \textbf{\texttt{actual\_power}, \texttt{total\_load}, \texttt{fan\_power}
  (from \texttt{base\_power}, \texttt{nominal\_power}, \texttt{power\_pf}).} Each power
  draw scales close to linearly with its driving quantity, with a small fixed
  overhead. All three take the same form, e.g.:
  \begin{equation*}
    \texttt{actual\_power}: y' = f\,y + 0.05\,(f-1).
  \end{equation*}
  \item \textbf{\texttt{hvac\_load}, \texttt{hvac\_power} (from \texttt{voltage\_factor},
  \texttt{duty\_cycle}).} Raising the driver lowers the HVAC draw by a damped offset:
  \begin{align*}
    \texttt{hvac\_load}&: y' = y - 0.4\,g(f) - 0.02\,(f-1), \\
    \texttt{hvac\_power}&: y' = y - 0.35\,g(f) - 0.02\,(f-1).
  \end{align*}
  \item \textbf{\texttt{heating\_demand} (from \texttt{rate\_of\_heating}).} Heating
  demand rises super-linearly when the heating rate goes up and falls sub-linearly
  when it goes down, so the relationship is asymmetric in $f$:
  \begin{equation*}
    y' =
    \begin{cases}
      f^{1.5}\,y, & f \ge 1, \\[2pt]
      f^{0.7}\,y, & f < 1.
    \end{cases}
  \end{equation*}
  \item \textbf{\texttt{air\_temperature}, \texttt{mass\_temperature} (from
  \texttt{heat\_fraction}, \texttt{temperature}).} Each temperature shifts by a damped
  amount plus a small linear term:
  \begin{align*}
    \texttt{air\_temperature}&: y' = y + 2.0\,g(f) + 0.05\,(f-1), \\
    \texttt{mass\_temperature}&: y' = y + 1.5\,g(f) + 0.04\,(f-1).
  \end{align*}
  \item \textbf{\texttt{is\_COOL\_on}, \texttt{is\_HEAT\_on}, \texttt{is\_AUX\_on}
  (from \texttt{thermostatic\_control\_range}, \texttt{recovery\_duty\_cycle},
  \texttt{period}).} Each on/off indicator shifts within its clipped window $[1,2]$:
  \begin{align*}
    \texttt{is\_COOL\_on}&: y' = \operatorname{clip}\!\big(y + 0.25\,g(f) + 0.01\,(f-1),\,1,\,2\big), \\
    \texttt{is\_HEAT\_on}&: y' = \operatorname{clip}\!\big(y - 0.25\,g(f) - 0.01\,(f-1),\,1,\,2\big), \\
    \texttt{is\_AUX\_on} &: y' = \operatorname{clip}\!\big(y - 0.20\,g(f) - 0.01\,(f-1),\,1,\,2\big).
  \end{align*}
\end{itemize}

\begin{table}[t]
\centering
\small
\setlength{\tabcolsep}{5pt}
\resizebox{\textwidth}{!}{%
\begin{tabular}{llll}
\toprule
Target class & Secondary covariate group & Weight $w_c$ & Effect ($r$ or $\delta$) \\
\midrule
\multirow{5}{*}{Power (mult.)}
  & \texttt{voltage\_factor}         & $0.50$ & $r = f^2$ \\
  & \texttt{breaker\_amps}           & $0.40$ & $r = 1 + 0.1\,(f-1)$ \\
  & \texttt{phi}                     & $0.30$ & $r = \cos^2\!\big(\operatorname{clip}(1.57\,(f-1),\pm1.57)\big)$ \\
  & other non-thermal               & $0.20$ & $r = f + 0.05\,(f-1)$ \\
  & thermal                         & $0.30$ & $r = 1 + 0.15\,g(f)$ \\
\midrule
\multirow{2}{*}{Demand (mult.)}
  & thermal                         & $0.40$ & $r = 1 + 0.15\,g(f)$ \\
  & non-thermal                     & $0.20$ & $r = 1 + 0.05\,g(f)$ \\
\midrule
\multirow{5}{*}{Temperature (add.)}
  & \texttt{heat\_fraction}          & $1.50$ & $\delta = w_c\,g(f) + 0.02\,(f-1)$ \\
  & \texttt{temperature}             & $2.50$ & $\delta = w_c\,g(f) + 0.02\,(f-1)$ \\
  & \texttt{thermostatic\_control\_range} & $0.50$ & $\delta = w_c\,g(f) + 0.02\,(f-1)$ \\
  & other thermal                   & $0.40$ & $\delta = w_c\,g(f) + 0.02\,(f-1)$ \\
  & non-thermal                     & $0.05$ & $\delta = w_c\,g(f) + 0.005\,(f-1)$ \\
\midrule
\multirow{2}{*}{State (add.)}
  & thermal                         & $0.40$ & $\delta = -0.1\,g(f)$ \\
  & non-thermal                     & $0.10$ & $\delta = 0.02\,g(f)$ \\
\bottomrule
\end{tabular}%
}
\caption{\textbf{GridLAB-D secondary covariate effects.} For a target, each secondary
covariate contributes an effect scaled by its weight $w_c$ and the joint dampening
$s$. Multiplicative classes use the raw effect $r$; additive classes use the offset
$\delta$. Here $f$ is the secondary covariate's own factor and
$g(f)=\operatorname{sign}(f-1)\,\lvert f-1\rvert^{0.4}$. Covariates split into a
\emph{thermal} group (\texttt{heat\_fraction}, \texttt{temperature},
\texttt{thermostatic\_control\_range}, \texttt{rate\_of\_heating},
\texttt{rate\_of\_cooling}, \texttt{duty\_cycle}, \texttt{recovery\_duty\_cycle},
\texttt{period}), which act on the building's thermal state, and a \emph{non-thermal}
group of all remaining covariates, which act on the electrical load.}
\label{tab:gridlab-secondary}
\end{table}

\paragraph{Combining multiple covariates.}
When an event sets more than one covariate, the target starts from its primary
relationship and each additional active covariate applies a further, dampened effect.
A target's class fixes whether these effects combine multiplicatively (\texttt{power},
\texttt{demand}) or additively (\texttt{temperature}, \texttt{state}). The covariates
also split into a \emph{thermal} group, which acts on the building's thermal state, and
a \emph{non-thermal} group, which acts on the electrical load; the two groups get
different weights and effect shapes, listed in \Cref{tab:gridlab-secondary}.
Let $n$ be the number of active secondary covariates in the event and set the joint
dampening
\begin{equation*}
  s =
  \begin{cases}
    1, & n \le 1, \\
    \tfrac{1}{\sqrt{n}}, & n > 1,
  \end{cases}
\end{equation*}
so that the more covariates move together, the less each one contributes. Each
secondary covariate $c$ also carries a fixed weight $w_c$ that orders it by
importance. Writing $r$ for a multiplicative effect and $\delta$ for an additive one,
the running target value is updated as
\begin{align*}
  \text{power, demand: } &y \leftarrow y\,\big[\,1 + (s\,w_c)\,(r-1)\,\big], \\
  \text{temperature: }   &y \leftarrow y + (s\cdot0.5)\,\delta, \\
  \text{state: }         &y \leftarrow y + (s\,w_c)\,\delta .
\end{align*}
The weights $w_c$ and the effects $r$ and $\delta$ are listed in
\Cref{tab:gridlab-secondary}. After all effects are applied, the state targets are
clipped back to their $[1,2]$ window.

The direction and qualitative form of every effect follow the mechanisms above. The
magnitudes, i.e.\ the weights $w_c$ and the constants in $r$ and $\delta$, are chosen
so that each covariate leaves a clear and separable imprint on the target, rather than
being derived from the simulator.

\subsubsection{EPANET}
\label{app:sim-epanet}

\paragraph{Base setting.}
We run EPANET~2.2 through WNTR on the stock \texttt{Net1} network (EPANET Example~1).
The network has $11$ nodes ($9$ junctions, one reservoir with id $9$, and one tank with
id $2$) and $13$ links ($12$ pipes and one pump with id $9$), solved with the
Hazen--Williams head-loss model in GPM units. A control opens the pump when the tank
level drops below a lower bound and closes it above an upper bound, so the tank cycles
over the run. Outputs are recorded at a fixed interval, which forms the baseline
series.

\paragraph{Targets and covariates.}
We take $37$ target series and use the covariates below, named as they appear in the
released benchmark:
\begin{itemize}
  \item \textbf{Pressure targets ($10$).} \texttt{pressure\_\{10,11,12,13,21,22,23,31,32\}}
  at the nine junctions, and \texttt{pressure\_2} at the tank.
  \item \textbf{Demand targets ($2$).} \texttt{demand\_9} (reservoir outflow) and
  \texttt{demand\_2} (tank flow).
  \item \textbf{Flowrate targets ($13$).} \texttt{flowrate\_p} for the twelve pipes
  $p \in \{10,11,12,21,22,31,110,111,112,113,121,122\}$, and \texttt{flowrate\_9} at
  the pump.
  \item \textbf{Velocity targets ($12$).} \texttt{velocity\_p} for the same twelve
  pipes.
  \item \textbf{Covariates.} \texttt{junction\_j\_base\_demand} for the nine junctions,
  \texttt{reservoir\_9\_base\_head}, \texttt{tank\_2\_init\_level},
  \texttt{pump\_9\_setting}, and \texttt{pipe\_p\_diameter} and
  \texttt{pipe\_p\_roughness} for the twelve pipes.
\end{itemize}
Each target has one \emph{primary} covariate that drives it most directly; the pairing
is given in \Cref{tab:epanet-primary}. Any covariate that an event also sets acts as a
\emph{secondary} driver (described below). Every event only \emph{raises} its
covariates (the factor $f>1$), so targets only increase under an event.

\paragraph{Covariate shapes.}
The covariates \texttt{pump\_9\_setting}, \texttt{reservoir\_9\_base\_head}, and
\texttt{tank\_2\_init\_level} are assigned the bell shape (a setting is ramped up for a
period and then brought back down). The remaining covariates are assigned the step
shape (a setting is moved to a level and held there for the duration of the event).
Each event draws its multiplicative factor for a covariate from the pool
$\{2, 4, 6, 8\}$; all factors exceed $1$, so the targets only increase under an event.

\begin{table}[t]
\centering
\small
\resizebox{\textwidth}{!}{%
\begin{tabular}{lllll}
\toprule
Target & Primary covariate & $a$ & $c$ & Combination \\
\midrule
\texttt{pressure\_j} (junctions)   & \texttt{junction\_j\_base\_demand} & $0.09$ & $1 + 0.05\,(j \bmod 10)$ & multiplicative \\
\texttt{pressure\_2} (tank)        & \texttt{tank\_2\_init\_level}      & $0.09$ & $1$                       & multiplicative \\
\texttt{demand\_9} (reservoir)     & \texttt{reservoir\_9\_base\_head}  & $0.09$ & $1$                       & multiplicative \\
\texttt{demand\_2} (tank)          & \texttt{tank\_2\_init\_level}      & $0.09$ & $1$                       & multiplicative \\
\texttt{flowrate\_p} (pipes)       & \texttt{pipe\_p\_diameter}         & $0.09$ & $1 + 0.1\,(p \bmod 10)$   & multiplicative \\
\texttt{flowrate\_9} (pump)        & \texttt{pump\_9\_setting}          & $0.09$ & $1$                       & multiplicative \\
\texttt{velocity\_p} (pipes)       & \texttt{pipe\_p\_roughness}        & $0.10$ & $1 + 0.02\,(p \bmod 10)$  & multiplicative \\
\bottomrule
\end{tabular}%
}
\caption{\textbf{EPANET target--covariate pairing.} Each target is driven by one
primary covariate through the shared form in the text, with per-element constants $a$
and $c$. All targets combine multiplicatively. Here $j$ ranges over the nine junction
ids and $p$ over the twelve pipe ids.}
\label{tab:epanet-primary}
\end{table}

\paragraph{Primary relationships.}
Each primary relationship maps the baseline value $y$ and the primary covariate's
factor $f$ to the target under the event. The relationships are derived from EPANET's
documented hydraulic model~\citep{rossman2000epanet};\footnote{\url{https://www.epa.gov/water-research/epanet}}
we restate them in closed form. Every target follows the same multiplicative form, with
a small per-element sensitivity:
\begin{equation*}
  y' = y\,\big[\,1 + a\,c\,g(f) + 0.01\,c\,(f-1)\,\big],
  \qquad
  g(f) = \operatorname{sign}(f-1)\,\lvert f-1\rvert^{0.4},
\end{equation*}
where the constants $a$ and $c$ for each target are listed in
\Cref{tab:epanet-primary}. The target rises with its driver by a damped amount; the
constant $c$ scales how sensitive each junction or pipe is, and because events only
raise covariates ($f>1$) the target always increases.

\begin{table}[t]
\centering
\small
\begin{tabular}{lll}
\toprule
Target class & Weight $w_c$ & Ratio $r$ \\
\midrule
Pressure & $0.10 + 0.05\,(h \bmod 5)$ & \multirow{4}{*}{$r = 1 + (s\,w_c)\,(f^{0.4}-1) + 0.01\,(s\,w_c)\,(f-1)$} \\
Demand   & $0.10 + 0.02\,(h \bmod 5)$ & \\
Flowrate & $0.10 + 0.04\,(h \bmod 6)$ & \\
Velocity & $0.10 + 0.03\,(h \bmod 6)$ & \\
\bottomrule
\end{tabular}
\caption{\textbf{EPANET secondary covariate effects.} Each secondary covariate
contributes a multiplicative ratio $r$ scaled by its weight $w_c$ and the joint
dampening $s$. All classes share the same ratio shape and differ only in the weight.
Here $f$ is the secondary covariate's own factor, and $h$ is a fixed per-covariate
constant (the sum of the character codes of its name) that spreads the weights across
covariates.}
\label{tab:epanet-secondary}
\end{table}

\paragraph{Combining multiple covariates.}
When an event sets more than one covariate, the target starts from its primary
relationship and each additional active covariate applies a further, dampened effect.
For EPANET every class combines multiplicatively.
Let $n$ be the number of active secondary covariates in the event and set the joint
dampening
\begin{equation*}
  s =
  \begin{cases}
    1, & n \le 1, \\
    \tfrac{1}{\sqrt{n}}, & n > 1,
  \end{cases}
\end{equation*}
so that the more covariates move together, the less each one contributes. Each
secondary covariate $c$ also carries a fixed weight $w_c$ that orders it by importance.
For a secondary covariate whose ratio is $r$, the running target value is updated as
\begin{equation*}
  y \leftarrow y\,\cdot\,r,
  \qquad
  r = 1 + (s\,w_c)\,(f^{0.4}-1) + 0.01\,(s\,w_c)\,(f-1).
\end{equation*}
The weights $w_c$ are listed in \Cref{tab:epanet-secondary}.

The direction and qualitative form of every effect follow the mechanisms above. The
magnitudes, i.e.\ the weights $w_c$ and the constants in the primary and secondary
relationships, are chosen so that each covariate leaves a clear and separable imprint
on the target, rather than being derived from the simulator.

\subsubsection{SUMO}
\label{app:sim-sumo}

\paragraph{Base setting.}
We run SUMO on a single signalized four-arm intersection: a center node with four
approach nodes at $\pm500\,$m, every edge two lanes at $13.89\,$m/s ($50\,$km/h) in
both directions, and a fixed traffic-light plan on a $90\,$s cycle
($42\,$s green, $3\,$s yellow, twice). Traffic follows a rush-hour demand of four
through-movements, with five periods per day (from $120$ vehicles/hour at night up to
$2000$ vehicles/hour at peak) repeated over $30$ days. Induction-loop and lane-area
detectors, plus an edge-level emissions record, are written every $300\,$s, which forms
the baseline series.

\paragraph{Targets and covariates.}
We generate tasks for $30$ target series and use $16$ covariates. One target fails
the $10\%$ verification criterion in every task and is dropped entirely, leaving
the $29$ targets below, named as they appear in the released benchmark. Detectors on the two measured edges carry the suffixes
\texttt{\_e\_0} and \texttt{\_e\_1}; we write \texttt{\{0,1\}} for the pair.
\begin{itemize}
  \item \textbf{Emissions targets ($4$).} \texttt{CO2\_abs\_edge\_e2c},
  \texttt{NOx\_abs\_edge\_e2c}, \texttt{PMx\_abs\_edge\_e2c},
  \texttt{fuel\_abs\_edge\_e2c}.
  \item \textbf{Flow targets ($5$).} \texttt{flow\_loop\_e\_\{0,1\}\_near},
  \texttt{flow\_loop\_e\_0\_far}, \texttt{flow\_loop\_exit\_e\_\{0,1\}}.
  \item \textbf{Occupancy targets ($6$).} \texttt{occupancy\_loop\_e\_\{0,1\}\_near},
  \texttt{occupancy\_loop\_exit\_e\_\{0,1\}}, \texttt{meanOccupancy\_e2\_e\_\{0,1\}}.
  \item \textbf{Speed targets ($6$).} \texttt{speed\_loop\_e\_\{0,1\}\_near},
  \texttt{speed\_loop\_exit\_e\_\{0,1\}}, \texttt{meanSpeed\_e2\_e\_\{0,1\}}.
  \item \textbf{Delay targets ($6$).} \texttt{meanHaltingDuration\_e2\_e\_\{0,1\}},
  \texttt{meanTimeLoss\_e2\_e\_\{0,1\}}, \texttt{meanVehicleNumber\_e2\_e\_\{0,1\}}.
  \item \textbf{Jam-length targets ($2$).}
  \texttt{meanMaxJamLengthInMeters\_e2\_e\_\{0,1\}}.
  \item \textbf{Covariates ($16$).} \texttt{minGap}, \texttt{accel}, \texttt{decel},
  \texttt{maxSpeed}, \texttt{sigma}, \texttt{tau}, \texttt{length}, \texttt{speedDev},
  \texttt{offset}, \texttt{width}, \texttt{height}, \texttt{lcStrategic},
  \texttt{lcCooperative}, \texttt{lcSpeedGain}, \texttt{impatience},
  \texttt{actionStepLength}.
\end{itemize}
Each target has one \emph{primary} covariate; the pairing is given in
\Cref{tab:sumo-primary}. Fifteen covariates act as a primary driver;
\texttt{speedDev} has no target in the released set and only acts as a
\emph{secondary} driver (described below), as do the other covariates when an event
sets several at once. Before the relationships are applied, every target series is
shifted up by a constant so all values stay positive.

\paragraph{Covariate shapes.}
The covariates \texttt{sigma}, \texttt{impatience}, and \texttt{lcSpeedGain} are
assigned the bell shape (a setting is ramped up for a period and then brought back
down). The remaining covariates are assigned the step shape (a setting is moved to a
level and held there for the duration of the event).
Each event draws its multiplicative factor for a covariate from the pool
$\{0.1, 0.25, 0.5, 2, 4, 6, 8\}$.

\begin{table}[t]
\centering
\small
\begin{tabular}{lll}
\toprule
Target & Primary covariate & Combination \\
\midrule
\texttt{meanMaxJamLengthInMeters\_e2\_e\_\{0,1\}} & \texttt{minGap}     & additive \\
\texttt{meanTimeLoss\_e2\_e\_\{0,1\}}            & \texttt{accel}       & additive \\
\texttt{meanHaltingDuration\_e2\_e\_\{0,1\}}     & \texttt{offset}      & additive \\
\texttt{meanVehicleNumber\_e2\_e\_\{0,1\}}       & \texttt{impatience}  & additive \\
\texttt{speed\_loop\_e\_\{0,1\}\_near}           & \texttt{decel}       & multiplicative \\
\texttt{meanSpeed\_e2\_e\_\{0,1\}}               & \texttt{maxSpeed}    & multiplicative \\
\texttt{speed\_loop\_exit\_e\_\{0,1\}}           & \texttt{lcSpeedGain} & multiplicative \\
\texttt{CO2/NOx/PMx/fuel\_abs\_edge\_e2c}        & \texttt{sigma}       & multiplicative \\
\texttt{flow\_loop\_e\_\{0,1\}\_near}            & \texttt{tau}         & multiplicative \\
\texttt{flow\_loop\_e\_0\_far}                   & \texttt{lcStrategic} & multiplicative \\
\texttt{flow\_loop\_exit\_e\_\{0,1\}}            & \texttt{lcCooperative}& multiplicative \\
\texttt{meanOccupancy\_e2\_e\_\{0,1\}}           & \texttt{length}      & multiplicative \\
\texttt{occupancy\_loop\_e\_0\_near}             & \texttt{width}       & multiplicative \\
\texttt{occupancy\_loop\_e\_1\_near}             & \texttt{height}      & multiplicative \\
\texttt{occupancy\_loop\_exit\_e\_\{0,1\}}       & \texttt{actionStepLength} & multiplicative \\
\bottomrule
\end{tabular}
\caption{\textbf{SUMO target--covariate pairing.} Each target is driven by one primary
covariate. The Combination column is the rule by which \emph{secondary} covariates
combine, set by the target's class (additive for delay and jam length, multiplicative
for emissions, flow, occupancy, and speed).}
\label{tab:sumo-primary}
\end{table}

\paragraph{Primary relationships.}
Each primary relationship maps the baseline value $y$ and the primary covariate's
factor $f$ to the target under the event. The relationships below are derived from
SUMO's documented car-following, emission, and detector models~\citep{lopez2018sumo};\footnote{\url{https://sumo.dlr.de/docs/}}
we restate them in closed form. A covariate can be raised or lowered, but each metric
here responds mainly to added pressure on the network, so we first map the factor
through an \emph{effective factor}
\begin{equation*}
  f_{\mathrm{eff}} =
  \begin{cases}
    f, & f \ge 1, \\[2pt]
    1 + 1.5\,(1/f - 1), & f < 1,
  \end{cases}
\end{equation*}
which turns a reduction into an equivalent increase. Every relationship below uses
$f_{\mathrm{eff}}$.
\begin{itemize}
  \item \textbf{Delay (from \texttt{accel}, \texttt{offset}, \texttt{impatience}).}
  Delays grow exponentially with the driver:
  \begin{align*}
    \texttt{meanTimeLoss}&: y' = y + 6.0\,(e^{1.1\,(f_{\mathrm{eff}}-1)} - 1), \\
    \texttt{meanHaltingDuration}&: y' = y + 8.0\,(e^{1.2\,(f_{\mathrm{eff}}-1)} - 1), \\
    \texttt{meanVehicleNumber}&: y' = y + 4.0\,(e^{1.15\,(f_{\mathrm{eff}}-1)} - 1).
  \end{align*}
  \item \textbf{Jam length (from \texttt{minGap}).} Queue spillback grows faster than
  linearly:
  \begin{equation*}
    \texttt{meanMaxJamLengthInMeters}: y' = y + 4.5\,(f_{\mathrm{eff}}-1)^{2.5}.
  \end{equation*}
  \item \textbf{Speed (from \texttt{decel}, \texttt{maxSpeed}, \texttt{lcSpeedGain}).}
  More pressure lowers speed through a decay:
  \begin{align*}
    \texttt{speed\_loop\_near}&: y' = y\,/\,f_{\mathrm{eff}}^{0.4}, \\
    \texttt{meanSpeed}&: y' = y\,/\,f_{\mathrm{eff}}^{0.35}, \\
    \texttt{speed\_loop\_exit}&: y' = y\,/\,f_{\mathrm{eff}}^{0.3}.
  \end{align*}
  \item \textbf{Emissions (from \texttt{sigma}).} All four pollutants scale
  exponentially:
  \begin{equation*}
    \texttt{CO2/NOx/PMx/fuel}: y' = y\,e^{0.12\,(f_{\mathrm{eff}}-1)}.
  \end{equation*}
  \item \textbf{Flow (from \texttt{tau}, \texttt{lcStrategic}, \texttt{lcCooperative}).}
  Flow follows a piecewise curve, with a rising branch below $f_{\mathrm{eff}}=2$ and a
  falling branch above it as stop-and-go sets in. With our factor pool and the
  effective-factor map, every event has $f_{\mathrm{eff}}\ge 2$ (the smallest is $f=2$
  itself, and reductions map to $2.5$ and above), so only the falling branch is reached
  and flow always decreases under an event. The curve is discontinuous at
  $f_{\mathrm{eff}}=2$, which affects no generated task since no event lands on the
  rising side:
  \begin{equation*}
    y' =
    \begin{cases}
      y\,\big(1 + 0.1\,(f_{\mathrm{eff}}-1)\big), & f_{\mathrm{eff}} < 2, \\[2pt]
      y\,\cdot\,2.2\,/\,f_{\mathrm{eff}}^{1.5}, & f_{\mathrm{eff}} \ge 2.
    \end{cases}
  \end{equation*}
  \item \textbf{Occupancy (from \texttt{length}, \texttt{width}, \texttt{height},
  \texttt{actionStepLength}).} Occupancy rises faster than linearly:
  \begin{align*}
    \texttt{meanOccupancy}&: y' = y + 0.15\,(f_{\mathrm{eff}}-1)^{2.2}, \\
    \texttt{occupancy\_loop\_e\_0\_near}&: y' = y\,\big(1 + 0.05\,(f_{\mathrm{eff}}-1)^{1.5}\big), \\
    \texttt{occupancy\_loop\_e\_1\_near}&: y' = y\,\big(1 + 0.04\,(f_{\mathrm{eff}}-1)^{1.5}\big), \\
    \texttt{occupancy\_loop\_exit}&: y' = y\,\big(1 + 0.02\,(f_{\mathrm{eff}}-1)^{1.9}\big).
  \end{align*}
\end{itemize}

\begin{table}[t]
\centering
\small
\setlength{\tabcolsep}{5pt}
\begin{tabular}{lcccccccccccccccc}
\toprule
& \rotatebox{90}{\texttt{minGap}} & \rotatebox{90}{\texttt{accel}}
& \rotatebox{90}{\texttt{decel}} & \rotatebox{90}{\texttt{maxSpeed}}
& \rotatebox{90}{\texttt{sigma}} & \rotatebox{90}{\texttt{tau}}
& \rotatebox{90}{\texttt{length}} & \rotatebox{90}{\texttt{speedDev}}
& \rotatebox{90}{\texttt{offset}} & \rotatebox{90}{\texttt{width}}
& \rotatebox{90}{\texttt{height}} & \rotatebox{90}{\texttt{lcStrategic}}
& \rotatebox{90}{\texttt{lcCooperative}} & \rotatebox{90}{\texttt{lcSpeedGain}}
& \rotatebox{90}{\texttt{impatience}} & \rotatebox{90}{\texttt{actionStepLength}} \\
\midrule
$w^{\text{add}}_c$ & $.60$ & $.20$ & $.20$ & $.20$ & $.20$ & $.20$ & $.50$ & $.20$ & $.70$ & $.20$ & $.20$ & $.20$ & $.20$ & $.20$ & $.45$ & $.20$ \\
$w^{\text{mul}}_c$ & $.45$ & $.15$ & $.30$ & $.15$ & $.35$ & $.50$ & $.35$ & $.25$ & $.30$ & $.15$ & $.15$ & $.20$ & $.20$ & $.20$ & $.15$ & $.15$ \\
\bottomrule
\end{tabular}
\caption{\textbf{SUMO secondary covariate weights.} $w^{\text{add}}_c$ is used for the
additive classes (delay, jam length) and $w^{\text{mul}}_c$ for the multiplicative
classes (emissions, flow, occupancy, speed). The effect shapes $e$ and $\rho$ are given
in the text.}
\label{tab:sumo-secondary}
\end{table}

\paragraph{Combining multiple covariates.}
When an event sets more than one covariate, the target starts from its primary
relationship and each additional active covariate applies a further, dampened effect.
A target's class fixes whether these effects combine additively (delay, jam length) or
multiplicatively (emissions, flow, occupancy, speed).
Let $n$ be the number of active secondary covariates in the event and set the joint
dampening
\begin{equation*}
  s =
  \begin{cases}
    1, & n \le 1, \\
    \tfrac{1}{\sqrt{n}}, & n > 1.
  \end{cases}
\end{equation*}
Each secondary covariate $c$ carries a weight $w_c$ (\Cref{tab:sumo-secondary}) and the
running target value is updated as
\begin{equation*}
  \text{additive: } y \leftarrow y + (s\,w^{\text{add}}_c)\,e(f_{\mathrm{eff}}),
  \qquad
  \text{multiplicative: } y \leftarrow y\,\big[\,1 + (s\,w^{\text{mul}}_c)\,(\rho(f_{\mathrm{eff}}) - 1)\,\big].
\end{equation*}
The additive shift $e(\cdot)$ and multiplicative ratio $\rho(\cdot)$ reuse the custom
curves above for a few sensitive covariates, and take a generic form otherwise:
\begin{itemize}
  \item $e(f_{\mathrm{eff}}) = 8.0\,(e^{1.2\,(f_{\mathrm{eff}}-1)}-1)$ for
  \texttt{offset}, $4.5\,(f_{\mathrm{eff}}-1)^{2.5}$ for \texttt{minGap},
  $4.0\,(e^{1.15\,(f_{\mathrm{eff}}-1)}-1)$ for \texttt{impatience},
  $10.0\,(f_{\mathrm{eff}}-1)^2$ for \texttt{length}, and
  $8.0\,(f_{\mathrm{eff}}-1)^2$ otherwise.
  \item $\rho(f_{\mathrm{eff}}) = f_{\mathrm{eff}}^{-0.4}$ for \texttt{decel},
  $f_{\mathrm{eff}}^{-0.35}$ for \texttt{maxSpeed}, $e^{0.12\,(f_{\mathrm{eff}}-1)}$
  for \texttt{sigma}, the flow piecewise form above for \texttt{tau}, and
  $1 + \tfrac{m}{2}\,(f_{\mathrm{eff}}-1)^{1.5}$ otherwise, with $m = 0.28$ for
  \texttt{actionStepLength} and $m = 0.25$ for the rest.
\end{itemize}

The direction and qualitative form of every effect follow the mechanisms above. The
magnitudes, i.e.\ the weights $w_c$ and the constants in $e$ and $\rho$, are chosen so
that each covariate leaves a clear and separable imprint on the target, rather than
being derived from the simulator.

\section{Aggregation and Metrics}
\label{app:metrics}

We provide here the details of how each task is scored and how the per-task scores are aggregated in the benchmark.

\subsection{Per-task scoring}
Each metric reduces a task's forecast to a single number. A point-forecast metric takes the median (point) forecast $\widehat{\Xb}_F=[\widehat{X}_{t+1},\dots,\widehat{X}_T]$ and the ground-truth future $\Xb_F=[X_{t+1},\dots,X_T]$; a probabilistic metric takes the predicted quantiles $\widehat{\Xb}^{(q)}_F$ over a set of levels $Q$. The forecast horizon runs from $t+1$ to $T$, so it has $T-t$ steps. 

\subsection{Aggregates}
Following \citet{shchur2025fevbench}, we report two aggregates over a task set $\mathcal{S}$, defined in \Cref{sec:benchmark-aggregation} and restated here. Let $\mathcal{M}$ be the set of evaluated models, $M=\lvert\mathcal{M}\rvert$, and $m_j(\mathcal{T})$ the error (wQL or WAPE) of model $j$ on task $\mathcal{T}$.

\paragraph{Skill score.}
We use the skill score~\citep{hyndman2018forecasting}, a model's relative reduction in error against a fixed reference model $\beta$, the naive baseline that repeats the last observed value. For model $j$ over $\mathcal{S}$,
\begin{equation*}
  \mathrm{Skill}_j(\mathcal{S})=1-\exp\!\left(\frac{1}{\lvert\mathcal{S}\rvert}\sum_{\mathcal{T}\in\mathcal{S}}\log\,\mathrm{clip}\!\left(\frac{m_j(\mathcal{T})}{m_\beta(\mathcal{T})}\right)\right),
  \qquad
  \mathrm{clip}(x)=\max\!\big(10^{-2},\min(x,10^{2})\big).
\end{equation*}
The clip constrains each ratio to $[10^{-2},10^{2}]$, limiting the influence of a few extreme tasks. The score is a geometric mean of the per-task error ratios: it equals $0$ when a model matches the baseline, is positive when it improves on the baseline, and approaches $1$ (or 100\%) as forecasts approach perfect accuracy. 

\paragraph{Average win rate.}
The average win rate is the fraction of tasks on which a model attains a lower error than each competing model, with ties counted as half a win. For model $j$ over $\mathcal{S}$,
\begin{equation*}
  \mathrm{Win}_j(\mathcal{S})=\frac{1}{\lvert\mathcal{S}\rvert\,(M-1)}\sum_{\mathcal{T}\in\mathcal{S}}\ \sum_{\substack{k\in\mathcal{M}\\ k\neq j}}\Big(\mathbb{1}\!\left[m_j(\mathcal{T})<m_k(\mathcal{T})\right]+\tfrac{1}{2}\,\mathbb{1}\!\left[m_j(\mathcal{T})=m_k(\mathcal{T})\right]\Big).
\end{equation*}
The win rate ranges from $0$ to $1$ (or 100\%). It is magnitude-invariant, reflecting only the ordering of models on each task and not the size of their differences, and its value depends on the composition of the model set $\mathcal{M}$, changing as models are added or removed.

\subsection{Metrics}
The main paper reports the weighted quantile loss (wQL) and the weighted absolute percentage error (WAPE), defined in \Cref{sec:benchmark-aggregation}. We restate them here.
For a quantile level $q$, let $\Lambda_q(X,\widehat{X})=q\,(X-\widehat{X})_{+}+(1-q)\,(\widehat{X}-X)_{+}$ be the quantile (pinball) loss, with $(\cdot)_{+}=\max(\cdot,0)$. wQL averages this over the levels $Q$ and normalizes by the target's aggregate magnitude; WAPE is the point-forecast analogue on the median forecast:
\begin{equation*}
  \mathrm{wQL}(\mathcal{T})=\frac{1}{\lvert Q\rvert}\sum_{q\in Q}\frac{\sum_{\tau=t+1}^{T}\Lambda_q\big(X_\tau,\widehat{X}^{(q)}_\tau\big)}{\sum_{\tau=t+1}^{T}\lvert X_\tau\rvert},
  \qquad
  \mathrm{WAPE}(\mathcal{T})=\frac{\sum_{\tau=t+1}^{T}\lvert X_\tau-\widehat{X}_\tau\rvert}{\sum_{\tau=t+1}^{T}\lvert X_\tau\rvert}.
\end{equation*}
We adopt these metrics for their scale-invariance: our targets span four simulators with disparate scales, and normalizing by each target's aggregate magnitude keeps errors comparable across targets and well suited to the relative aggregation above.

\section{Experimental Setup}
\label{app:experimental-setup}

Every model is given the target's history together with the covariates: the past covariates over the history and the future covariates over the forecast horizon. The foundation models produce nine quantile levels, $0.1, 0.2, \dots, 0.9$, which we score with the metrics of \Cref{app:metrics}. The LLMs output a point forecast, which we score as the median: their wQL is the quantile loss at the $0.5$ level only (\Cref{sec:benchmark-models}).
We prompt the LLMs for the median rather than many quantiles because asking for many quantiles lengthens the output and causes formatting errors that persist across retries. Scoring the median alone inherently disadvantages the LLMs under probabilistic metrics, since the foundation models' wQL averages over nine levels, $0.1$ to $0.9$, most of them away from the median, where a calibrated forecast incurs a lower loss; we leave more efficient ways to elicit quantiles from LLMs to future work.

We run the LLMs through Google Cloud Vertex Model Garden and each foundation model on a single A100 GPU per task. The inference times in \Cref{tab:leaderboard} thus compare on-device GPU times with API round-trip times. These differ, but they are the conditions under which each family is used in practice, and represent the best each can achieve.

\subsection{Foundation models}
We use the following public checkpoints.
\begin{itemize}[leftmargin=*]
  \item Chronos-2: \texttt{amazon/chronos-2}\footnote{\url{https://huggingface.co/amazon/chronos-2}}
  \item TimesFM-3.0: \texttt{google/timesfm-3.0-pytorch}\footnote{\url{https://huggingface.co/google/timesfm-3.0-pytorch}}
  \item t0-gamma: \texttt{theforecastingcompany/t0-gamma}\footnote{\url{https://huggingface.co/theforecastingcompany/t0-gamma}}
  \item Timestral: \texttt{mistralai/Timestral-209M}\footnote{\url{https://huggingface.co/mistralai/Timestral-209M}}
  \item TiRex-2: \texttt{NX-AI/TiRex-2}\footnote{\url{https://huggingface.co/NX-AI/TiRex-2}}
  \item TS-ICL: \texttt{taharnbl/TS-ICL}\footnote{\url{https://huggingface.co/taharnbl/TS-ICL}}
  \item TabPFN-3.5: \texttt{Prior-Labs/tabpfn\_3\_5}\footnote{\url{https://huggingface.co/Prior-Labs/tabpfn_3_5}}, which we run for time series through the TabPFN-TS harness~\citep{tabpfnts}.
\end{itemize}

\subsection{Large language models}
We evaluate two frontier LLMs: Gemini-3.1-Pro (\texttt{gemini-3.1-pro-preview}) and Claude-Opus-4.8 (\texttt{claude-opus-4-8}). We run both at medium and high reasoning effort.

We pass the target history and the covariates to the model as text, as four CSV files, and ask it to return the forecast in a fixed tagged format. The variable names in these files are replaced by dummy identifiers such as \texttt{<target\_1>} and \texttt{<covariate\_1>}. The exact prompt is:

\begin{quote}
\small
\begin{verbatim}
# Task Definition

## File Descriptions
- covariates.csv: Time-series data of input variables (features) that
  might influence the system or be affected by events.
- variates.csv: Time-series data of output variables (targets) that
  you need to forecast.
- future_variates_blanked.csv: Time-series data of output variables
  (targets) that you need to forecast.
- future_covariates.csv: Time-series data of input variables (features)
  for the future timestamps - the same timestamps you need to forecast
  the output variables for.

## Goal
You must provide the forecast values in a strict structured format. Do
not include any conversational text, analysis, or explanations in your
final response. Only output the forecast data within the specified tags.
For each variable listed in future_variates_blanked.csv, output the
forecast values enclosed in tags named after the variable. The content
inside the tags must be a list of (timestamp, value) pairs separated by
commas. You must output the variables in the exact same order as the
columns appear in the header of future_variates_blanked.csv (ignoring
the timestamp column).

Example Format:
<variable_name>
(timestamp, value), (timestamp, value), ...
</variable_name>

Replace variable_name with the actual name of the variable you are
forecasting. Ensure you provide values for all expected timestamps
found in future_variates_blanked.csv. 
\end{verbatim}
\end{quote}

If a call fails or the output cannot be parsed into a forecast for every target and timestamp, we retry the same call, up to $10$ attempts per task, with a short wait that grows with each attempt. %
A task with no valid forecast after all attempts is imputed with the naive baseline forecast. We find that this happens on at most $16$ of $37{,}665$ tasks for any LLM, under $0.05\%$ in every case, so it does not affect the reported aggregates. We report the number of failures in \Cref{tab:imputed-windows}.

\input{tables/table_v10_imputed.tex}

\subsection{Regression baselines}
We fit each regression baseline separately on the history of every task. Each training row is one time step of the history: the input is the covariates at that step, $\zb_\tau$, and the label is the target $X_\tau$. At test time, the model predicts each forecast step $X_{t+h}$ separately, given its known covariates $\zb_{t+h}$. We use Linear Regression and Ridge Regression from scikit-learn, with the covariates standardized on the history and a regularization strength of $10$ for Ridge Regression, and XGBoost with $150$ trees of depth $6$, a learning rate of $0.05$, and row and column subsampling of $0.9$. These settings are fixed across all tasks and are not tuned. Like the LLMs, the regression baselines output a point forecast, which we score as the median. We run them on CPU. Tuned variants of these models are explored in \Cref{app:tuned-regression}.

\section{Limitations and Future Work}
\label{app:future-work}

We discuss the limitations of \bench{} and several areas for future work.

\paragraph{Harder system dynamics.}
Our tasks keep the system in a single regime: each event sets the covariates, the target responds, and the system returns to its usual range. Within this range, each relationship is exact and does not reverse. Many real-world systems may not always be so well-behaved. An intervention can push a system into a new state, such as a power grid moving from normal operation into an overload. Modelling such regime changes is important, and common in domains such as finance. An intervention can also drive a system into a hard limit, where the target saturates and stops responding. Capturing these limits is important, so a model learns when a covariate stops affecting the target. Further, interventions can also stack on one another, so the system never settles back before the next one begins~\citep{robins2000msm,bica2020crn}. Building benchmarks that cover these cases would test whether a model can reason about the behavior of the system as a whole.

\paragraph{Larger covariate spaces.}
Our tasks use a refined set of covariates, where every one of them affects the target, however, real systems can expose a vast number of possible drivers, of which only a few matter, and sometimes none do. Further, many drivers could be unrelated to the target, while some could be correlational. A model must then find the drivers that truly matter, and ignore the rest~\citep{runge2019detecting,assaad2022survey}. Metadata and other side information could help a model identify the true drivers. Building benchmarks with such irrelevant and misleading covariates is an important future direction.

\paragraph{Grounding in real-world data.}
\bench{} is designed to discover failure modes in a controlled semi-synthetic environment, where the covariate--target relationships are known exactly and the ground truth carries no noise unrelated to the covariates, so a model's error reflects how it handles the covariates. We do not test how these findings hold up under the irreducible noise and unobserved drivers of real systems. Validating our proposed axes on real-world data with covariates would be a useful next step and may reveal further findings.
Building benchmarks for what-if forecasting from purely real-world data is a grand challenge. Real data records only one course of events, where it is impossible to simulate counterfactuals~\citep{holland1986statistics,imbens2015causal}.
Synthetic control offers a partial answer: it estimates the missing counterfactual for a unit from a weighted combination of comparable units whose outcomes were recorded~\citep{abadie2010synthetic}.
Bringing these ideas to forecasting would move benchmarks closer to real applications, while keeping the ground truth needed to measure a model's response.

\paragraph{Natural-language what-if queries.}
Our tasks still require engineering the right covariates: we assume the right drivers have been chosen already, and are set by hand to simulate scenarios. A more general system would instead take a what-if query as posed, in natural language, and construct the covariates itself. Benchmarks that evaluate this end to end, from a free-form query to a forecast are promising directions for future work~\citep{williams2025contextiskey,liu2024timemmd,lee2025timecap}. Such queries carry their own nuances: they may be precise, broad, or ambiguous, and may name the drivers only indirectly. LLMs and agents are natural candidates to interpret them, turning a query into concrete covariates and opening scenario analysis to many more users, while keeping those covariates explicit for inspection. A capable system should also know when a query cannot be answered from the available history, and say so, instead of returning a confident but unsupported forecast. Benchmarks that capture these needs are important for bringing such systems into real-world use and deployment.

\paragraph{From evaluation to training.}
Most broadly, the simulators behind \bench{} need not serve only for evaluation; they can also be used for training. A model that learns a system in-context and forecasts its response to unseen interventions would do more than predict: it could plan, weigh possible actions, and help make decisions about the system it has learned, akin to \emph{world models}~\citep{ha2018worldmodels,hafner2023dreamerv3,lecun2022jepa}.
We see \bench{} as a first step toward forecasters that can answer such questions.

%% file: tables/table_v10_imputed.tex
\begin{table}[H]
\centering
\small
\begin{tabular}{lrr}
\toprule
\textbf{Model} & \textbf{Imputed windows} & \textbf{\% of windows} \\
\midrule
Claude-Opus-4.8 (High) & 16 & 0.04\% \\
Claude-Opus-4.8 (Medium) & 2 & 0.01\% \\
Gemini-3.1-Pro (High) & 10 & 0.03\% \\
Gemini-3.1-Pro (Medium) & 0 & 0.00\% \\
\bottomrule
\end{tabular}
\caption{Number of task windows on which each LLM returned no valid forecast after all retries, out of $37{,}665$ windows. These windows are imputed with the naive baseline forecast, so they receive zero skill. All other models return a forecast on every window and have no imputed windows.}
\label{tab:imputed-windows}
\end{table}

%% file: inputs/appendix_v2.tex
\providecommand{\bench}{\textsc{What-IfBench}}

\section{Skill across the number of events, event length and context length}
\label{app:axis-ablations}

\begin{figure}[H]
    \centering
    \includegraphics[width=\linewidth]{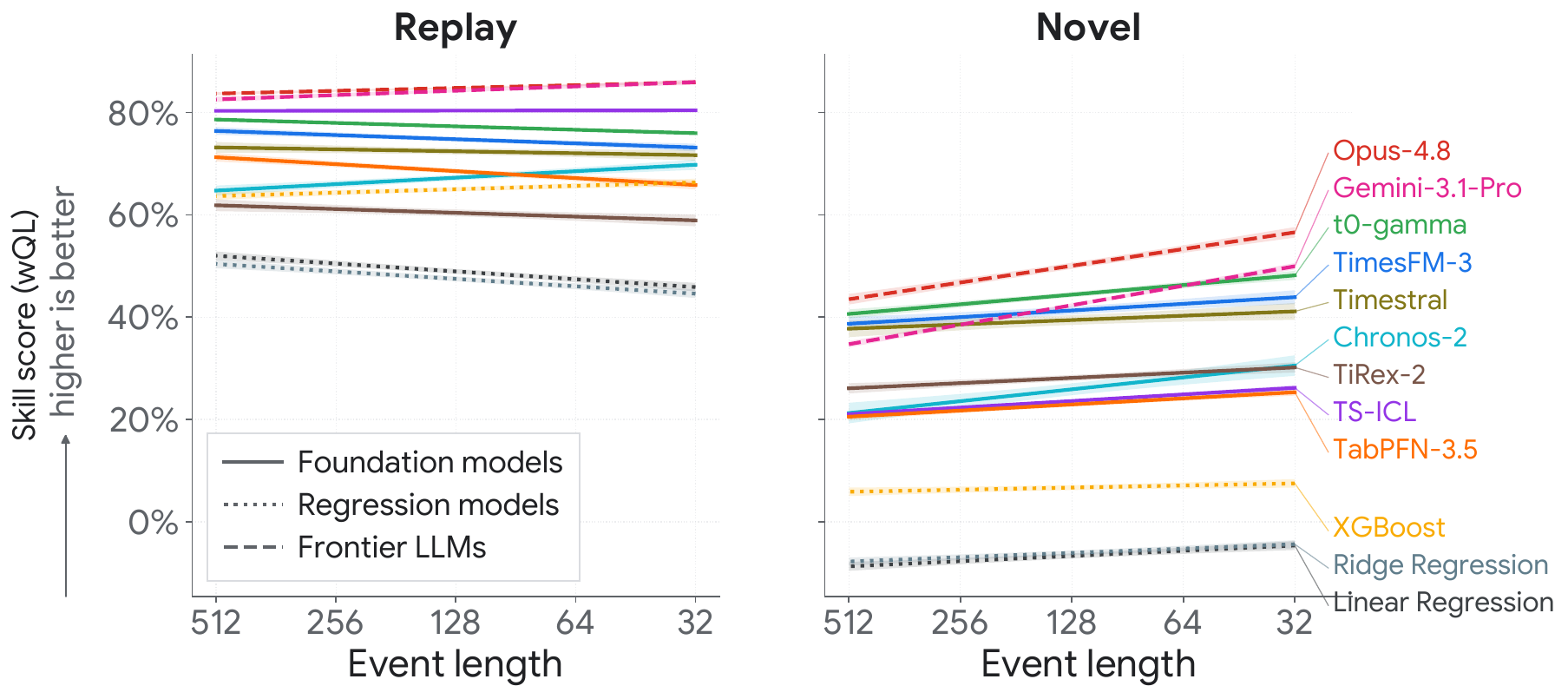}
    \caption{Skill scores with different event lengths on \bench{}, on tasks with
    Replay (left) and Novel (right) futures respectively. Lines are least-squares trends of skill on $\log_2$ of the event length; they summarise the direction of change.
    }
    \label{fig:v10-eventlen-pair-wql}
\end{figure}

\begin{figure}[H]
    \centering
    \includegraphics[width=\linewidth]{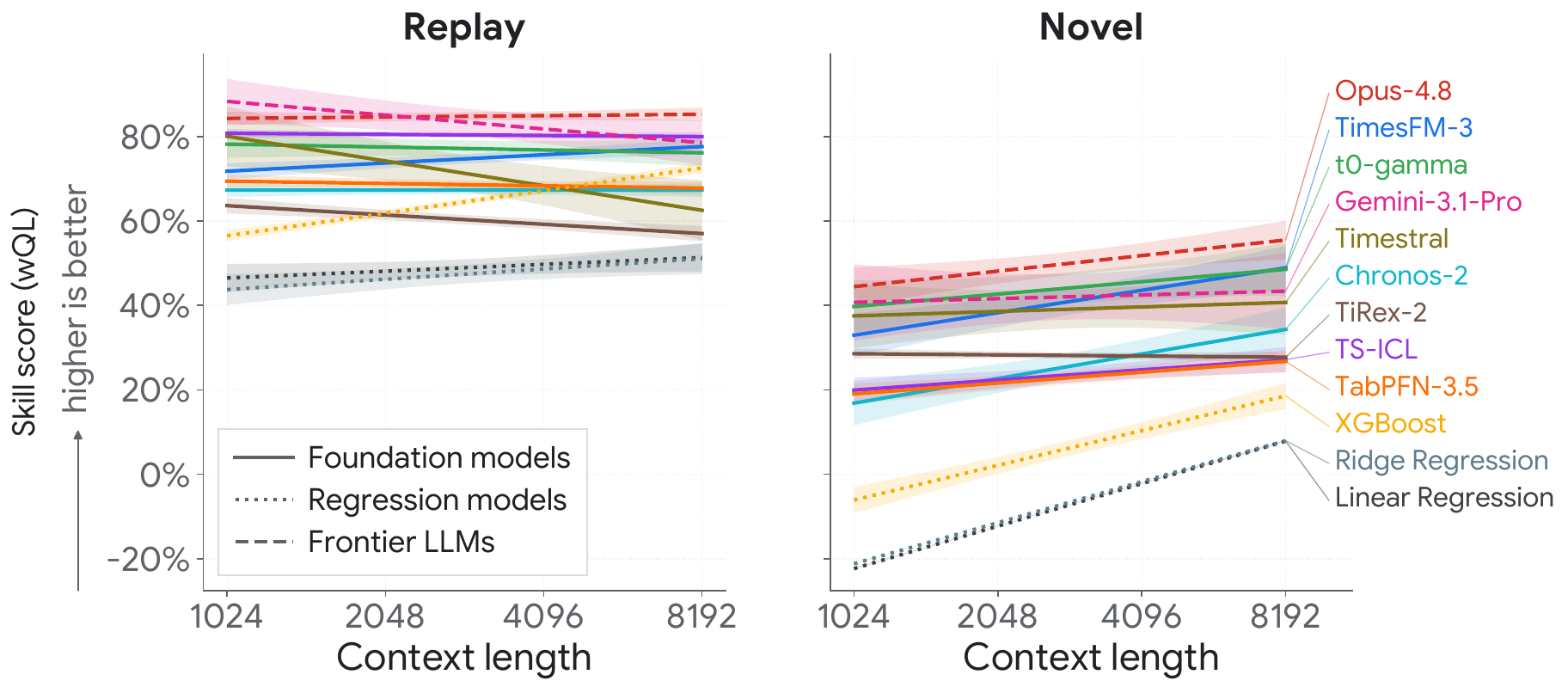}
    \caption{Skill scores with different context lengths on \bench{}, on tasks with
    Replay (left) and Novel (right) futures respectively. All models receive the same input and produce the same output: each consumes the full history at every context length, without truncation, and forecasts the same horizon. Tasks are grouped by their nominal context length; for the longest events, the actual context is longer (\Cref{app:task-layout}). Lines are least-squares trends of skill on $\log_2$ of the context length; they summarise the direction of change.}
    \label{fig:v10-ctxlen-pair-wql}
\end{figure}

The main text reports how skill varies with the number of historical events. Here
we show the analogous trends for event length and context length, again split by
Replay and Novel futures. Both echo the number-of-events result: on Novel futures,
skill improves as the context holds more past events: that is, when events are
shorter (\Cref{fig:v10-eventlen-pair-wql}) or the context is longer
(\Cref{fig:v10-ctxlen-pair-wql}). We also report how the regression baselines and TiRex-2 respond to the number of events (\Cref{tab:linear-events,tab:tirex-events}).

\paragraph{Linear and Ridge Regression and the number of events.}
Most models change little on Replay futures as the number of events grows, but Linear and Ridge Regression get worse on two simulators. \Cref{tab:linear-events} compares tasks that differ only in the number of events, sharing the target, event length, context length and number of covariates. On SUMO, their Replay skill falls from about $80\%$ to $5\%$ and their Novel skill from $-4\%$ to $-30\%$; on OpenDC, Replay skill falls from $80\%$ to $54\%$ and Novel skill from $29\%$ to $-35\%$. On EPANET and GridLAB-D, both improve with more events. We believe this is because, with only a few events, a single linear fit can pass close to the few covariate values it has seen, while with many events it has to average over many points and loses the fit, even at values it has already seen; this would explain why skill drops on both Replay and Novel futures.

\input{tables/table_linear_events.tex}

\paragraph{TiRex-2 and the number of events.}
TiRex-2 is the only foundation model whose skill falls with more events on both Replay and Novel futures (\Cref{fig:v10-kevents-pair-wql}). Comparing tasks that differ only in the number of events (\Cref{tab:tirex-events}), its skill falls from $61.4\%$ to $52.2\%$ on Replay futures and from $33.8\%$ to $26.5\%$ on Novel futures, while TimesFM-3 gains on both ($+2.0$ and $+4.6$ points). The drop comes from two simulators: on SUMO, Replay skill falls from $78.5\%$ to $55.3\%$, on all $29$ targets, and on OpenDC, Novel skill falls from $23.1\%$ to $-24.9\%$, on $14$ of $22$ targets. On EPANET and GridLAB-D, it improves slightly. Other foundation models show weaker versions of the same pattern, for example Chronos-2 on SUMO Replay futures ($-21.4$ points), so these settings become harder with many events for several models, and TiRex-2 is the most affected.

\input{tables/table_tirex_events.tex}

\paragraph{Gemini-3.1-Pro and the context length.}
Gemini-3.1-Pro's drop at $64$ events in \Cref{fig:v10-kevents-pair-wql} comes from the longest context, where all such tasks lie: its skill falls sharply at $C_4$ on both Replay and Novel futures, while more events at a fixed context change it little (\Cref{tab:llm-context-events}). On Replay futures, Claude-Opus-4.8 and TimesFM-3 change much less with context. We believe Gemini-3.1-Pro handles the longest inputs less well.

\input{tables/table_llm_context_events.tex}

\section{Skill scores under WAPE}
\label{app:metric-skill}

\begin{figure}[H]
  \centering
  \includegraphics[width=0.82\linewidth]{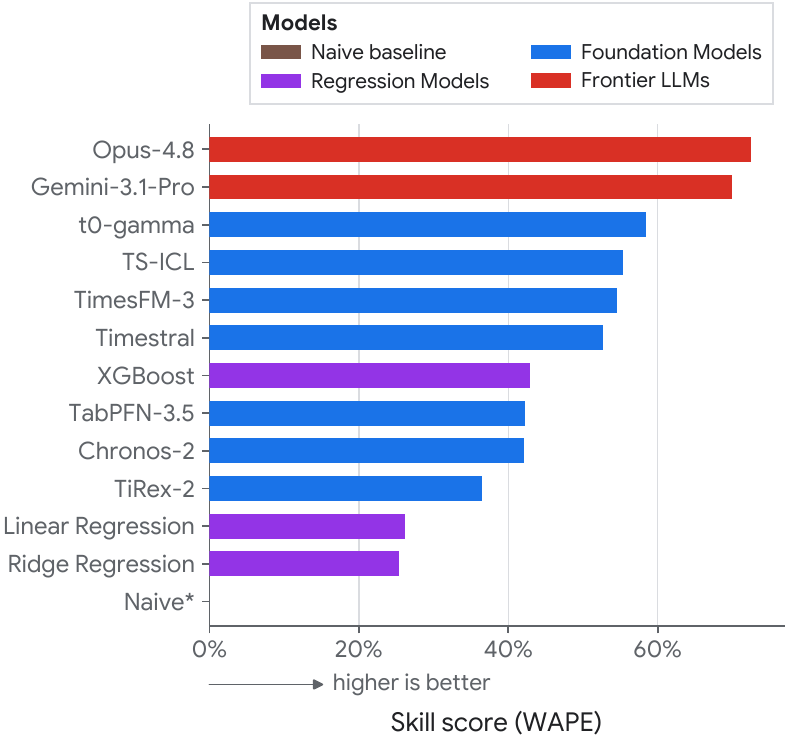}
  \caption{Aggregate skill leaderboard on \bench{} as per the WAPE metric.}
  \label{fig:v10-skill-lead-wape}
\end{figure}

\begin{figure}[H]
  \centering
  \includegraphics[width=0.82\linewidth]{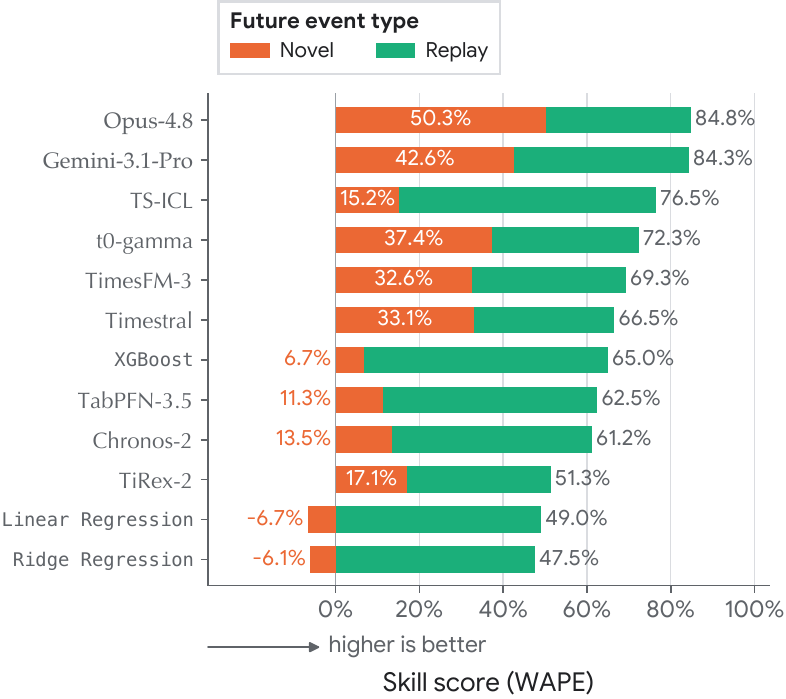}
  \caption{Replay vs Novel skill of \fw{palatino_llms}, \fw{optima_numerical} \fw{optima_foundation} \fw{optima_models}, and \fw{menlo_regression} \fw{menlo_models} on \bench{} as per the WAPE metric.}
  \label{fig:v10-skill-rn-wape}
\end{figure}

\Cref{fig:v10-skill-lead-wape} gives the aggregate skill leaderboard under WAPE, and
\cref{fig:v10-skill-rn-wape} splits it into Replay vs Novel skill.

\begin{figure}[H]
  \centering
  \includegraphics[width=\linewidth]{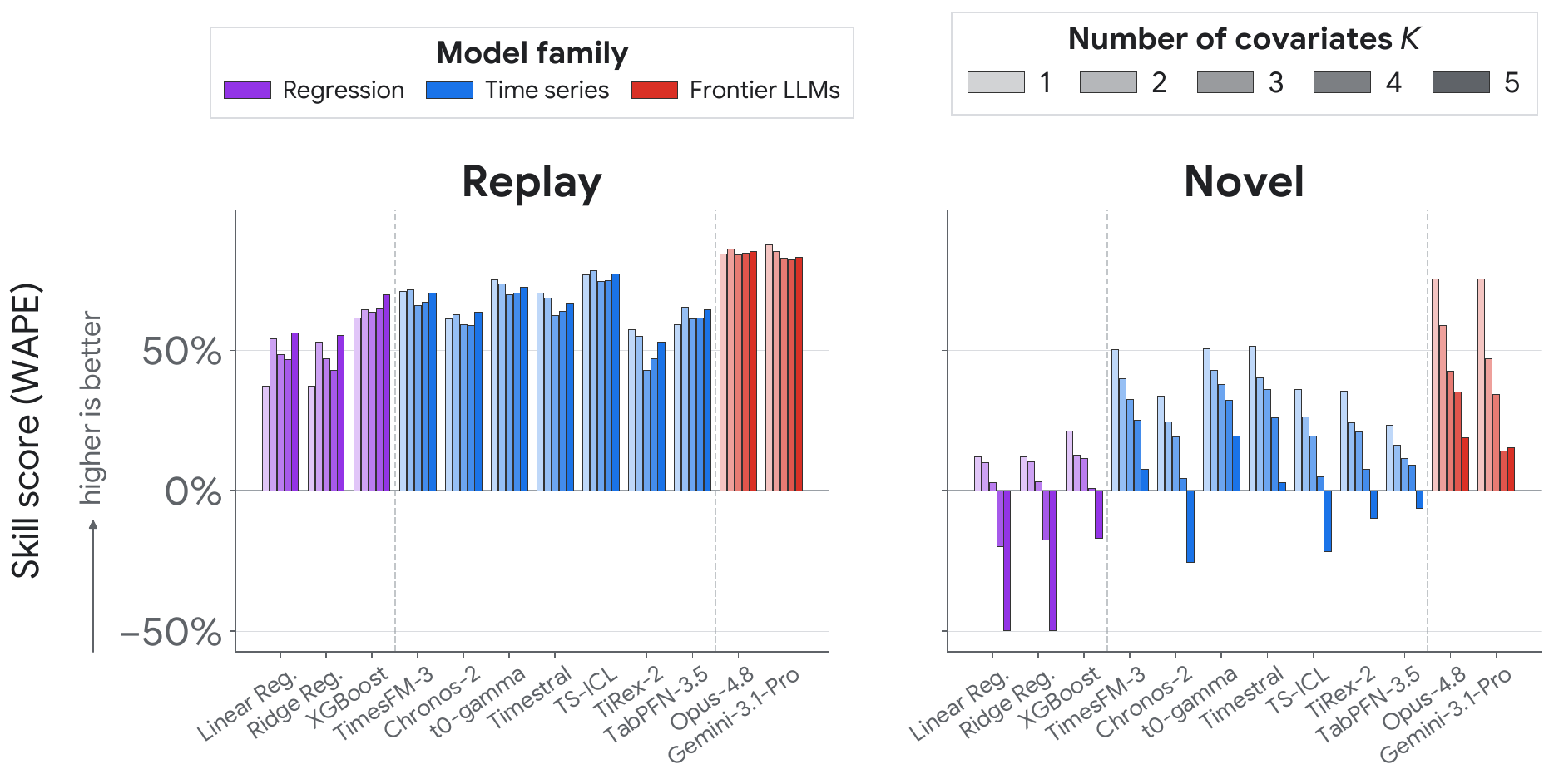}
  \caption{Skill scores as per the WAPE metric on tasks with different number of covariates on \bench{}, on Replay (left) and Novel (right) future types respectively.}
  \label{fig:v10-by-k-wape}
\end{figure}

\begin{figure}[H]
    \centering
    \includegraphics[width=\linewidth]{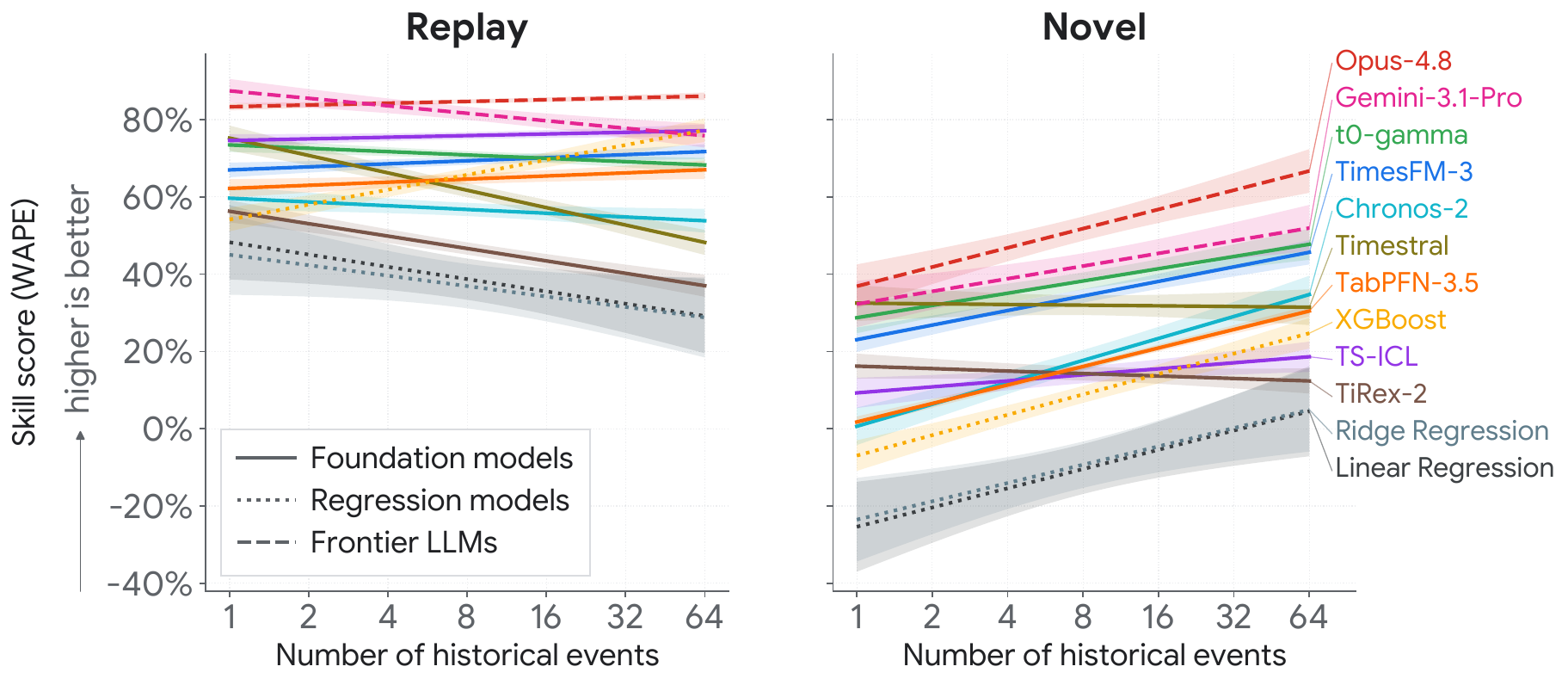}
    \caption{Skill scores as per the WAPE metric with different number of historical events on \bench{}, on tasks with Replay (left) and Novel (right) futures respectively. Lines as in \Cref{fig:v10-kevents-pair-wql}.}
    \label{fig:v10-kevents-pair-wape}
\end{figure}

\begin{figure}[H]
    \centering
    \includegraphics[width=\linewidth]{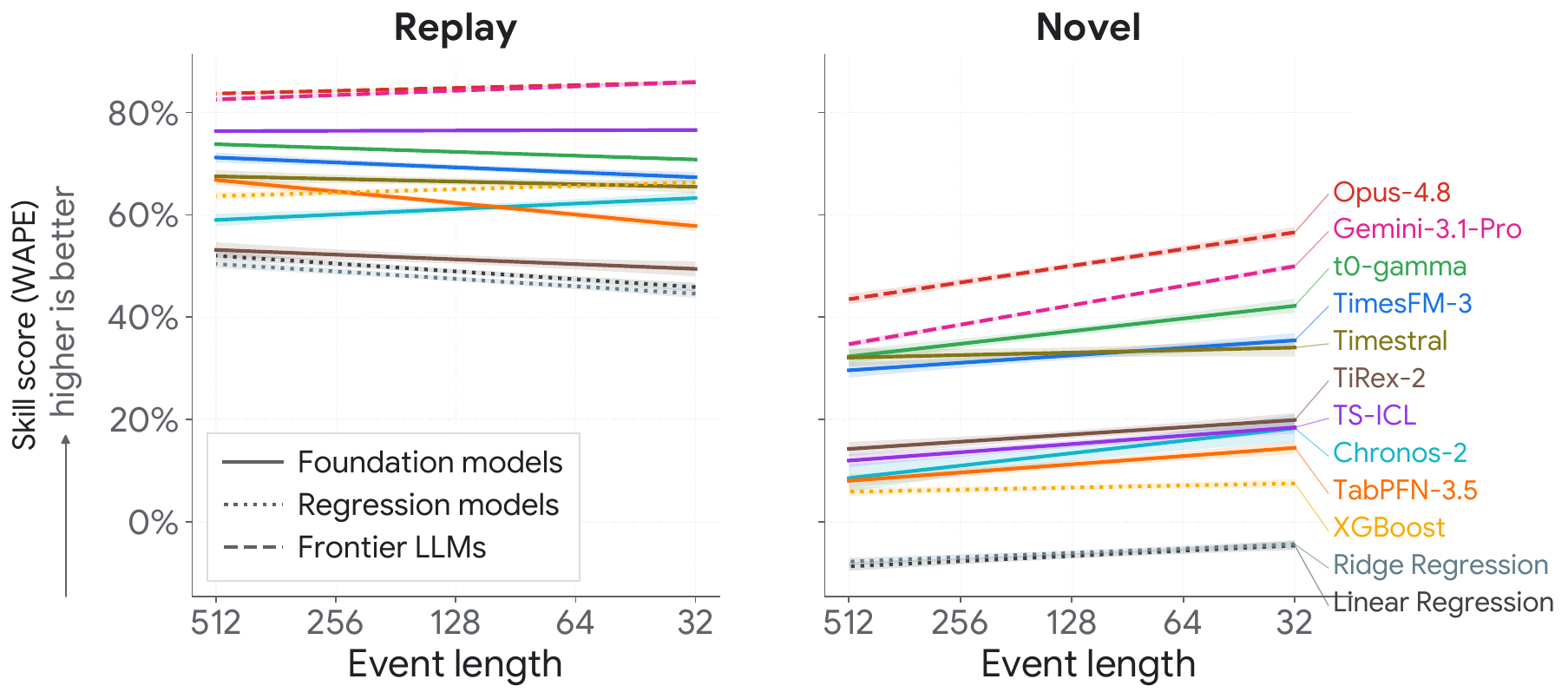}
    \caption{Skill scores as per the WAPE metric with different event lengths on \bench{}, on tasks with Replay (left) and Novel (right) futures respectively. Lines as in \Cref{fig:v10-kevents-pair-wql}.}
    \label{fig:v10-eventlen-pair-wape}
\end{figure}

\begin{figure}[H]
    \centering
    \includegraphics[width=\linewidth]{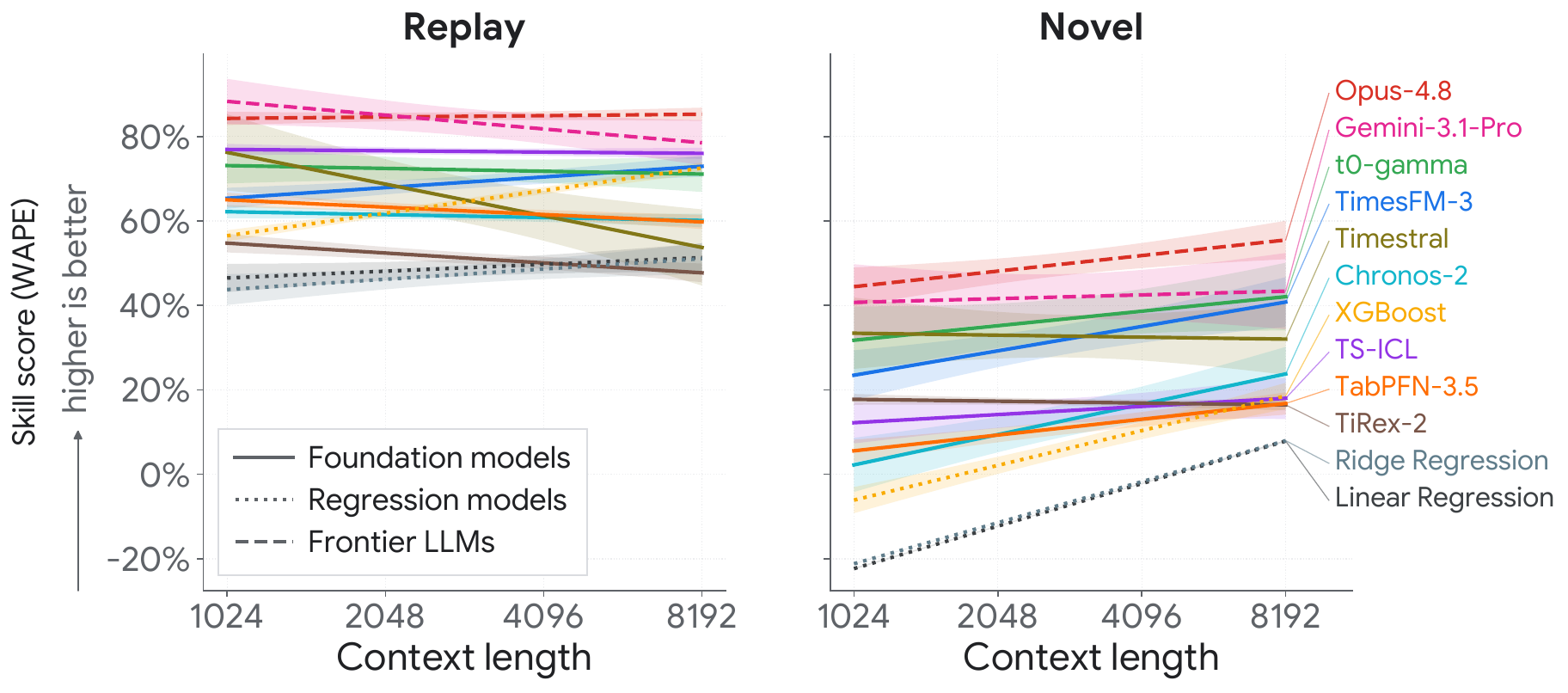}
    \caption{Skill scores as per the WAPE metric with different context lengths on \bench{}, on tasks with Replay (left) and Novel (right) futures respectively. All models receive the same input and produce the same output: each consumes the full history at every context length, without truncation, and forecasts the same horizon. Tasks are grouped by their nominal context length; for the longest events, the actual context is longer (\Cref{app:task-layout}). Lines as in \Cref{fig:v10-kevents-pair-wql}.}
    \label{fig:v10-ctxlen-pair-wape}
\end{figure}

Further, the main text shows how wQL skill falls as the number of covariates $K$
grows (\cref{fig:v10-by-k-wql}). The same trend holds under WAPE: every model loses skill on Novel futures as $K$ grows. We provide here the aggregate skill leaderboard partitioned by the number of covariates, under WAPE, in
\cref{fig:v10-by-k-wape}.
Finally, \cref{fig:v10-kevents-pair-wape,fig:v10-eventlen-pair-wape,fig:v10-ctxlen-pair-wape} give the WAPE counterparts of the number-of-events, event-length, and context-length figures (\cref{fig:v10-kevents-pair-wql,fig:v10-eventlen-pair-wql,fig:v10-ctxlen-pair-wql}); the trends match those under wQL.

\section{Results by Simulator}
\label{app:by-simulator}

The main text pools the four simulators. Here we separate the results by simulator: OpenDC (data centers), GridLAB-D (power grids), EPANET (water networks), and SUMO (road traffic). The simulators differ in size: EPANET contributes 13,265 tasks, SUMO 11,600, OpenDC 8,800, and GridLAB-D 4,000, so the pooled score is weighted toward the larger ones. All scores are wQL skill scores, measured against the naive baseline on the windows of the same simulator.

\paragraph{Overall score by simulator.}
\Cref{tab:leaderboard-by-simulator} gives the overall skill score of each model on each simulator, next to the pooled score over all four.

\input{tables/table_v10_by_simulator.tex}

\takeaway{No model is best everywhere. Claude-Opus-4.8 (High) leads on OpenDC, GridLAB-D, and SUMO, but t0-gamma leads on EPANET. Scores are highest on SUMO for both LLM families and for six of the seven foundation models.}

\begin{figure}[H]
  \centering
  \begin{minipage}{0.49\linewidth}\centering\includegraphics[width=\linewidth]{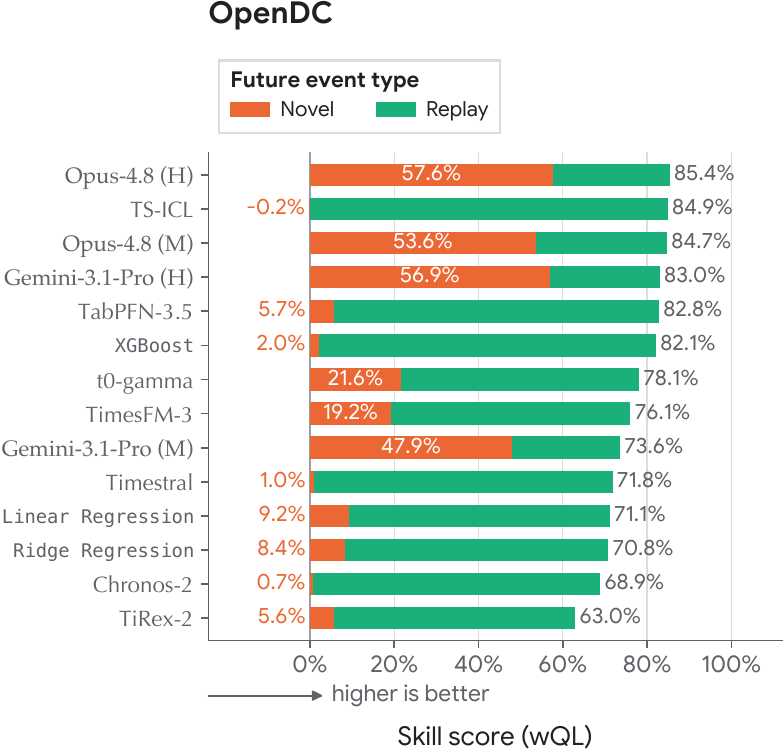}\end{minipage}\hfill
  \begin{minipage}{0.49\linewidth}\centering\includegraphics[width=\linewidth]{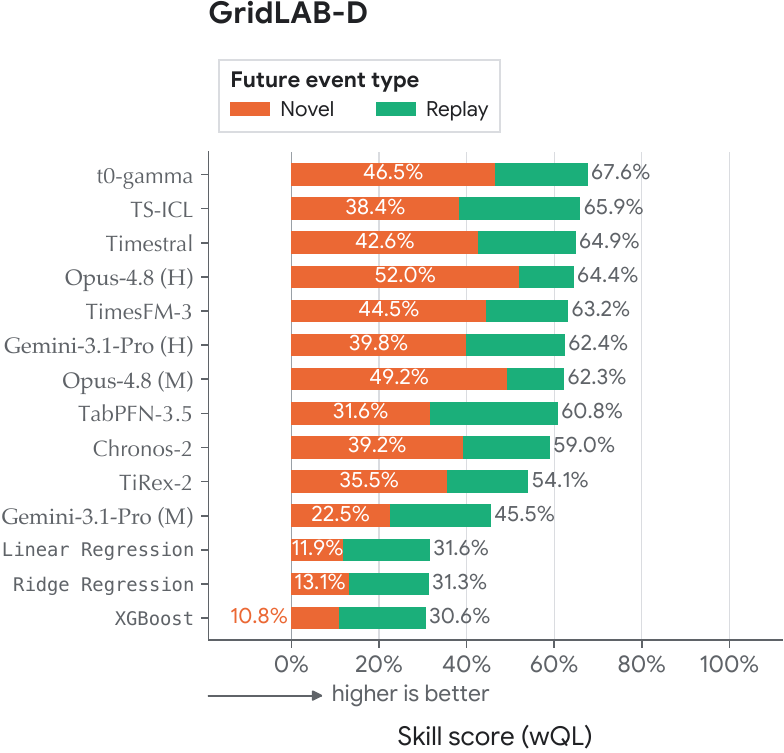}\end{minipage}\\[1.2em]
  \begin{minipage}{0.49\linewidth}\centering\includegraphics[width=\linewidth]{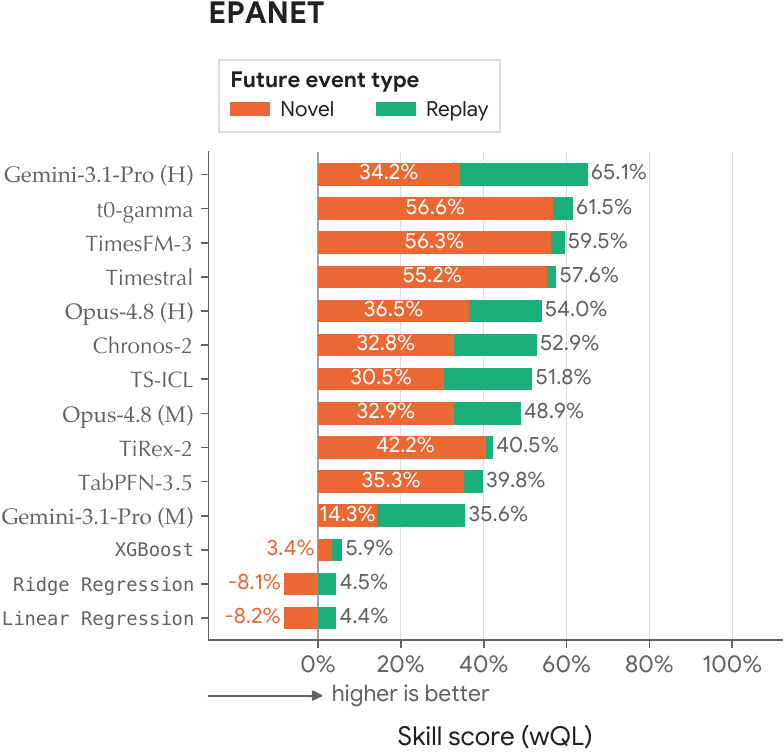}\end{minipage}\hfill
  \begin{minipage}{0.49\linewidth}\centering\includegraphics[width=\linewidth]{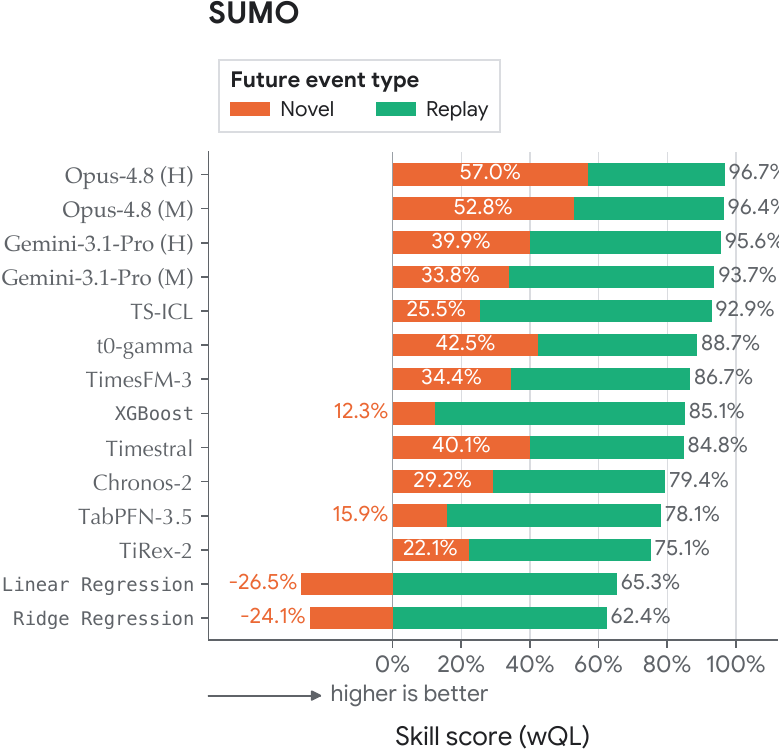}\end{minipage}
  \caption{Skill score (wQL) of \fw{palatino_llms}, \fw{optima_numerical} \fw{optima_foundation} \fw{optima_models}, and \fw{menlo_regression} \fw{menlo_models} on Replay and Novel futures, separately for each simulator. In each panel, the orange segment is a model's skill on Novel futures and the full bar is its skill on Replay futures; models are sorted by Replay skill within each panel, and all panels share the same horizontal axis. (H) and (M) mark High and Medium reasoning effort.}
  \label{fig:by-simulator-replay-novel}
\end{figure}

\paragraph{Replay and Novel futures by simulator.}
\Cref{fig:by-simulator-replay-novel} splits each score into Replay futures, which repeat covariate values already seen in the history, and Novel futures, which set the covariates beyond that range.

\takeaway{Novel futures are much harder than Replay futures on OpenDC and SUMO. On these two simulators, several models fall to about zero or below on Novel futures, for example TS-ICL on OpenDC and the linear and ridge regressions on SUMO. On EPANET and GridLAB-D, the drop from Replay to Novel is smaller.}

\begin{figure}[H]
  \centering
  \includegraphics[width=0.85\linewidth]{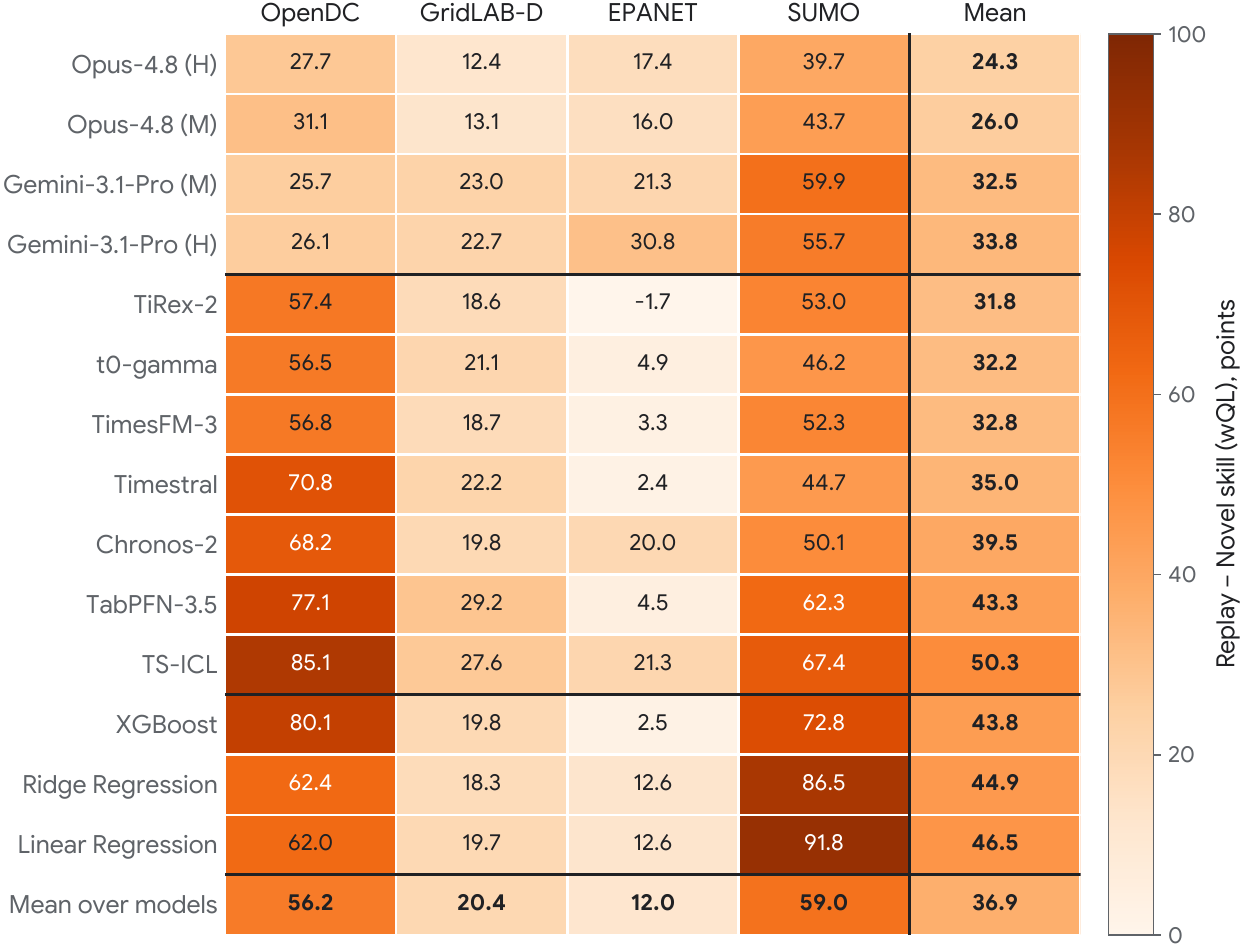}
  \caption{Replay--Novel gap (Replay skill minus Novel skill, wQL, in percentage points) for each model and simulator. Larger values mean a larger drop from Replay to Novel futures. The last column is the mean over simulators and the last row is the mean over models. Rows are sorted by the mean gap within each group of models.}
  \label{fig:by-simulator-gap}
\end{figure}

\paragraph{The Replay--Novel gap.}
To compare the drop across models and simulators in one view, we define the \emph{Replay--Novel gap} of a model on a simulator as its Replay skill minus its Novel skill, in percentage points. Both skills are measured against the naive baseline. The gap is the number of skill points a model loses when the future sets the covariates beyond the range seen in the history. A gap of zero means the model handles Novel futures as well as Replay futures. A large gap means the model forecasts well when covariate values repeat, but does not carry what it has learned to values it has not seen. \Cref{fig:by-simulator-gap} shows the gap for every model and simulator, with the mean over simulators in the last column and the mean over models in the last row.

\takeaway{The gap between Replay and Novel skill is much larger on OpenDC and SUMO than on EPANET and GridLAB-D. Novel futures are therefore the harder case on OpenDC and SUMO, while on EPANET and GridLAB-D models lose relatively little. On OpenDC the four LLMs have the smallest gaps, and on SUMO the two Claude models do.}

\section{Skill Against a Covariate-Aware Baseline}
\label{app:linear-baseline}

The skill scores in the main text are measured against a naive baseline that copies the last value. We use it because it does not use covariates, so it is the same kind of forecast on every task and in every slice of the benchmark. Here we measure skill against a covariate-aware baseline, Linear Regression, to see how much models gain over a simple model that also uses the covariates. It sets a higher bar: it has a lower wQL than the naive baseline on 69.1\% of the Replay tasks and on 55.7\% of the Novel tasks.

We do not use it as the main reference because its strength depends on the slice of the benchmark. Its skill against the naive baseline is $49.0\%$ on Replay futures and $-6.7\%$ on Novel futures; on SUMO, it is $65.3\%$ and $-26.5\%$. A skill score against it therefore mixes how well a model forecasts with how well linear regression forecasts on the same slice. This makes comparisons across Replay and Novel futures, across simulators, and across the number of covariates harder to interpret.

Even so, linear regression is a valid reference for \bench{}: it uses the covariates, as a practitioner's first model would, and it sets a higher bar than copying the last value. We therefore offer it alongside the naive baseline, as an option in our benchmark leaderboard. Other models, such as XGBoost, can also be used as reference baselines for the skill score.

\takeaway{Measured against linear regression instead of the naive baseline, skill scores differ in three ways. First, every model scores much lower against linear regression than against the naive baseline, but the order of the models does not change. Second, compared with the skill against the naive baseline, skill against linear regression is much lower on Replay futures, where linear regression is strong, but not on Novel futures, where it is close to the naive baseline: for every model, skill on Novel futures is slightly higher against linear regression. As a result, the gap between Replay and Novel skill mostly disappears for the foundation models, while the LLMs, TS-ICL, and XGBoost keep it. Third, the trends with the number of covariates, the number of events, and the event length differ from those against the naive baseline, and some are reversed, because linear regression is affected by these factors too. The LLMs have the highest skill on Replay futures, and Claude-Opus-4.8 (High) has the highest on Novel futures, against either reference.}

\paragraph{Leaderboard.}
\Cref{tab:leaderboard-vs-linear} gives the skill score of each model against both references, under wQL and WAPE.

\input{tables/table_v10_vs_linear.tex}

\takeaway{Every model has a much lower skill score against linear regression than against the naive baseline, but the order of the models does not change. For example, under wQL, Claude-Opus-4.8 (High) goes from 72.5\% to 63.1\% and TimesFM-3 from 61.6\% to 46.9\%. The same holds under WAPE.}

\begin{figure}[H]
  \centering
  \begin{minipage}{0.49\linewidth}\centering\includegraphics[width=\linewidth]{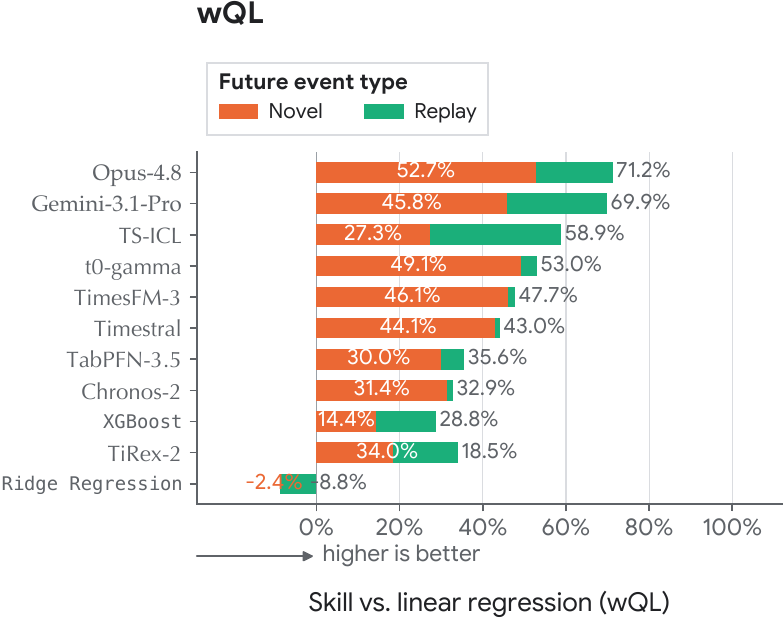}\end{minipage}\hfill
  \begin{minipage}{0.49\linewidth}\centering\includegraphics[width=\linewidth]{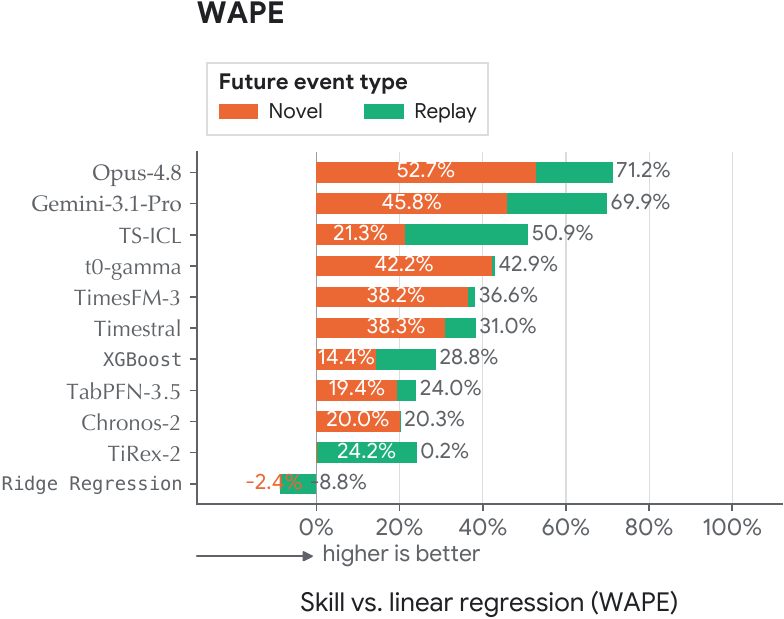}\end{minipage}
  \caption{Skill score against linear regression of \fw{palatino_llms}, \fw{optima_numerical} \fw{optima_foundation} \fw{optima_models}, and \fw{menlo_regression} \fw{menlo_models} on Replay and Novel futures, under wQL (left) and WAPE (right). The orange segment is a model's skill on Novel futures, and the full bar is its skill on Replay futures. Models are sorted by Replay skill.}
  \label{fig:vs-linear-replay-novel}
\end{figure}

\paragraph{Replay and Novel futures.}
\Cref{fig:vs-linear-replay-novel} shows the skill of each model on Replay and Novel futures against linear regression.

\takeaway{Linear regression itself drops sharply from Replay to Novel futures (from $49.0\%$ to $-6.7\%$ against the naive baseline). This is why the gap between Replay and Novel skill is much smaller against linear regression than against the naive baseline: the reference is much weaker on Novel futures than on Replay futures. For TimesFM-3, Timestral, Chronos-2, and TiRex-2, skill on Novel futures is similar to or higher than skill on Replay futures, and t0-gamma and TabPFN-3.5 lose only about $4$ to $6$ points, which shows that these models are more reliable than linear regression in novel scenarios. Against linear regression, the LLMs, TS-ICL, and XGBoost still have a lower skill on Novel futures than on Replay futures. Note that skill scores against linear regression are overall much smaller than against the naive baseline. The same holds under WAPE.}

\begin{figure}[H]
  \centering
  \includegraphics[width=0.95\linewidth]{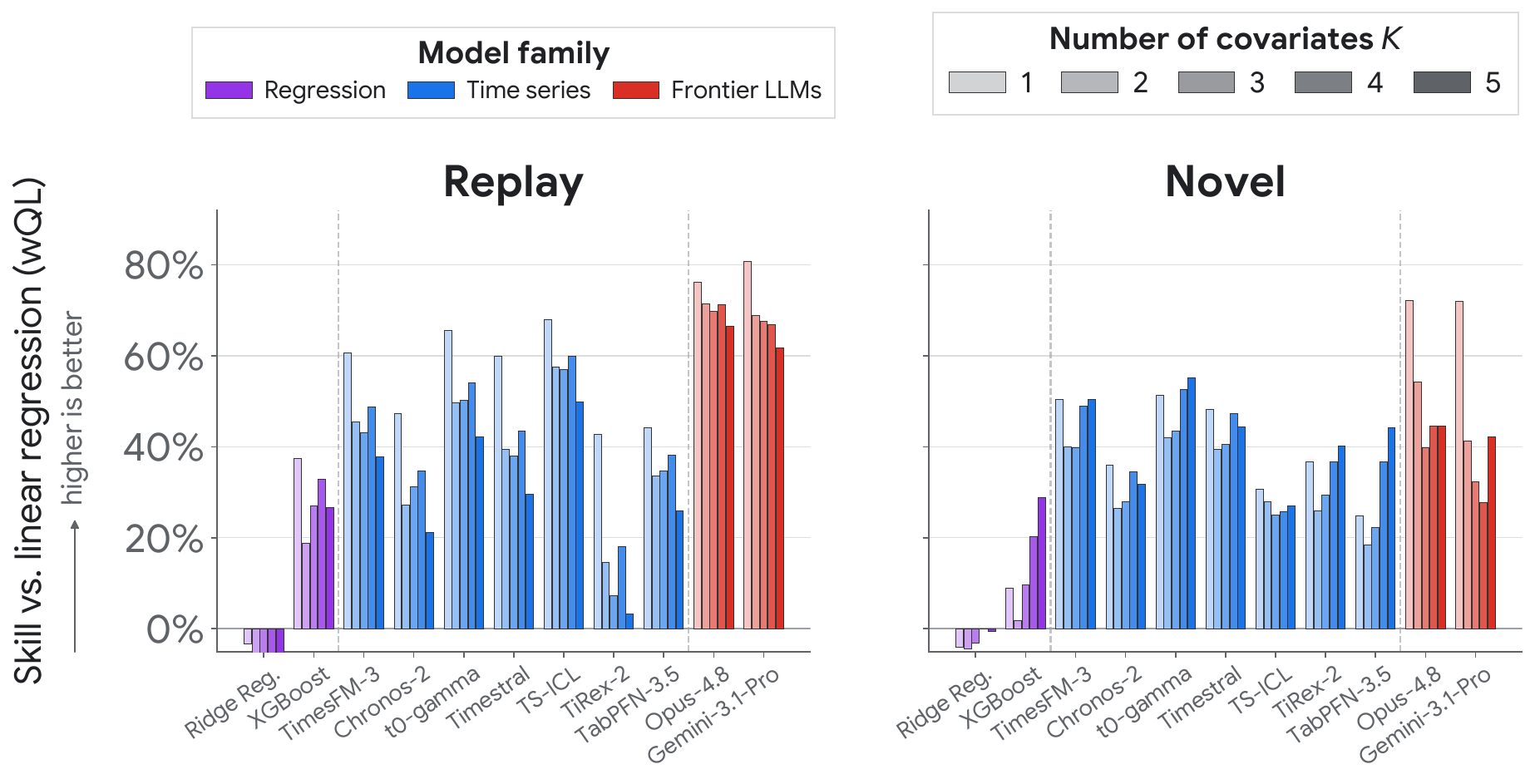}
  \caption{Skill score (wQL) against linear regression for different numbers of covariates $K$, on Replay (left) and Novel (right) futures.}
  \label{fig:vs-linear-by-k-wql}
\end{figure}

\begin{figure}[H]
  \centering
  \includegraphics[width=0.95\linewidth]{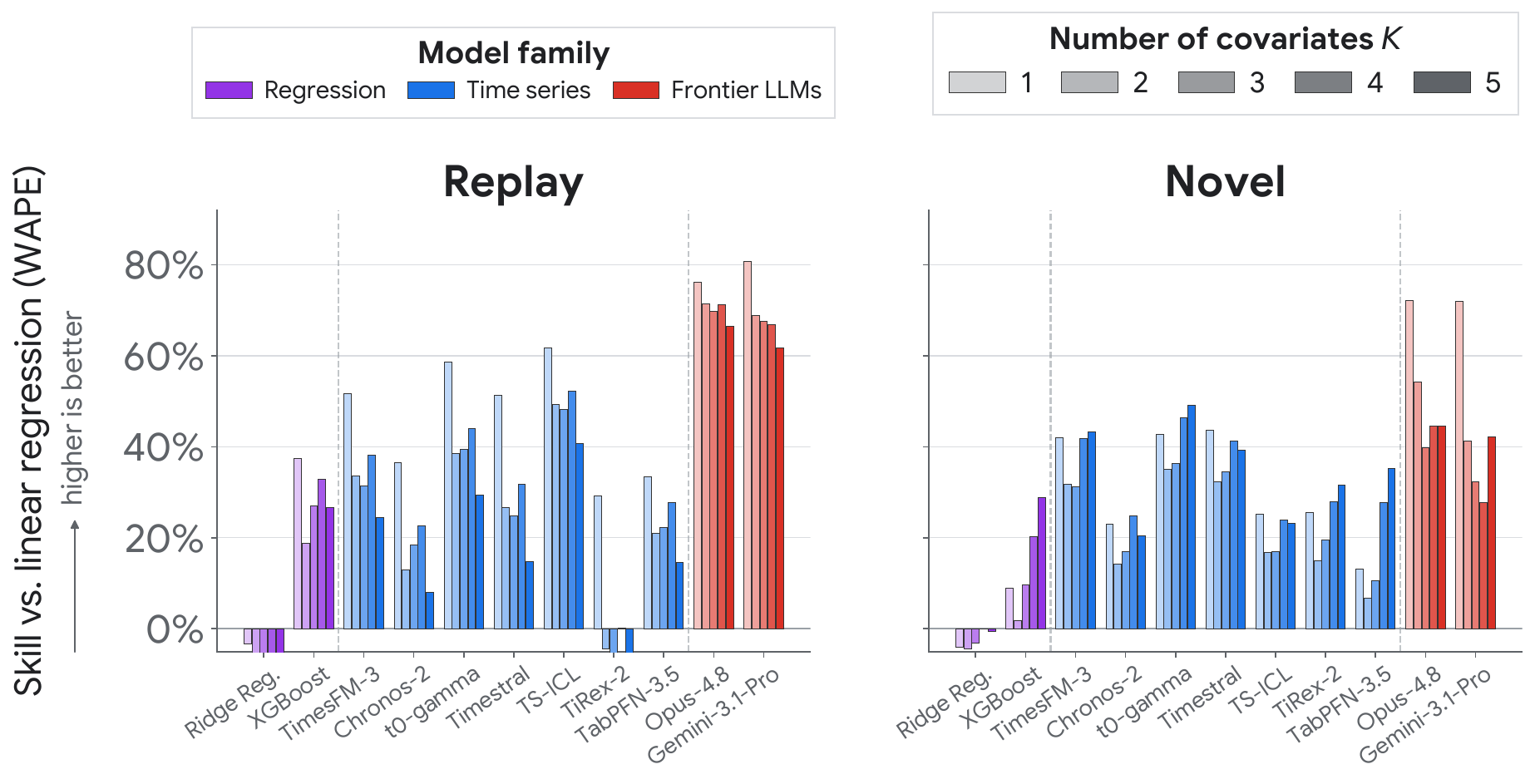}
  \caption{Skill score (WAPE) against linear regression for different numbers of covariates $K$, on Replay (left) and Novel (right) futures.}
  \label{fig:vs-linear-by-k-wape}
\end{figure}

\paragraph{Number of covariates.}
\Cref{fig:vs-linear-by-k-wql,fig:vs-linear-by-k-wape} show how skill varies with the number of covariates $K$, under wQL and WAPE.

\subsubsection*{wQL}

\subsubsection*{WAPE}

\takeaway{On Replay futures, skill against linear regression falls as covariates are added, whereas against the naive baseline it is nearly flat. This is because linear regression stays strong on Replay futures as covariates are added (its skill against the naive baseline is $37.3\%$ at $K=1$ and $56.2\%$ at $K=5$), while the models do not improve. On Novel futures, skill against the naive baseline falls sharply with more covariates. Against linear regression, it falls little or not at all for the foundation models, because linear regression itself falls as fast (from $12.1\%$ at $K=1$ to $-49.8\%$ at $K=5$ against the naive baseline). Against linear regression, the skill of the LLMs on Novel futures still falls as covariates are added. Skill scores against linear regression are overall much smaller than against the naive baseline. The same holds under WAPE.}

\begin{figure}[H]
  \centering
  \includegraphics[width=0.95\linewidth]{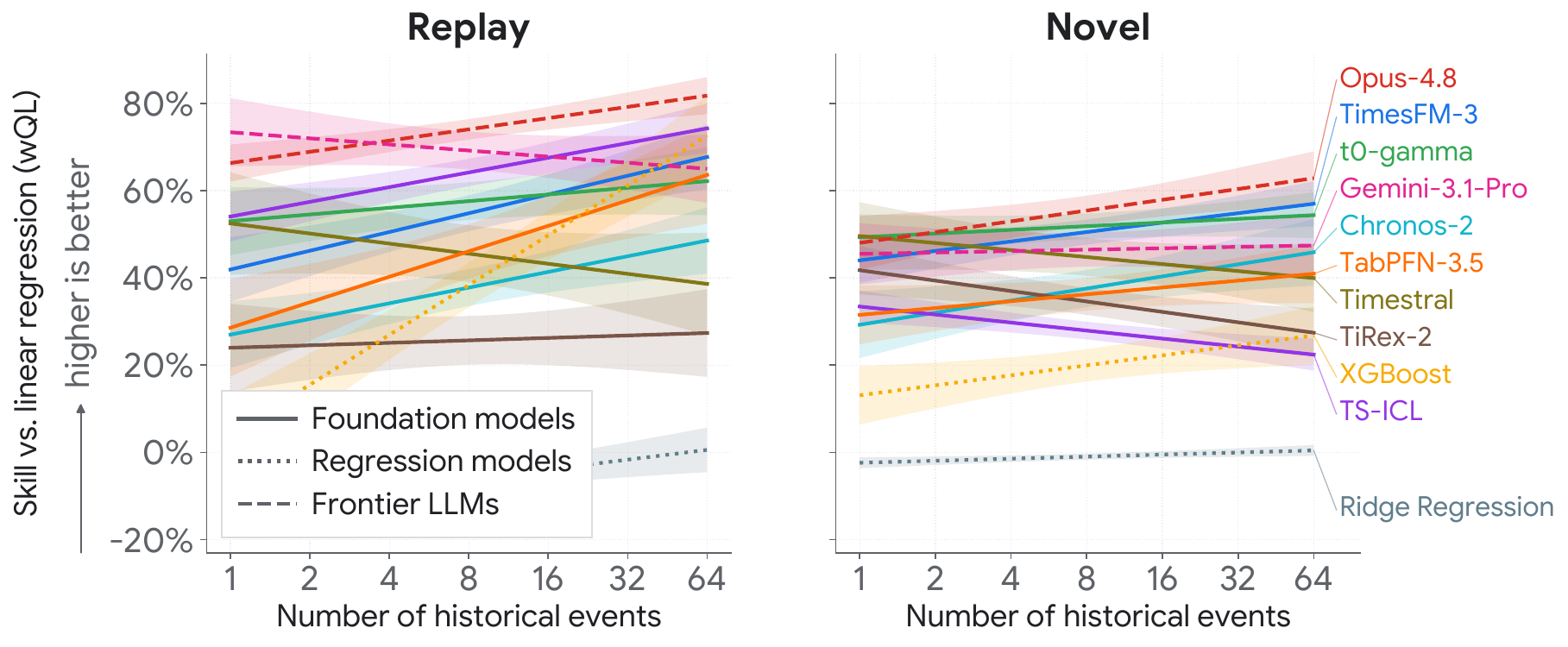}
  \caption{Skill score (wQL) against linear regression for different numbers of historical events, on Replay (left) and Novel (right) futures. Lines as in \Cref{fig:v10-kevents-pair-wql}.}
  \label{fig:vs-linear-kevents-wql}
\end{figure}

\begin{figure}[H]
  \centering
  \includegraphics[width=0.95\linewidth]{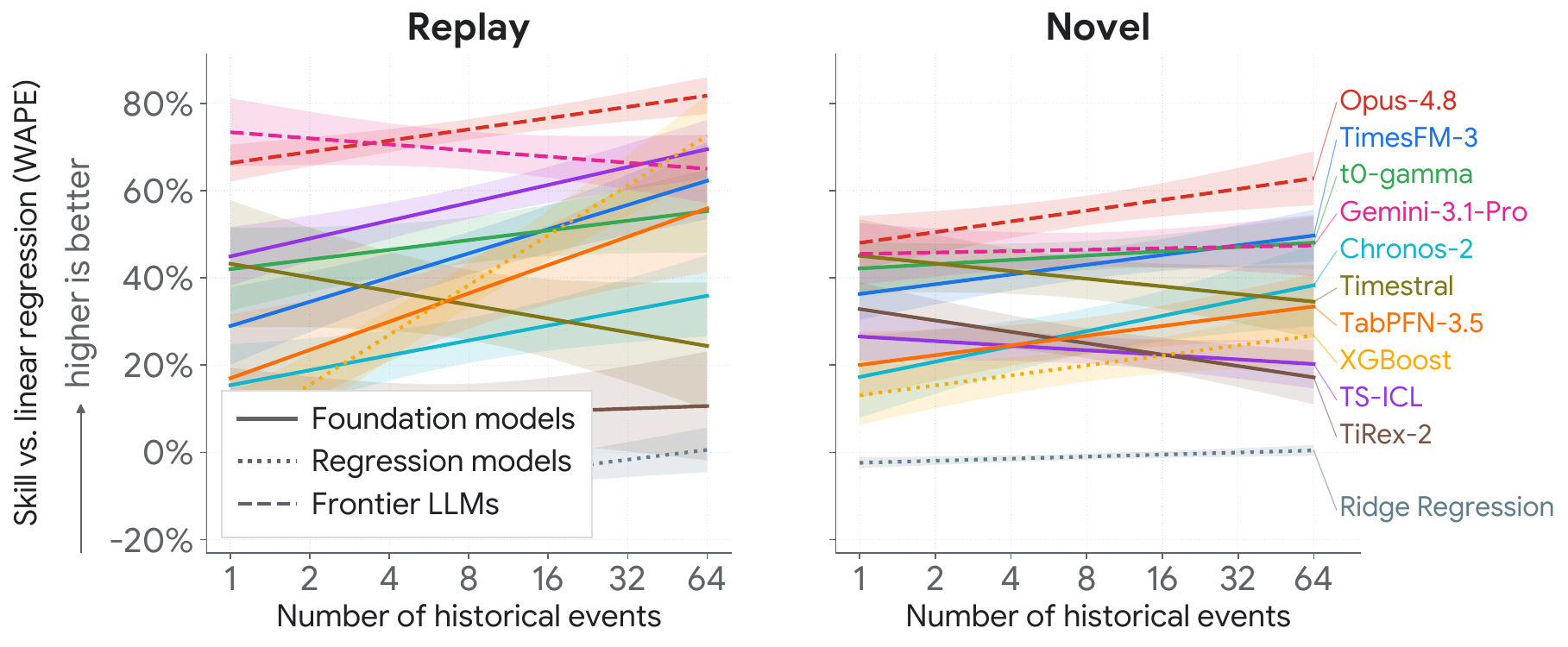}
  \caption{Skill score (WAPE) against linear regression for different numbers of historical events, on Replay (left) and Novel (right) futures. Lines as in \Cref{fig:v10-kevents-pair-wql}.}
  \label{fig:vs-linear-kevents-wape}
\end{figure}

\paragraph{Number of historical events.}
\Cref{fig:vs-linear-kevents-wql,fig:vs-linear-kevents-wape} show how skill varies with the number of historical events in the context, under wQL and WAPE.

\subsubsection*{wQL}

\subsubsection*{WAPE}

\takeaway{Against the naive baseline, skill on Replay futures is flat as the number of historical events grows, and skill on Novel futures rises. Against linear regression, more events raise skill on Replay futures for most models, and raise it less on Novel futures. Linear regression itself improves on Novel futures with more events (from $-27.3\%$ at one event to $9.3\%$ at 64, against the naive baseline), so most models gain only somewhat more from history than it does there. On Replay futures, the models gain more from history than linear regression does. The same holds under WAPE.}

\begin{figure}[H]
  \centering
  \includegraphics[width=0.95\linewidth]{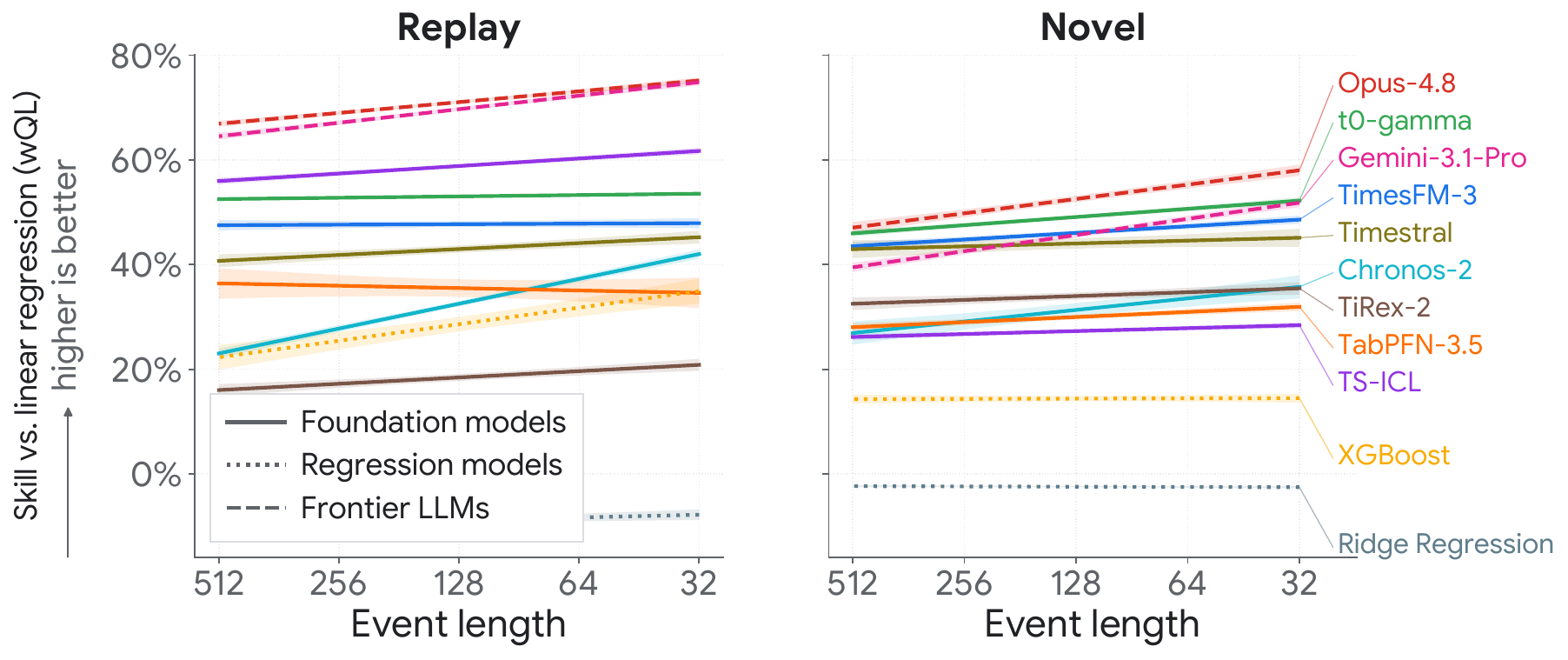}
  \caption{Skill score (wQL) against linear regression for different event lengths, on Replay (left) and Novel (right) futures. Lines as in \Cref{fig:v10-kevents-pair-wql}.}
  \label{fig:vs-linear-eventlen-wql}
\end{figure}

\begin{figure}[H]
  \centering
  \includegraphics[width=0.95\linewidth]{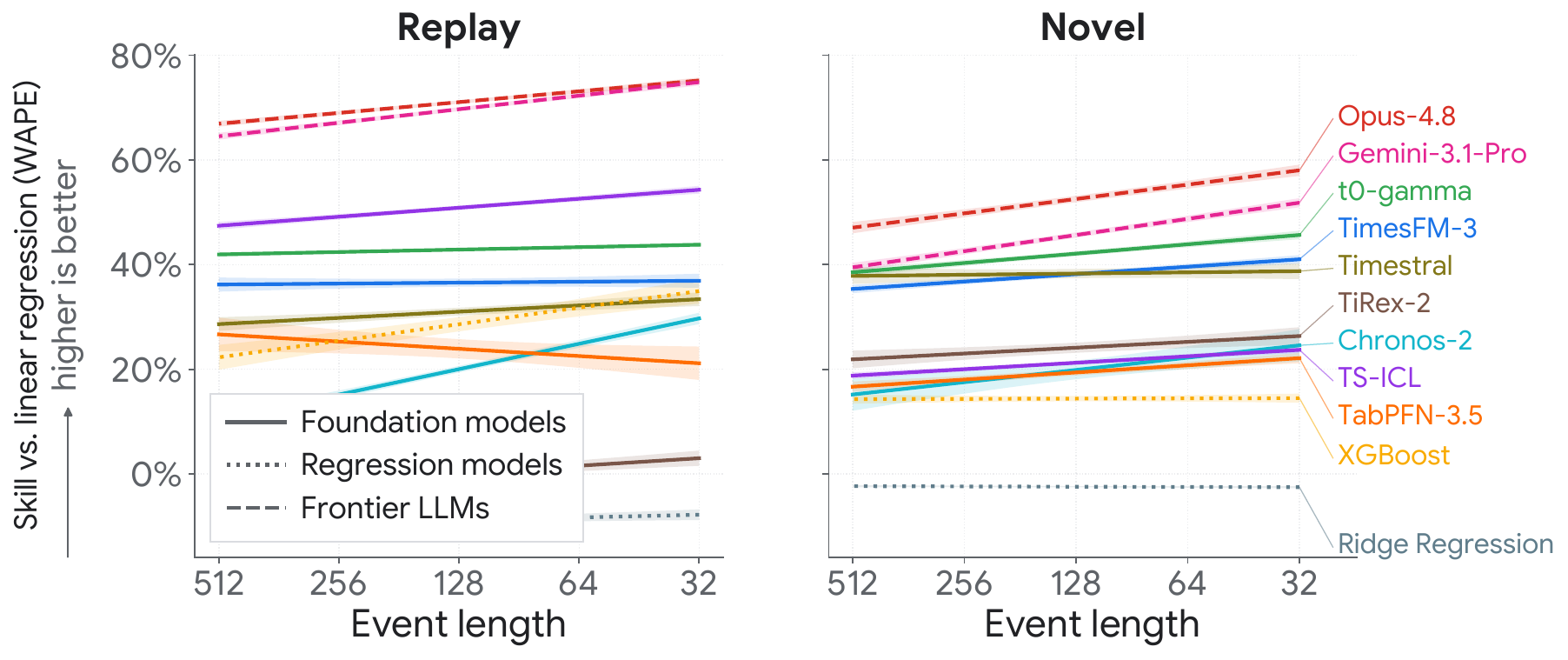}
  \caption{Skill score (WAPE) against linear regression for different event lengths, on Replay (left) and Novel (right) futures. Lines as in \Cref{fig:v10-kevents-pair-wql}.}
  \label{fig:vs-linear-eventlen-wape}
\end{figure}

\paragraph{Event length.}
\Cref{fig:vs-linear-eventlen-wql,fig:vs-linear-eventlen-wape} show how skill varies with the length of the events, under wQL and WAPE.

\subsubsection*{wQL}

\subsubsection*{WAPE}

\takeaway{Against the naive baseline, skill on Novel futures falls as events get longer. Linear regression itself falls only slightly on Novel futures with longer events (from $-4.1\%$ at length 32 to $-7.5\%$ at length 512, against the naive baseline). Against it, the skill of the models therefore still falls as events get longer, by about as much as against the naive baseline. Skill on Replay futures changes little with event length against the naive baseline, but falls slightly against linear regression, which itself improves on Replay futures with longer events. The same holds under WAPE.}

\paragraph{Other views.}
Our code supports further views that we do not show here, such as results for each simulator, and other aggregations and metrics, such as win rates. It can also compute every table and figure in this appendix against a different reference model.

\input{inputs/appendix_tuned_regression_draft.tex}

\section{Win Rates}
\label{app:win-rates}

\begin{figure}[H]
  \centering
  \includegraphics[width=\linewidth]{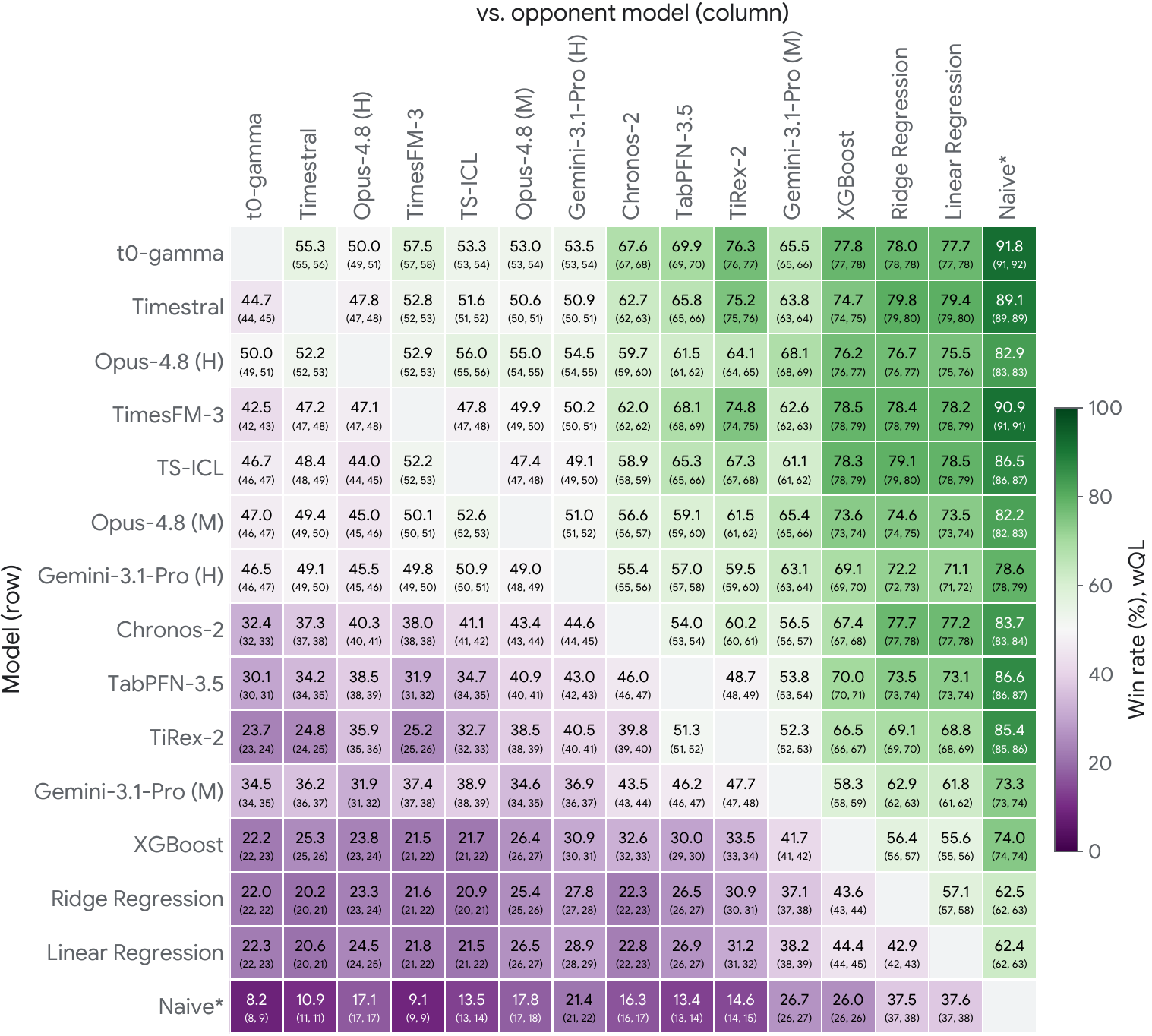}
  \caption{Pairwise win rates on \bench{} under the wQL metric. Each cell is the win rate (\%) of the row model against the column model, with a bootstrap 95\% confidence interval below it; the two cells across the diagonal add up to $100\%$. Models are ordered by their average win rate. Naive* does not consume covariates.}
  \label{fig:pairwise-winrate-wql}
\end{figure}

\begin{figure}[H]
  \centering
  \includegraphics[width=\linewidth]{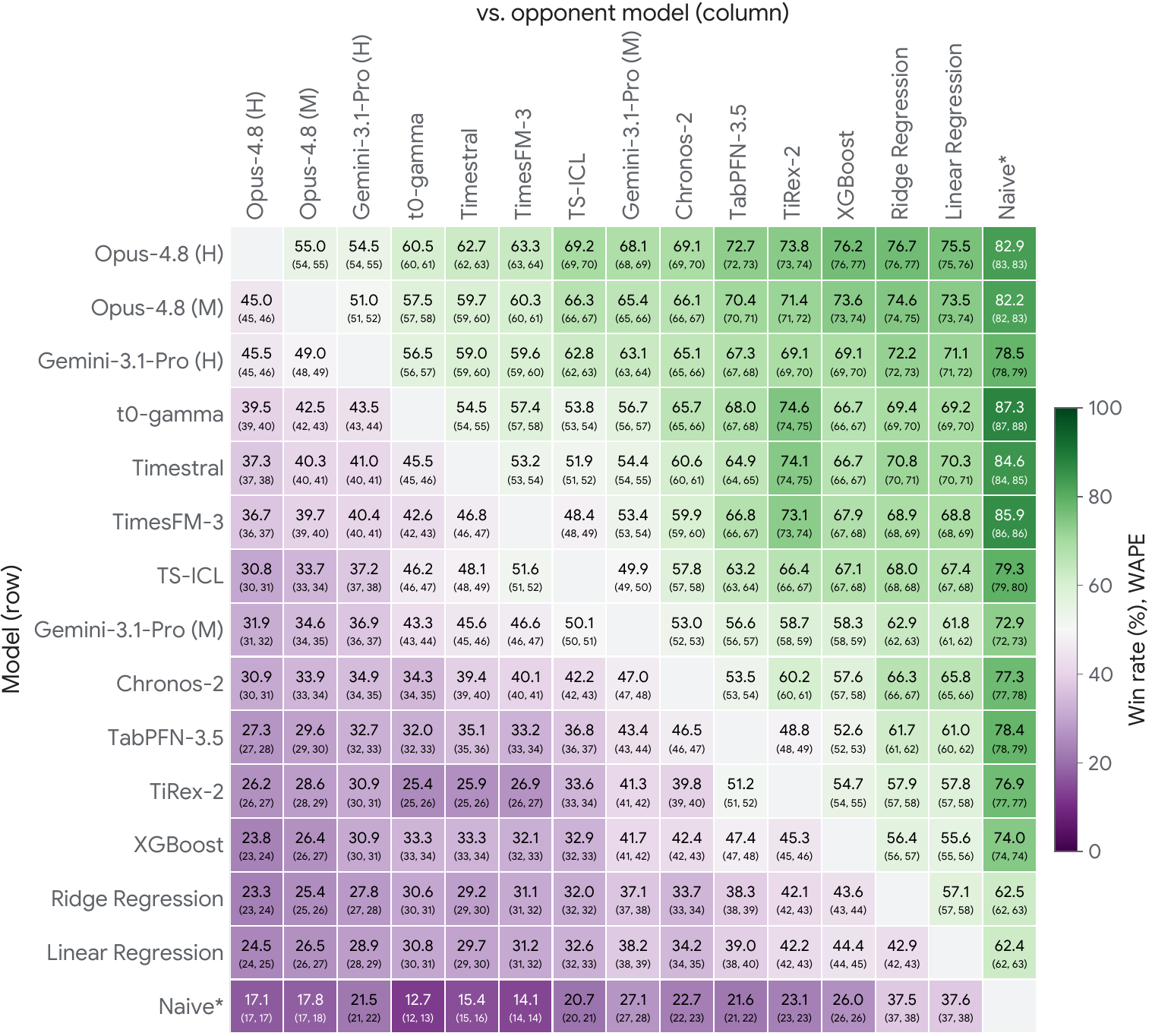}
  \caption{Pairwise win rates on \bench{} under the WAPE metric. Each cell is the win rate (\%) of the row model against the column model, with a bootstrap 95\% confidence interval below it; the two cells across the diagonal add up to $100\%$. Models are ordered by their average win rate. Naive* does not consume covariates.}
  \label{fig:pairwise-winrate-wape}
\end{figure}

The main text reports skill scores. Here we report \emph{win rates}: the fraction of task windows on which one model has a lower error than another, with ties counted as half a win. \Cref{tab:leaderboard-winrate} gives each model's average win rate over all other models. \Cref{fig:pairwise-winrate-wql,fig:pairwise-winrate-wape} show every pairwise comparison.

\input{tables/table_v10_winrate_paper.tex}

\clearpage  %

\clearpage  %

%% file: tables/table_linear_events.tex
\begin{table}[H]
\centering
\small
\setlength{\tabcolsep}{4pt}
\begin{tabular}{l rrrc rrrc}
\toprule
 & \multicolumn{4}{c}{\textbf{Replay}} & \multicolumn{4}{c}{\textbf{Novel}} \\
\cmidrule(lr){2-5} \cmidrule(lr){6-9}
\textbf{Simulator} & \textbf{Fewest} & \textbf{Most} & \textbf{Change} & \textbf{Worse} & \textbf{Fewest} & \textbf{Most} & \textbf{Change} & \textbf{Worse} \\
\midrule
OpenDC & $80.2\%$ & $54.4\%$ & $-25.8$ & 22/22 & $28.7\%$ & $-35.2\%$ & $-63.9$ & 11/22 \\
GridLAB-D & $34.8\%$ & $39.8\%$ & $+5.0$ & 4/10 & $6.2\%$ & $18.2\%$ & $+12.0$ & 4/10 \\
EPANET & $24.4\%$ & $39.4\%$ & $+15.1$ & 6/33$^\dagger$ & $10.9\%$ & $32.6\%$ & $+21.7$ & 10/33$^\dagger$ \\
SUMO & $79.8\%$ & $5.5\%$ & $-74.3$ & 25/29 & $-4.2\%$ & $-29.7\%$ & $-25.5$ & 22/29 \\
\bottomrule
\end{tabular}
\caption{Skill score (wQL) of Linear Regression with the fewest and the most historical events, on Replay and Novel futures. Each comparison uses tasks that share the target, event length, context length and number of covariates, where the most events are at least $8\times$ the fewest. Change is in percentage points. Worse is the number of targets whose skill is lower with the most events, out of the benchmark targets (\Cref{tab:instantiation}) that have at least one such comparison. $^\dagger$Out of the $37$ EPANET targets: the other $4$ (\texttt{demand\_9}, \texttt{flowrate\_10}, \texttt{flowrate\_9}, \texttt{velocity\_10}) have at most $2$ events in every task, so no such comparison exists for them. Ridge Regression is within $1$ point of Linear Regression in every cell.}
\label{tab:linear-events}
\end{table}

%% file: tables/table_tirex_events.tex
\begin{table}[H]
\centering
\small
\setlength{\tabcolsep}{4pt}
\begin{tabular}{l rrrc rrrc}
\toprule
 & \multicolumn{4}{c}{\textbf{Replay}} & \multicolumn{4}{c}{\textbf{Novel}} \\
\cmidrule(lr){2-5} \cmidrule(lr){6-9}
\textbf{Simulator} & \textbf{Fewest} & \textbf{Most} & \textbf{Change} & \textbf{Worse} & \textbf{Fewest} & \textbf{Most} & \textbf{Change} & \textbf{Worse} \\
\midrule
OpenDC & $63.3\%$ & $58.8\%$ & $-4.4$ & 17/22 & $23.1\%$ & $-24.9\%$ & $-48.1$ & 14/22 \\
GridLAB-D & $53.8\%$ & $55.7\%$ & $+1.9$ & 4/10 & $35.1\%$ & $37.3\%$ & $+2.2$ & 3/10 \\
EPANET & $40.9\%$ & $43.6\%$ & $+2.7$ & 15/33$^\dagger$ & $43.6\%$ & $47.7\%$ & $+4.0$ & 13/33$^\dagger$ \\
SUMO & $78.5\%$ & $55.3\%$ & $-23.2$ & 29/29 & $26.4\%$ & $18.5\%$ & $-7.9$ & 27/29 \\
\bottomrule
\end{tabular}
\caption{Skill score (wQL) of TiRex-2 with the fewest and the most historical events, on Replay and Novel futures. Each comparison uses tasks that share the target, event length, context length and number of covariates, where the most events are at least $8\times$ the fewest. Change is in percentage points. Worse is the number of targets whose skill is lower with the most events, out of the benchmark targets (\Cref{tab:instantiation}) that have at least one such comparison. $^\dagger$Out of the $37$ EPANET targets: the other $4$ (\texttt{demand\_9}, \texttt{flowrate\_10}, \texttt{flowrate\_9}, \texttt{velocity\_10}) have at most $2$ events in every task, so no such comparison exists for them.}
\label{tab:tirex-events}
\end{table}

%% file: tables/table_llm_context_events.tex
\begin{table}[H]
\centering
\footnotesize
\setlength{\tabcolsep}{3pt}
\begin{tabular}{l cc cc}
\toprule
 & \multicolumn{2}{c}{\textbf{Replay}} & \multicolumn{2}{c}{\textbf{Novel}} \\
\cmidrule(lr){2-3} \cmidrule(lr){4-5}
\textbf{Model} & \textbf{\shortstack{Context\\$C_3 \rightarrow C_4$}} & \textbf{\shortstack{Events\\fewest $\rightarrow$ most}} & \textbf{\shortstack{Context\\$C_3 \rightarrow C_4$}} & \textbf{\shortstack{Events\\fewest $\rightarrow$ most}} \\
\midrule
Gemini-3.1-Pro (High) & $87.2 \rightarrow 73.7$ \;($\mathbf{-13.5}$) & $82.9 \rightarrow 80.2$ \;($-2.7$) & $52.3 \rightarrow 35.1$ \;($\mathbf{-17.2}$) & $42.8 \rightarrow 49.6$ \;($+6.9$) \\
Gemini-3.1-Pro (Medium) & $75.9 \rightarrow 63.1$ \;($\mathbf{-12.9}$) & $73.0 \rightarrow 64.6$ \;($-8.4$) & $35.9 \rightarrow 24.2$ \;($\mathbf{-11.6}$) & $31.6 \rightarrow 31.1$ \;($-0.5$) \\
Claude-Opus-4.8 (High) & $86.6 \rightarrow 83.8$ \;($-2.8$) & $83.2 \rightarrow 85.5$ \;($+2.3$) & $57.8 \rightarrow 51.3$ \;($-6.5$) & $48.8 \rightarrow 60.7$ \;($+11.9$) \\
Claude-Opus-4.8 (Medium) & $85.5 \rightarrow 82.2$ \;($-3.3$) & $81.5 \rightarrow 84.0$ \;($+2.5$) & $53.7 \rightarrow 44.3$ \;($-9.4$) & $42.6 \rightarrow 55.1$ \;($+12.5$) \\
TimesFM-3 & $78.2 \rightarrow 76.4$ \;($-1.8$) & $73.5 \rightarrow 75.5$ \;($+2.0$) & $50.9 \rightarrow 44.6$ \;($-6.3$) & $45.2 \rightarrow 49.7$ \;($+4.6$) \\
\bottomrule
\end{tabular}
\caption{Skill score (wQL, \%) of the LLMs and TimesFM-3 when moving from the second-longest to the longest context ($C_3 \rightarrow C_4$), and from the fewest to the most historical events at a fixed target, event length, context length and number of covariates (most events at least $8\times$ the fewest). Change in percentage points in brackets. Gemini-3.1-Pro drops at the longest context, but changes little with the number of events.}
\label{tab:llm-context-events}
\end{table}

%% file: tables/table_v10_by_simulator.tex
\begin{table}[H]
\centering
\footnotesize
\setlength{\tabcolsep}{3pt}
\renewcommand{\arraystretch}{0.96}
\begin{tabular*}{\textwidth}{@{\extracolsep{\fill}}l l l l l l @{}}
\toprule
\textbf{\shortstack[l]{Model\\ (\# Tasks)}} & \textbf{\shortstack{OpenDC\\ (8,800)}} & \textbf{\shortstack{GridLAB-D\\ (4,000)}} & \textbf{\shortstack{EPANET\\ (13,265)}} & \textbf{\shortstack{SUMO\\ (11,600)}} & \textbf{\shortstack{All\\ (37,665)}} \\
\midrule
\multicolumn{6}{@{}l}{\textbf{\textit{Frontier LLMs}}} \\
\quad Claude-Opus-4.8 (High) & \makebox[2.9em][r]{{\setlength{\fboxsep}{1pt}\colorbox{black!14}{\textbf{75.1\%}}}}\,\textcolor{blue!55!black}{\rule{5.26mm}{1.1ex}} & \makebox[2.9em][r]{{\setlength{\fboxsep}{1pt}\colorbox{black!14}{\textbf{58.7\%}}}}\,\textcolor{blue!55!black}{\rule{4.11mm}{1.1ex}} & \makebox[2.9em][r]{45.9\%}\,\textcolor{blue!55!black}{\rule{3.21mm}{1.1ex}} & \makebox[2.9em][r]{{\setlength{\fboxsep}{1pt}\colorbox{black!14}{\textbf{88.1\%}}}}\,\textcolor{blue!55!black}{\rule{6.17mm}{1.1ex}} & \makebox[2.9em][r]{{\setlength{\fboxsep}{1pt}\colorbox{black!14}{\textbf{72.5\%}}}}\,\textcolor{black!35}{\rule{5.08mm}{1.1ex}} \\
\quad Claude-Opus-4.8 (Medium) & \makebox[2.9em][r]{\underline{73.4\%}}\,\textcolor{blue!55!black}{\rule{5.14mm}{1.1ex}} & \makebox[2.9em][r]{56.2\%}\,\textcolor{blue!55!black}{\rule{3.93mm}{1.1ex}} & \makebox[2.9em][r]{41.4\%}\,\textcolor{blue!55!black}{\rule{2.90mm}{1.1ex}} & \makebox[2.9em][r]{\underline{87.0\%}}\,\textcolor{blue!55!black}{\rule{6.09mm}{1.1ex}} & \makebox[2.9em][r]{\underline{70.3\%}}\,\textcolor{black!35}{\rule{4.92mm}{1.1ex}} \\
\quad Gemini-3.1-Pro (High) & \makebox[2.9em][r]{73.0\%}\,\textcolor{blue!55!black}{\rule{5.11mm}{1.1ex}} & \makebox[2.9em][r]{52.4\%}\,\textcolor{blue!55!black}{\rule{3.67mm}{1.1ex}} & \makebox[2.9em][r]{52.0\%}\,\textcolor{blue!55!black}{\rule{3.64mm}{1.1ex}} & \makebox[2.9em][r]{83.7\%}\,\textcolor{blue!55!black}{\rule{5.86mm}{1.1ex}} & \makebox[2.9em][r]{70.0\%}\,\textcolor{black!35}{\rule{4.90mm}{1.1ex}} \\
\quad Gemini-3.1-Pro (Medium) & \makebox[2.9em][r]{62.9\%}\,\textcolor{blue!55!black}{\rule{4.40mm}{1.1ex}} & \makebox[2.9em][r]{35.0\%}\,\textcolor{blue!55!black}{\rule{2.45mm}{1.1ex}} & \makebox[2.9em][r]{25.7\%}\,\textcolor{blue!55!black}{\rule{1.80mm}{1.1ex}} & \makebox[2.9em][r]{79.6\%}\,\textcolor{blue!55!black}{\rule{5.57mm}{1.1ex}} & \makebox[2.9em][r]{58.2\%}\,\textcolor{black!35}{\rule{4.07mm}{1.1ex}} \\
\addlinespace[0.5ex]
\multicolumn{6}{@{}l}{\textbf{\textit{Numerical Foundation Models}}} \\
\quad t0-gamma & \makebox[2.9em][r]{58.6\%}\,\textcolor{blue!55!black}{\rule{4.10mm}{1.1ex}} & \makebox[2.9em][r]{\underline{58.4\%}}\,\textcolor{blue!55!black}{\rule{4.09mm}{1.1ex}} & \makebox[2.9em][r]{{\setlength{\fboxsep}{1pt}\colorbox{black!14}{\textbf{59.1\%}}}}\,\textcolor{blue!55!black}{\rule{4.14mm}{1.1ex}} & \makebox[2.9em][r]{74.6\%}\,\textcolor{blue!55!black}{\rule{5.22mm}{1.1ex}} & \makebox[2.9em][r]{64.5\%}\,\textcolor{black!35}{\rule{4.52mm}{1.1ex}} \\
\quad TimesFM-3 & \makebox[2.9em][r]{56.0\%}\,\textcolor{blue!55!black}{\rule{3.92mm}{1.1ex}} & \makebox[2.9em][r]{54.8\%}\,\textcolor{blue!55!black}{\rule{3.84mm}{1.1ex}} & \makebox[2.9em][r]{\underline{57.9\%}}\,\textcolor{blue!55!black}{\rule{4.05mm}{1.1ex}} & \makebox[2.9em][r]{70.4\%}\,\textcolor{blue!55!black}{\rule{4.93mm}{1.1ex}} & \makebox[2.9em][r]{61.6\%}\,\textcolor{black!35}{\rule{4.31mm}{1.1ex}} \\
\quad TS-ICL & \makebox[2.9em][r]{61.1\%}\,\textcolor{blue!55!black}{\rule{4.28mm}{1.1ex}} & \makebox[2.9em][r]{54.2\%}\,\textcolor{blue!55!black}{\rule{3.79mm}{1.1ex}} & \makebox[2.9em][r]{42.1\%}\,\textcolor{blue!55!black}{\rule{2.95mm}{1.1ex}} & \makebox[2.9em][r]{77.0\%}\,\textcolor{blue!55!black}{\rule{5.39mm}{1.1ex}} & \makebox[2.9em][r]{61.3\%}\,\textcolor{black!35}{\rule{4.29mm}{1.1ex}} \\
\quad Timestral & \makebox[2.9em][r]{47.2\%}\,\textcolor{blue!55!black}{\rule{3.30mm}{1.1ex}} & \makebox[2.9em][r]{55.1\%}\,\textcolor{blue!55!black}{\rule{3.86mm}{1.1ex}} & \makebox[2.9em][r]{56.4\%}\,\textcolor{blue!55!black}{\rule{3.95mm}{1.1ex}} & \makebox[2.9em][r]{69.8\%}\,\textcolor{blue!55!black}{\rule{4.89mm}{1.1ex}} & \makebox[2.9em][r]{59.2\%}\,\textcolor{black!35}{\rule{4.14mm}{1.1ex}} \\
\quad TabPFN-3.5 & \makebox[2.9em][r]{59.7\%}\,\textcolor{blue!55!black}{\rule{4.18mm}{1.1ex}} & \makebox[2.9em][r]{48.3\%}\,\textcolor{blue!55!black}{\rule{3.38mm}{1.1ex}} & \makebox[2.9em][r]{37.6\%}\,\textcolor{blue!55!black}{\rule{2.63mm}{1.1ex}} & \makebox[2.9em][r]{57.1\%}\,\textcolor{blue!55!black}{\rule{4.00mm}{1.1ex}} & \makebox[2.9em][r]{50.8\%}\,\textcolor{black!35}{\rule{3.56mm}{1.1ex}} \\
\quad Chronos-2 & \makebox[2.9em][r]{44.4\%}\,\textcolor{blue!55!black}{\rule{3.11mm}{1.1ex}} & \makebox[2.9em][r]{50.1\%}\,\textcolor{blue!55!black}{\rule{3.51mm}{1.1ex}} & \makebox[2.9em][r]{43.7\%}\,\textcolor{blue!55!black}{\rule{3.06mm}{1.1ex}} & \makebox[2.9em][r]{61.8\%}\,\textcolor{blue!55!black}{\rule{4.33mm}{1.1ex}} & \makebox[2.9em][r]{50.8\%}\,\textcolor{black!35}{\rule{3.56mm}{1.1ex}} \\
\quad TiRex-2 & \makebox[2.9em][r]{40.9\%}\,\textcolor{blue!55!black}{\rule{2.86mm}{1.1ex}} & \makebox[2.9em][r]{45.6\%}\,\textcolor{blue!55!black}{\rule{3.19mm}{1.1ex}} & \makebox[2.9em][r]{41.4\%}\,\textcolor{blue!55!black}{\rule{2.90mm}{1.1ex}} & \makebox[2.9em][r]{56.0\%}\,\textcolor{blue!55!black}{\rule{3.92mm}{1.1ex}} & \makebox[2.9em][r]{46.7\%}\,\textcolor{black!35}{\rule{3.27mm}{1.1ex}} \\
\addlinespace[0.5ex]
\multicolumn{6}{@{}l}{\textbf{\textit{Regression Models$^\dagger$}}} \\
\quad XGBoost & \makebox[2.9em][r]{58.2\%}\,\textcolor{blue!55!black}{\rule{4.07mm}{1.1ex}} & \makebox[2.9em][r]{21.4\%}\,\textcolor{blue!55!black}{\rule{1.50mm}{1.1ex}} & \makebox[2.9em][r]{4.6\%}\,\textcolor{blue!55!black}{\rule{0.32mm}{1.1ex}} & \makebox[2.9em][r]{63.9\%}\,\textcolor{blue!55!black}{\rule{4.47mm}{1.1ex}} & \makebox[2.9em][r]{42.9\%}\,\textcolor{black!35}{\rule{3.00mm}{1.1ex}} \\
\quad Linear Regression & \makebox[2.9em][r]{48.8\%}\,\textcolor{blue!55!black}{\rule{3.42mm}{1.1ex}} & \makebox[2.9em][r]{22.4\%}\,\textcolor{blue!55!black}{\rule{1.57mm}{1.1ex}} & \makebox[2.9em][r]{-1.7\%}\,\textcolor{blue!55!black}{\rule{0.00mm}{1.1ex}} & \makebox[2.9em][r]{33.7\%}\,\textcolor{blue!55!black}{\rule{2.36mm}{1.1ex}} & \makebox[2.9em][r]{26.2\%}\,\textcolor{black!35}{\rule{1.83mm}{1.1ex}} \\
\quad Ridge Regression & \makebox[2.9em][r]{48.3\%}\,\textcolor{blue!55!black}{\rule{3.38mm}{1.1ex}} & \makebox[2.9em][r]{22.8\%}\,\textcolor{blue!55!black}{\rule{1.60mm}{1.1ex}} & \makebox[2.9em][r]{-1.7\%}\,\textcolor{blue!55!black}{\rule{0.00mm}{1.1ex}} & \makebox[2.9em][r]{31.7\%}\,\textcolor{blue!55!black}{\rule{2.22mm}{1.1ex}} & \makebox[2.9em][r]{25.4\%}\,\textcolor{black!35}{\rule{1.78mm}{1.1ex}} \\
\addlinespace[0.5ex]
\multicolumn{6}{@{}l}{\textbf{\textit{Baseline}}} \\
\quad Naive* & \makebox[2.9em][r]{0.0\%}\,\textcolor{blue!55!black}{\rule{0.00mm}{1.1ex}} & \makebox[2.9em][r]{0.0\%}\,\textcolor{blue!55!black}{\rule{0.00mm}{1.1ex}} & \makebox[2.9em][r]{0.0\%}\,\textcolor{blue!55!black}{\rule{0.00mm}{1.1ex}} & \makebox[2.9em][r]{0.0\%}\,\textcolor{blue!55!black}{\rule{0.00mm}{1.1ex}} & \makebox[2.9em][r]{0.0\%}\,\textcolor{black!35}{\rule{0.00mm}{1.1ex}} \\
\bottomrule
\end{tabular*}
\caption{Skill score (wQL) of each model on \bench{}, separately for each of the four simulators, and pooled over all of them (All). The number under each simulator name is its number of tasks (\# Tasks). The best score per column is {\setlength{\fboxsep}{1pt}\colorbox{black!14}{\textbf{bolded and highlighted}}} and the second-best is \underline{underlined}. Note that Naive* represents a naive baseline which does not consume covariates. All other models are evaluated with covariates. $^\dagger$The simplest, untuned versions of the regression models; tuned variants are in \Cref{app:tuned-regression}.}
\label{tab:leaderboard-by-simulator}
\end{table}

%% file: tables/table_v10_vs_linear.tex
\begin{table}[H]
\centering
\footnotesize
\setlength{\tabcolsep}{3pt}
\renewcommand{\arraystretch}{0.96}
\begin{tabular*}{\textwidth}{@{\extracolsep{\fill}}l llll@{}}
\toprule
 & \multicolumn{2}{c}{\textbf{wQL}} & \multicolumn{2}{c}{\textbf{WAPE}} \\
\cmidrule(lr){2-3} \cmidrule(lr){4-5}
\textbf{Model} & \textbf{vs. Naive} & \textbf{vs. Linear} & \textbf{vs. Naive} & \textbf{vs. Linear} \\
\midrule
\multicolumn{5}{@{}l}{\textbf{\textit{Frontier LLMs}}} \\
\quad Claude-Opus-4.8 (High) & \makebox[3.0em][r]{{\setlength{\fboxsep}{1pt}\colorbox{black!14}{\textbf{72.5\%}}}}\,\textcolor{black!35}{\rule{5.08mm}{1.1ex}} & \makebox[3.0em][r]{{\setlength{\fboxsep}{1pt}\colorbox{black!14}{\textbf{63.1\%}}}}\,\textcolor{blue!55!black}{\rule{4.42mm}{1.1ex}} & \makebox[3.0em][r]{{\setlength{\fboxsep}{1pt}\colorbox{black!14}{\textbf{72.5\%}}}}\,\textcolor{black!35}{\rule{5.08mm}{1.1ex}} & \makebox[3.0em][r]{{\setlength{\fboxsep}{1pt}\colorbox{black!14}{\textbf{63.1\%}}}}\,\textcolor{blue!55!black}{\rule{4.42mm}{1.1ex}} \\
\quad Claude-Opus-4.8 (Medium) & \makebox[3.0em][r]{\underline{70.3\%}}\,\textcolor{black!35}{\rule{4.92mm}{1.1ex}} & \makebox[3.0em][r]{\underline{60.1\%}}\,\textcolor{blue!55!black}{\rule{4.21mm}{1.1ex}} & \makebox[3.0em][r]{\underline{70.3\%}}\,\textcolor{black!35}{\rule{4.92mm}{1.1ex}} & \makebox[3.0em][r]{\underline{60.1\%}}\,\textcolor{blue!55!black}{\rule{4.21mm}{1.1ex}} \\
\quad Gemini-3.1-Pro (High) & \makebox[3.0em][r]{70.0\%}\,\textcolor{black!35}{\rule{4.90mm}{1.1ex}} & \makebox[3.0em][r]{59.6\%}\,\textcolor{blue!55!black}{\rule{4.17mm}{1.1ex}} & \makebox[3.0em][r]{70.0\%}\,\textcolor{black!35}{\rule{4.90mm}{1.1ex}} & \makebox[3.0em][r]{59.6\%}\,\textcolor{blue!55!black}{\rule{4.17mm}{1.1ex}} \\
\quad Gemini-3.1-Pro (Medium) & \makebox[3.0em][r]{58.2\%}\,\textcolor{black!35}{\rule{4.07mm}{1.1ex}} & \makebox[3.0em][r]{43.4\%}\,\textcolor{blue!55!black}{\rule{3.04mm}{1.1ex}} & \makebox[3.0em][r]{58.2\%}\,\textcolor{black!35}{\rule{4.07mm}{1.1ex}} & \makebox[3.0em][r]{43.4\%}\,\textcolor{blue!55!black}{\rule{3.04mm}{1.1ex}} \\
\addlinespace[0.5ex]
\multicolumn{5}{@{}l}{\textbf{\textit{Numerical Foundation Models}}} \\
\quad t0-gamma & \makebox[3.0em][r]{64.5\%}\,\textcolor{black!35}{\rule{4.52mm}{1.1ex}} & \makebox[3.0em][r]{51.1\%}\,\textcolor{blue!55!black}{\rule{3.58mm}{1.1ex}} & \makebox[3.0em][r]{58.4\%}\,\textcolor{black!35}{\rule{4.09mm}{1.1ex}} & \makebox[3.0em][r]{42.5\%}\,\textcolor{blue!55!black}{\rule{2.98mm}{1.1ex}} \\
\quad TimesFM-3 & \makebox[3.0em][r]{61.6\%}\,\textcolor{black!35}{\rule{4.31mm}{1.1ex}} & \makebox[3.0em][r]{46.9\%}\,\textcolor{blue!55!black}{\rule{3.28mm}{1.1ex}} & \makebox[3.0em][r]{54.5\%}\,\textcolor{black!35}{\rule{3.82mm}{1.1ex}} & \makebox[3.0em][r]{37.4\%}\,\textcolor{blue!55!black}{\rule{2.62mm}{1.1ex}} \\
\quad TS-ICL & \makebox[3.0em][r]{61.3\%}\,\textcolor{black!35}{\rule{4.29mm}{1.1ex}} & \makebox[3.0em][r]{45.3\%}\,\textcolor{blue!55!black}{\rule{3.17mm}{1.1ex}} & \makebox[3.0em][r]{55.3\%}\,\textcolor{black!35}{\rule{3.87mm}{1.1ex}} & \makebox[3.0em][r]{37.8\%}\,\textcolor{blue!55!black}{\rule{2.65mm}{1.1ex}} \\
\quad Timestral & \makebox[3.0em][r]{59.2\%}\,\textcolor{black!35}{\rule{4.14mm}{1.1ex}} & \makebox[3.0em][r]{43.5\%}\,\textcolor{blue!55!black}{\rule{3.05mm}{1.1ex}} & \makebox[3.0em][r]{52.7\%}\,\textcolor{black!35}{\rule{3.69mm}{1.1ex}} & \makebox[3.0em][r]{34.8\%}\,\textcolor{blue!55!black}{\rule{2.44mm}{1.1ex}} \\
\quad TabPFN-3.5 & \makebox[3.0em][r]{50.8\%}\,\textcolor{black!35}{\rule{3.56mm}{1.1ex}} & \makebox[3.0em][r]{32.8\%}\,\textcolor{blue!55!black}{\rule{2.30mm}{1.1ex}} & \makebox[3.0em][r]{42.3\%}\,\textcolor{black!35}{\rule{2.96mm}{1.1ex}} & \makebox[3.0em][r]{21.7\%}\,\textcolor{blue!55!black}{\rule{1.52mm}{1.1ex}} \\
\quad Chronos-2 & \makebox[3.0em][r]{50.8\%}\,\textcolor{black!35}{\rule{3.56mm}{1.1ex}} & \makebox[3.0em][r]{32.2\%}\,\textcolor{blue!55!black}{\rule{2.25mm}{1.1ex}} & \makebox[3.0em][r]{42.0\%}\,\textcolor{black!35}{\rule{2.94mm}{1.1ex}} & \makebox[3.0em][r]{20.2\%}\,\textcolor{blue!55!black}{\rule{1.41mm}{1.1ex}} \\
\quad TiRex-2 & \makebox[3.0em][r]{46.7\%}\,\textcolor{black!35}{\rule{3.27mm}{1.1ex}} & \makebox[3.0em][r]{26.7\%}\,\textcolor{blue!55!black}{\rule{1.87mm}{1.1ex}} & \makebox[3.0em][r]{36.5\%}\,\textcolor{black!35}{\rule{2.56mm}{1.1ex}} & \makebox[3.0em][r]{13.0\%}\,\textcolor{blue!55!black}{\rule{0.91mm}{1.1ex}} \\
\addlinespace[0.5ex]
\multicolumn{5}{@{}l}{\textbf{\textit{Regression Models$^\dagger$}}} \\
\quad XGBoost & \makebox[3.0em][r]{42.9\%}\,\textcolor{black!35}{\rule{3.00mm}{1.1ex}} & \makebox[3.0em][r]{21.9\%}\,\textcolor{blue!55!black}{\rule{1.53mm}{1.1ex}} & \makebox[3.0em][r]{42.9\%}\,\textcolor{black!35}{\rule{3.00mm}{1.1ex}} & \makebox[3.0em][r]{21.9\%}\,\textcolor{blue!55!black}{\rule{1.53mm}{1.1ex}} \\
\quad Linear Regression & \makebox[3.0em][r]{26.2\%}\,\textcolor{black!35}{\rule{1.83mm}{1.1ex}} & \makebox[3.0em][r]{0.0\%}\,\textcolor{blue!55!black}{\rule{0.00mm}{1.1ex}} & \makebox[3.0em][r]{26.2\%}\,\textcolor{black!35}{\rule{1.83mm}{1.1ex}} & \makebox[3.0em][r]{0.0\%}\,\textcolor{blue!55!black}{\rule{0.00mm}{1.1ex}} \\
\quad Ridge Regression & \makebox[3.0em][r]{25.4\%}\,\textcolor{black!35}{\rule{1.78mm}{1.1ex}} & \makebox[3.0em][r]{-5.6\%}\,\textcolor{blue!55!black}{\rule{0.00mm}{1.1ex}} & \makebox[3.0em][r]{25.4\%}\,\textcolor{black!35}{\rule{1.78mm}{1.1ex}} & \makebox[3.0em][r]{-5.6\%}\,\textcolor{blue!55!black}{\rule{0.00mm}{1.1ex}} \\
\addlinespace[0.5ex]
\multicolumn{5}{@{}l}{\textbf{\textit{Baseline}}} \\
\quad Naive* & \makebox[3.0em][r]{0.0\%}\,\textcolor{black!35}{\rule{0.00mm}{1.1ex}} & \makebox[3.0em][r]{-35.5\%}\,\textcolor{blue!55!black}{\rule{0.00mm}{1.1ex}} & \makebox[3.0em][r]{0.0\%}\,\textcolor{black!35}{\rule{0.00mm}{1.1ex}} & \makebox[3.0em][r]{-35.5\%}\,\textcolor{blue!55!black}{\rule{0.00mm}{1.1ex}} \\
\bottomrule
\end{tabular*}
\caption{Skill score of each model on \bench{} under wQL and WAPE, measured against two reference models: the naive baseline that copies the last value (as in the main text) and a covariate-aware linear regression (Linear Regression). Each reference has a skill of 0 by construction. The best score per column is {\setlength{\fboxsep}{1pt}\colorbox{black!14}{\textbf{bolded and highlighted}}} and the second-best is \underline{underlined}. Note that Naive* represents a naive baseline which does not consume covariates. All other models are evaluated with covariates. $^\dagger$The simplest, untuned versions of the regression models; tuned variants are in \Cref{app:tuned-regression}.}
\label{tab:leaderboard-vs-linear}
\end{table}

%% file: inputs/appendix_tuned_regression_draft.tex
\section{Tuned Regression Baselines}
\label{app:tuned-regression}

The regression baselines in the main text are kept as simple as possible: each one is fit on the history of a task with one training row per time step, predicting the target from the covariates at that step, with default settings. Here we explore whether a minimal amount of tuning improves them, to check whether our conclusions depend on how weak these baselines are. \textbf{Tuning improves their scores, but does not change any of our conclusions.} We tune three things: how the training rows are built, which features the models use, and how many past values of the target they see (and, for Ridge Regression, its regularization strength). We select the variants on the test set, so we do not consider them valid baselines; this is also why we use the plain models for the main benchmark. The tuning is also not exhaustive: we do not tune other model hyperparameters, such as the learning rate or the tree depth of XGBoost. Even so, it gives a sense of how much these simple baselines can be improved.

\paragraph{Paired training.} With one training row per time step, the regression models cannot directly use past values of the target: a feature such as the target one step earlier is unknown at test time for every forecast step after the first, unless forecasts are fed back as inputs. We instead take every feature relative to a \emph{forecast origin}: the last time step at which the target is known when the forecast is made, which is $t$ at test time. We design the training rows as pairs of an origin $\tau$ and a horizon $h$, where each row predicts the target $X_{\tau+h}$. We sample the origins from the history, as N-BEATS does to create its training examples~\citep{oreshkin2020nbeats}. For each task, we draw a horizon $h$ uniformly from $1$ to the prediction length $T-t$, and then an origin $\tau$ uniformly from the history, such that $\tau+h\le t$ and enough history precedes $\tau$ to compute the features. We draw up to $8{,}000$ such pairs per task. At test time, the origin is $\tau=t$, and the model predicts each step $X_{t+h}$, $h=1,\dots,T-t$, separately, using the known future covariates.

\paragraph{Variants.} We vary the features and the number of target lags. For predicting $X_{t+h}$ from the origin $t$, the features are:
\begin{itemize}[noitemsep, leftmargin=*]
  \item $\zb_{t+h}$: the covariates at the step being predicted (always used, and the only feature of the plain models);
  \item $\zb_{t+h-1}$: the covariates one step earlier;
  \item $\zb_t$: the covariates at the origin;
  \item $\Delta\zb = \zb_{t+h}-\zb_t$: the change in the covariates since the origin;
  \item $s_t$: the mean, standard deviation, and slope of the last $32$ target values up to the origin, $X_{t-31},\dots,X_t$;
  \item $h$: the horizon.
\end{itemize}
In training, the sampled origin $\tau$ takes the place of $t$. Lags $\le \ell$ are the values of the target at lags $1, 2, 3, 6, 12, 24, 48, 96, 128, 256$ up to $\ell$, where lag $L$ is $X_{t-L+1}$. We explored $34$ variants across the three models. \Cref{tab:tuned-regression} shows a selection that adds features and lags step by step, as a guide for tuning these models, together with the best variant of each model.

\input{tables/table_tuned_regression.tex}

\takeaway{Minimal tuning raises the skill of the regression baselines by $7$ to $9$ points. Paired training alone does not help; the gains come from the features it allows, mainly the recent values of the target, as lags or as a summary. Even tuned, XGBoost only reaches TabPFN-3.5 and Chronos-2, and stays well below t0-gamma, TimesFM-3, TS-ICL, Timestral, and the LLMs. The drop from Replay to Novel futures stays as large as for the plain models.}

\begin{figure}[H]
  \centering
  \includegraphics[width=0.6\linewidth]{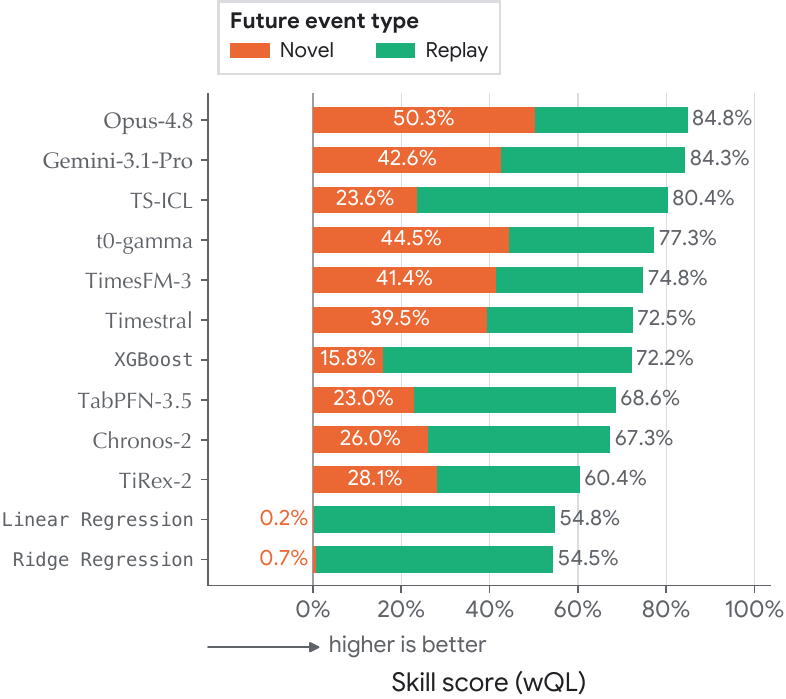}
  \caption{Skill score (wQL) on Replay and Novel futures, with the tuned variants of Linear Regression, Ridge Regression, and XGBoost (best on the test set) in place of the plain ones. The orange segment is a model's skill on Novel futures, and the full bar is its skill on Replay futures. Models are sorted by Replay skill.}
  \label{fig:tuned-replay-novel}
\end{figure}

\paragraph{Replay and Novel futures.} \Cref{fig:tuned-replay-novel} shows the best variant of each model on Replay and Novel futures, next to the other models in the paper.

\takeaway{Tuning raises the regression baselines on both Replay and Novel futures, but they still drop sharply between the two: tuned XGBoost goes from $72.2\%$ to $15.8\%$, and tuned Linear Regression from $54.8\%$ to $0.2\%$. Tuned XGBoost is ahead of several foundation models on Replay futures, but below every foundation model on Novel futures.}

\begin{figure}[H]
  \centering
  \includegraphics[width=\linewidth]{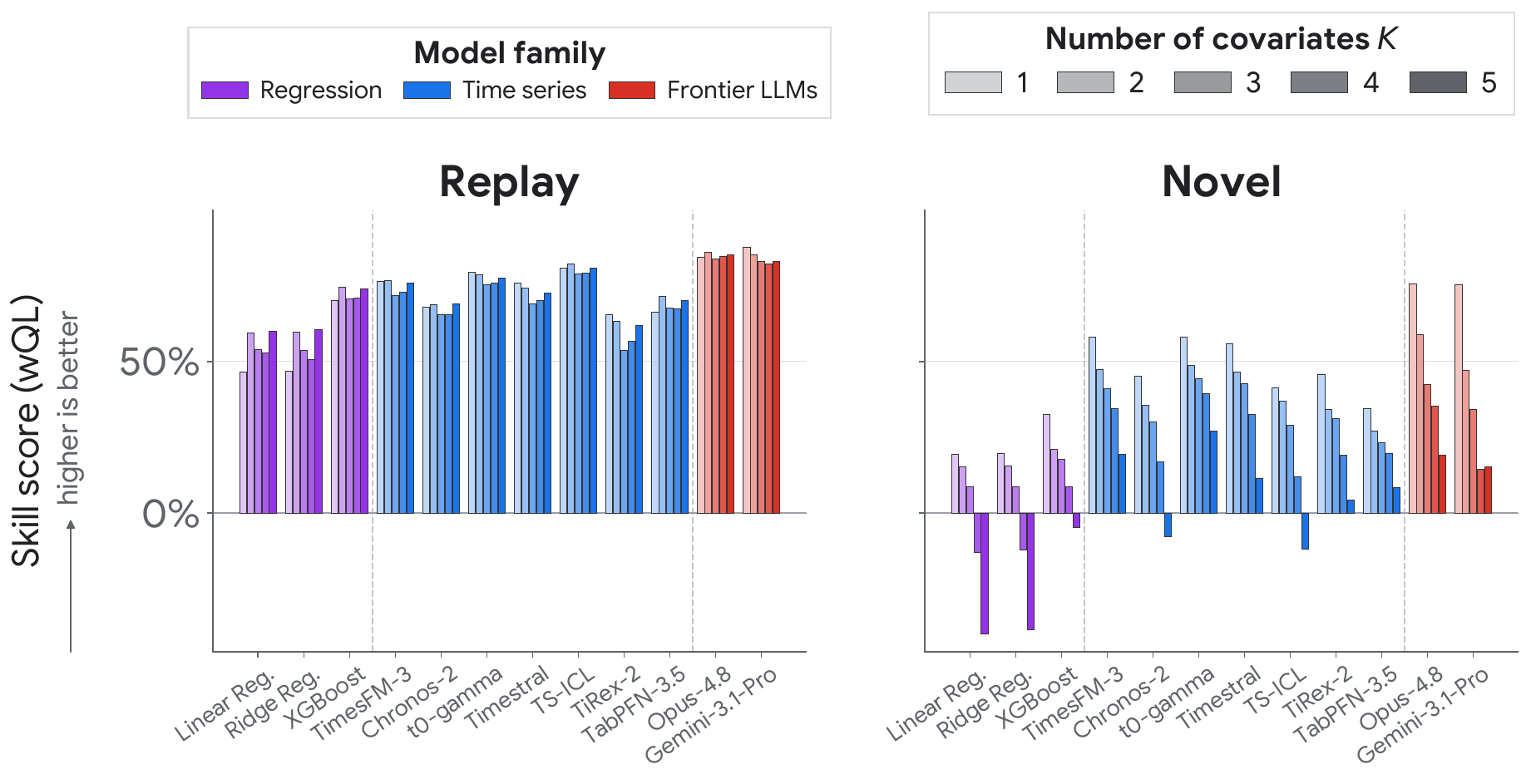}
  \caption{Skill score (wQL) for different numbers of covariates, on Replay (left) and Novel (right) futures, with the tuned regression variants.}
  \label{fig:tuned-by-k}
\end{figure}

\begin{figure}[H]
  \centering
  \includegraphics[width=\linewidth]{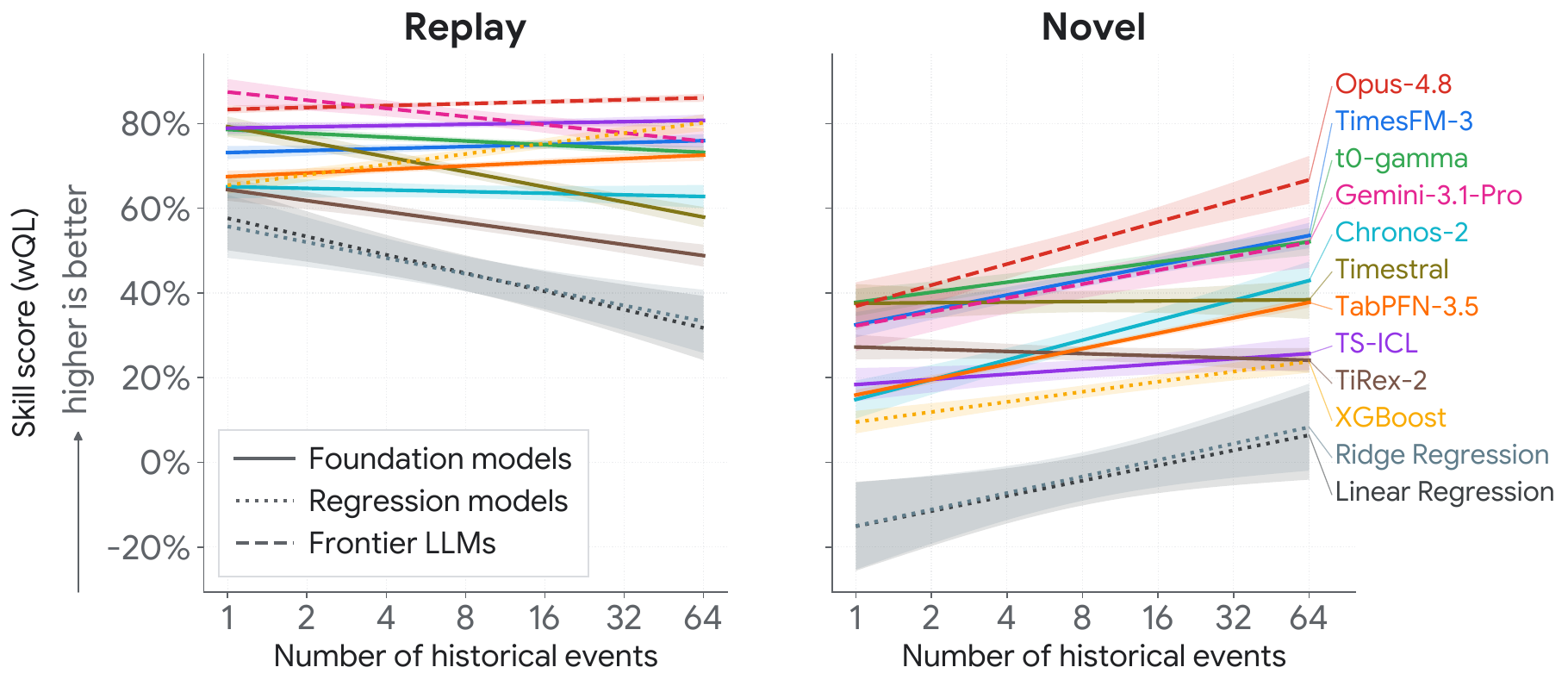}
  \caption{Skill score (wQL) for different numbers of historical events, on Replay (left) and Novel (right) futures, with the tuned regression variants. Lines as in \Cref{fig:v10-kevents-pair-wql}.}
  \label{fig:tuned-kevents}
\end{figure}

\begin{figure}[H]
  \centering
  \includegraphics[width=\linewidth]{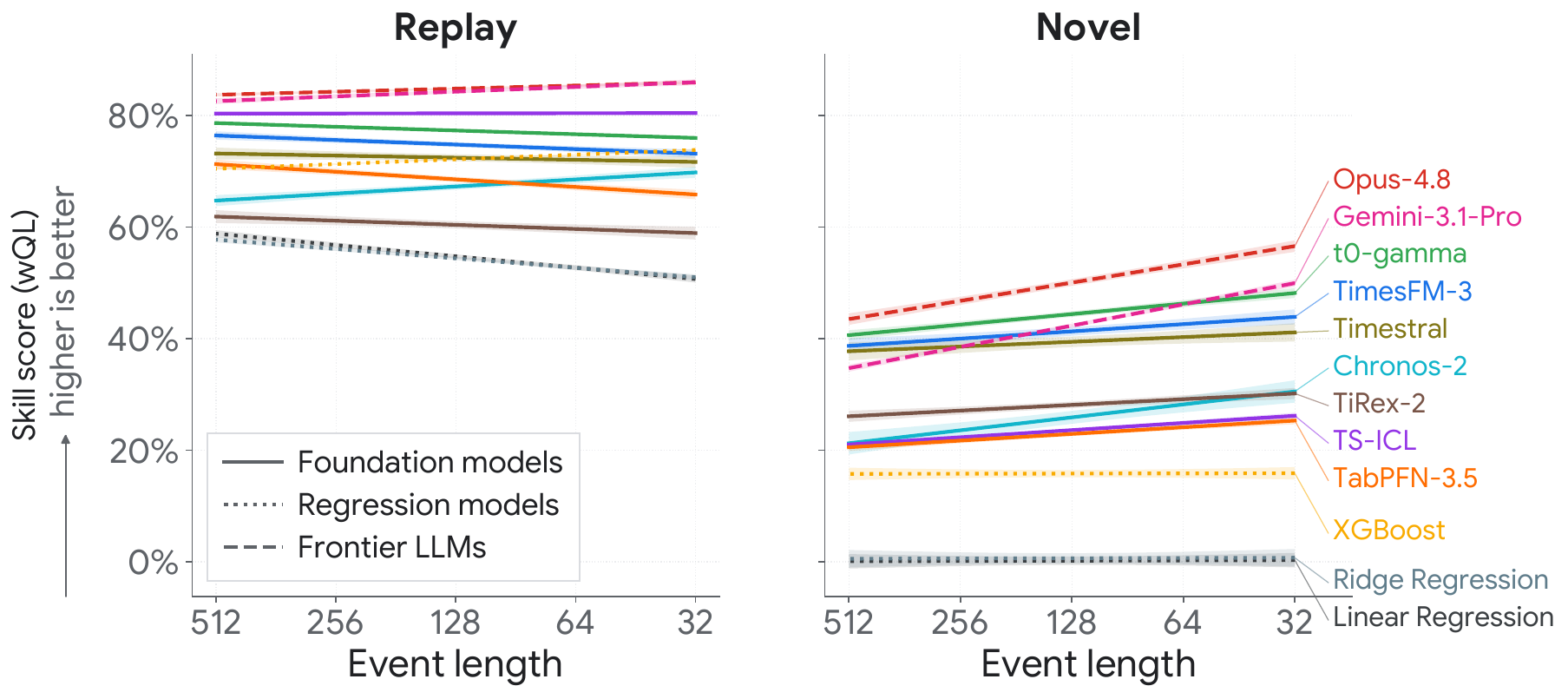}
  \caption{Skill score (wQL) for different event lengths, on Replay (left) and Novel (right) futures, with the tuned regression variants. Lines as in \Cref{fig:v10-kevents-pair-wql}.}
  \label{fig:tuned-eventlen}
\end{figure}

\begin{figure}[H]
  \centering
  \includegraphics[width=\linewidth]{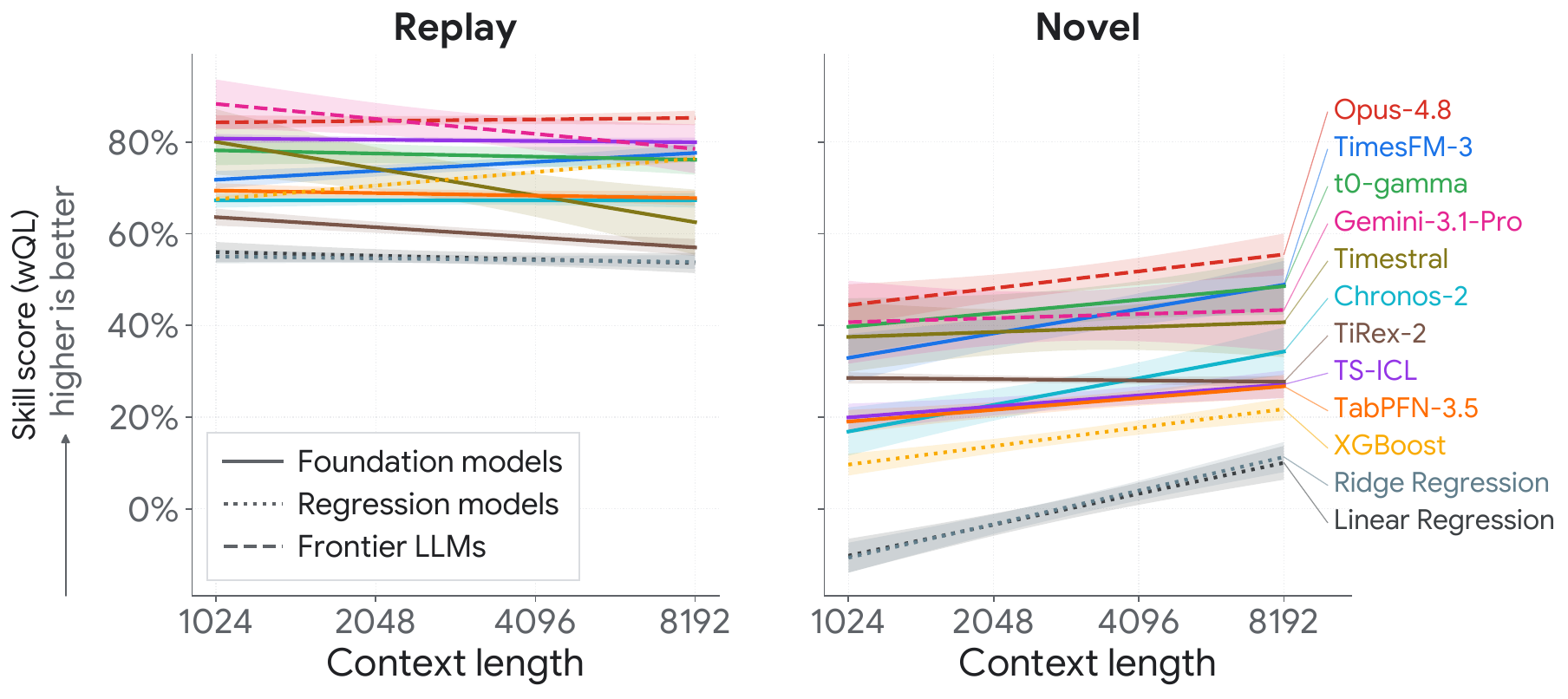}
  \caption{Skill score (wQL) for different context lengths, on Replay (left) and Novel (right) futures, with the tuned regression variants. Lines as in \Cref{fig:v10-kevents-pair-wql}.}
  \label{fig:tuned-ctxlen}
\end{figure}

\paragraph{Other axes.} \Cref{fig:tuned-by-k,fig:tuned-kevents,fig:tuned-eventlen,fig:tuned-ctxlen} show the tuned variants along the number of covariates, the number of historical events, the event length, and the context length.

\takeaway{The tuned variants follow the same trends as the plain ones. On Novel futures, their skill falls as more covariates are set together, rises with more historical events and a longer context, and changes little with the event length. Along every axis, they stay well below the best models on Novel futures.}

%% file: tables/table_tuned_regression.tex
\begin{table}[H]
\centering
\small
\begin{tabular}{lll rrrr}
\toprule
\textbf{Training} & \textbf{Features} & \textbf{Lags} & \textbf{Skill} & \textbf{Replay} & \textbf{Novel} & \textbf{Gap} \\
\midrule
\multicolumn{7}{@{}l}{\textbf{\textit{Linear Regression}}} \\
\quad Plain (main text) & $\zb_{t+h}$ & --- & 26.2\% & 49.0\% & -6.7\% & 55.7 \\
\quad Paired & $\zb_{t+h}$ & --- & 25.8\% & 48.7\% & -7.1\% & 55.8 \\
\quad Paired & $\zb_{t+h}, s_t$ & --- & 30.7\% & 52.3\% & -0.6\% & 52.9 \\
\quad \textbf{Paired} & $\boldsymbol{\zb_{t+h}, \zb_t, s_t, h}$ & \textbf{---} & \textbf{32.9\%} & \textbf{54.8\%} & \textbf{0.2\%} & \textbf{54.6} \\
\quad Paired & all & --- & 28.9\% & 52.5\% & -6.5\% & 59.0 \\
\quad Paired & all & $\le 3$ & 29.5\% & 53.1\% & -5.7\% & 58.8 \\
\quad Paired & all & $\le 24$ & 29.4\% & 53.0\% & -5.8\% & 58.8 \\
\midrule
\multicolumn{7}{@{}l}{\textbf{\textit{Ridge Regression}}} \\
\quad Plain (main text) & $\zb_{t+h}$ & --- & 25.4\% & 47.5\% & -6.1\% & 53.6 \\
\quad Paired & $\zb_{t+h}$ & --- & 25.4\% & 47.9\% & -6.8\% & 54.7 \\
\quad Paired & $\zb_{t+h}$ & 1 & 30.5\% & 51.5\% & 0.5\% & 51.0 \\
\quad Paired & $\zb_{t+h}$ & $\le 24$ & 30.6\% & 51.6\% & 0.6\% & 50.9 \\
\quad Paired & $\zb_{t+h}$ & $\le 48$ & 30.8\% & 51.7\% & 0.9\% & 50.7 \\
\quad \textbf{Paired} & \textbf{all} & $\boldsymbol{\le 48}$ & \textbf{32.7\%} & \textbf{54.5\%} & \textbf{0.7\%} & \textbf{53.8} \\
\midrule
\multicolumn{7}{@{}l}{\textbf{\textit{XGBoost}}} \\
\quad Plain (main text) & $\zb_{t+h}$ & --- & 42.9\% & 65.0\% & 6.7\% & 58.3 \\
\quad Paired & $\zb_{t+h}$ & --- & 42.5\% & 64.7\% & 6.4\% & 58.3 \\
\quad Paired & $\zb_{t+h}$ & 1 & 45.8\% & 67.0\% & 11.1\% & 56.0 \\
\quad Paired & $\zb_{t+h}$ & $\le 24$ & 46.7\% & 68.4\% & 10.1\% & 58.3 \\
\quad Paired & $\zb_{t+h}$ & $\le 48$ & 47.2\% & 68.7\% & 11.2\% & 57.5 \\
\quad Paired & all & $\le 24$ & 50.4\% & 71.4\% & 13.9\% & 57.5 \\
\quad Paired & all & $\le 48$ & 50.8\% & 71.8\% & 14.2\% & 57.6 \\
\quad \textbf{Paired} & \textbf{all} & $\boldsymbol{\le 96}$ & \textbf{51.6\%} & \textbf{72.2\%} & \textbf{15.8\%} & \textbf{56.3} \\
\quad Paired & all & $\le 128$ & 51.4\% & 72.0\% & 15.7\% & 56.3 \\
\quad Paired & all & $\le 256$ & 50.7\% & 71.4\% & 15.0\% & 56.4 \\
\bottomrule
\end{tabular}
\caption{Skill score (wQL) of the tuned regression variants, measured against the naive baseline. Rows add features and target lags step by step; the best variant of each model is in bold. \emph{all} is $\zb_{t+h}, \zb_{t+h-1}, \zb_t, \Delta\zb, s_t, h$. Ridge Regression uses a regularization strength of $10$ in every row shown. Gap is Replay skill minus Novel skill, in percentage points. The variants are selected on the test set and are not valid baselines.}
\label{tab:tuned-regression}
\end{table}

%% file: tables/table_v10_winrate_paper.tex
\begin{table}[H]
\centering
\small
\renewcommand{\arraystretch}{0.96}
\vspace{-1.5mm}
\begin{tabular*}{\textwidth}{@{\extracolsep{\fill}}l ll c@{}}
\toprule
\textbf{Model} & \textbf{wQL Win Rate} & \textbf{WAPE Win Rate} & \textbf{Inf. time / series (s)} \\
\midrule
\multicolumn{4}{@{}l}{\textbf{\textit{Frontier LLMs}}} \\
\quad Claude-Opus-4.8 (High) & \makebox[3.4em][r]{63.2\%}\,\textcolor{blue!55!black}{\rule{12.64mm}{1.3ex}} & \makebox[3.4em][r]{\textbf{68.6\%}}\,\textcolor{black!35}{\rule{13.72mm}{1.3ex}} & 270.13 \\
\quad Claude-Opus-4.8 (Medium) & \makebox[3.4em][r]{60.1\%}\,\textcolor{blue!55!black}{\rule{12.02mm}{1.3ex}} & \makebox[3.4em][r]{\underline{65.5\%}}\,\textcolor{black!35}{\rule{13.10mm}{1.3ex}} & 206.29 \\
\quad Gemini-3.1-Pro (High) & \makebox[3.4em][r]{58.4\%}\,\textcolor{blue!55!black}{\rule{11.68mm}{1.3ex}} & \makebox[3.4em][r]{63.4\%}\,\textcolor{black!35}{\rule{12.68mm}{1.3ex}} & 355.28 \\
\quad Gemini-3.1-Pro (Medium) & \makebox[3.4em][r]{46.0\%}\,\textcolor{blue!55!black}{\rule{9.20mm}{1.3ex}} & \makebox[3.4em][r]{50.9\%}\,\textcolor{black!35}{\rule{10.18mm}{1.3ex}} & 97.35 \\
\addlinespace[0.5ex]
\multicolumn{4}{@{}l}{\textbf{\textit{Numerical Foundation Models}}} \\
\quad t0-gamma & \makebox[3.4em][r]{\textbf{66.2\%}}\,\textcolor{blue!55!black}{\rule{13.24mm}{1.3ex}} & \makebox[3.4em][r]{60.6\%}\,\textcolor{black!35}{\rule{12.12mm}{1.3ex}} & 0.043 \\
\quad Timestral & \makebox[3.4em][r]{\underline{63.5\%}}\,\textcolor{blue!55!black}{\rule{12.70mm}{1.3ex}} & \makebox[3.4em][r]{58.3\%}\,\textcolor{black!35}{\rule{11.66mm}{1.3ex}} & 0.048 \\
\quad TimesFM-3 & \makebox[3.4em][r]{62.7\%}\,\textcolor{blue!55!black}{\rule{12.54mm}{1.3ex}} & \makebox[3.4em][r]{57.1\%}\,\textcolor{black!35}{\rule{11.42mm}{1.3ex}} & 0.419 \\
\quad TS-ICL & \makebox[3.4em][r]{61.6\%}\,\textcolor{blue!55!black}{\rule{12.32mm}{1.3ex}} & \makebox[3.4em][r]{54.8\%}\,\textcolor{black!35}{\rule{10.96mm}{1.3ex}} & 0.083 \\
\quad Chronos-2 & \makebox[3.4em][r]{53.8\%}\,\textcolor{blue!55!black}{\rule{10.76mm}{1.3ex}} & \makebox[3.4em][r]{48.8\%}\,\textcolor{black!35}{\rule{9.76mm}{1.3ex}} & 0.067 \\
\quad TabPFN-3.5 & \makebox[3.4em][r]{50.3\%}\,\textcolor{blue!55!black}{\rule{10.06mm}{1.3ex}} & \makebox[3.4em][r]{44.2\%}\,\textcolor{black!35}{\rule{8.84mm}{1.3ex}} & 21.26 \\
\quad TiRex-2 & \makebox[3.4em][r]{46.7\%}\,\textcolor{blue!55!black}{\rule{9.34mm}{1.3ex}} & \makebox[3.4em][r]{41.2\%}\,\textcolor{black!35}{\rule{8.24mm}{1.3ex}} & 0.079 \\
\addlinespace[0.5ex]
\multicolumn{4}{@{}l}{\textbf{\textit{Regression Models$^\dagger$}}} \\
\quad XGBoost & \makebox[3.4em][r]{35.4\%}\,\textcolor{blue!55!black}{\rule{7.08mm}{1.3ex}} & \makebox[3.4em][r]{41.1\%}\,\textcolor{black!35}{\rule{8.22mm}{1.3ex}} & 0.068 \\
\quad Ridge Regression & \makebox[3.4em][r]{31.5\%}\,\textcolor{blue!55!black}{\rule{6.30mm}{1.3ex}} & \makebox[3.4em][r]{36.7\%}\,\textcolor{black!35}{\rule{7.34mm}{1.3ex}} & 0.002 \\
\quad Linear Regression & \makebox[3.4em][r]{31.1\%}\,\textcolor{blue!55!black}{\rule{6.22mm}{1.3ex}} & \makebox[3.4em][r]{36.2\%}\,\textcolor{black!35}{\rule{7.24mm}{1.3ex}} & 0.002 \\
\addlinespace[0.5ex]
\multicolumn{4}{@{}l}{\textbf{\textit{Baseline}}} \\
\quad Naive* & \makebox[3.4em][r]{19.3\%}\,\textcolor{blue!55!black}{\rule{3.86mm}{1.3ex}} & \makebox[3.4em][r]{22.5\%}\,\textcolor{black!35}{\rule{4.50mm}{1.3ex}} & 0.072 \\
\bottomrule
\end{tabular*}
\caption{Average win rate of each model on \bench{} under wQL (probabilistic) and WAPE (point) metrics, alongside inference latency. A model's win rate is the fraction of task windows on which it has a lower error than another model, averaged over all other models in the table (ties count half). The best score per column is \textbf{bolded} and the second-best is \underline{underlined}. Note that Naive* represents a naive baseline which does not consume covariates. All other models are evaluated with covariates. $^\dagger$The simplest, untuned versions of the regression models; tuned variants are in \Cref{app:tuned-regression}.}
\label{tab:leaderboard-winrate}
\vspace{-2mm}
\end{table}

%% file: inputs/appendix_task_examples_v2.tex
\section{Example tasks}
\label{app:task-examples}

We show example tasks here to make the benchmark construction concrete. As described in the main text (\Cref{sec:benchmark-construction}), the axes along which we build \bench{} are:
\begin{itemize}
  \item \textbf{Future type}: Replay or Novel.
  \item \textbf{Number of covariates} $K \in \{1,2,3,4,5\}$.
  \item \textbf{Context / prediction length} $c$: from $C_1$ (context $1024$, prediction $128$) to $C_4$ (context about $8{,}000$, prediction $512$ to $864$, depending on the simulator), with longer contexts for the longest events (\Cref{app:task-layout}).
  \item \textbf{Event length} $\ell \in \{32,64,128,256,512\}$.
  \item \textbf{Number of events} $E$: from the fewest (one or two) to the most the context permits, up to $64$.
\end{itemize}
As set out in the construction recipe (\Cref{sec:benchmark-construction}), this is repeated for each target in each simulator.

Crucially, variations along one axis are built with every other axis held fixed. For example, the Replay and Novel tasks come from the same history, and only the future differs. This lets each axis be read on its own.

Each example below shows the history and the future of the target, together with its covariates. Except for the covariate axis, every example uses a single covariate, to keep the figures easy to read. History and future are separated by a dashed forecast-start line; in-context events are shaded grey and the future event red. Only the Replay future is drawn for every axis except the Replay-versus-Novel one. Some panels also show the pre-intervention target, the dotted line marking what the target would be with no event; it is shown purely for illustration and is not part of any task, and models never receive it.

\subsection{Replay and Novel futures}
\label{app:task-examples-future}

Each figure shows a Replay task and a Novel task over the same history. The history and its covariate are identical across the two; only the future event changes. Replay reuses covariate values already seen in the history, while Novel pushes them beyond that range.

\begin{figure}[htbp]
  \centering
  \includegraphics[width=\linewidth]{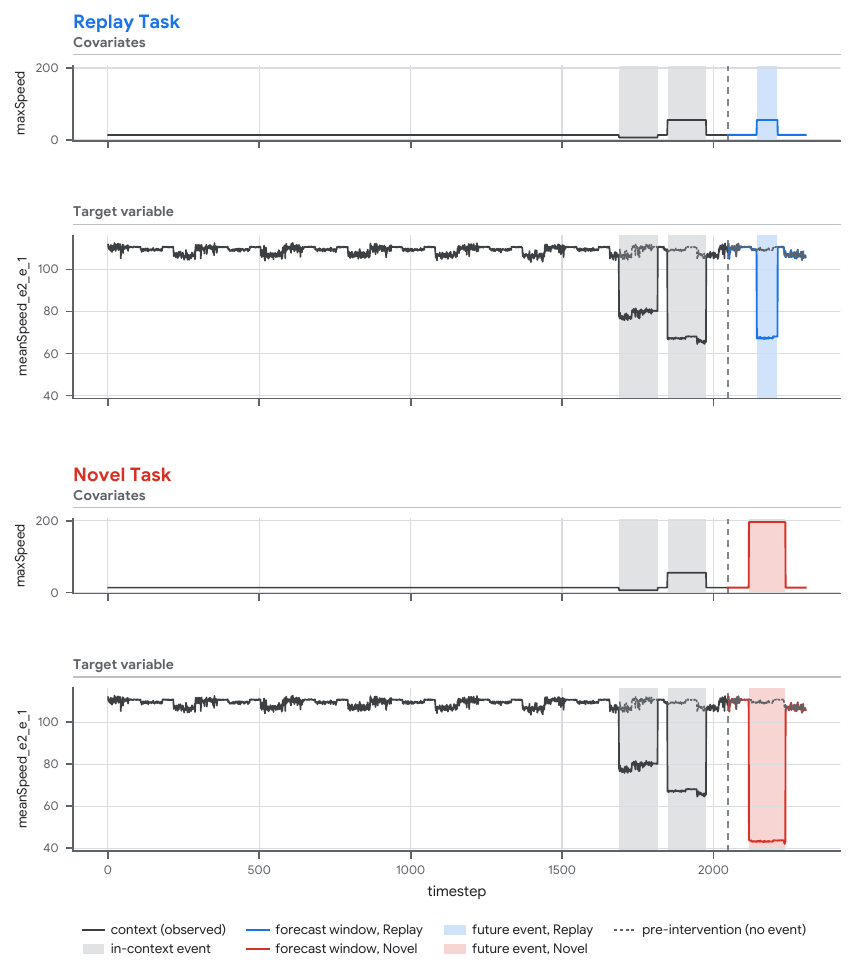}
  \caption{A Replay task and a Novel task over a shared history. The dotted pre-intervention target is shown for illustration only and is not part of the task.}
  \label{fig:ex-future-1}
\end{figure}

\begin{figure}[htbp]
  \centering
  \includegraphics[width=\linewidth]{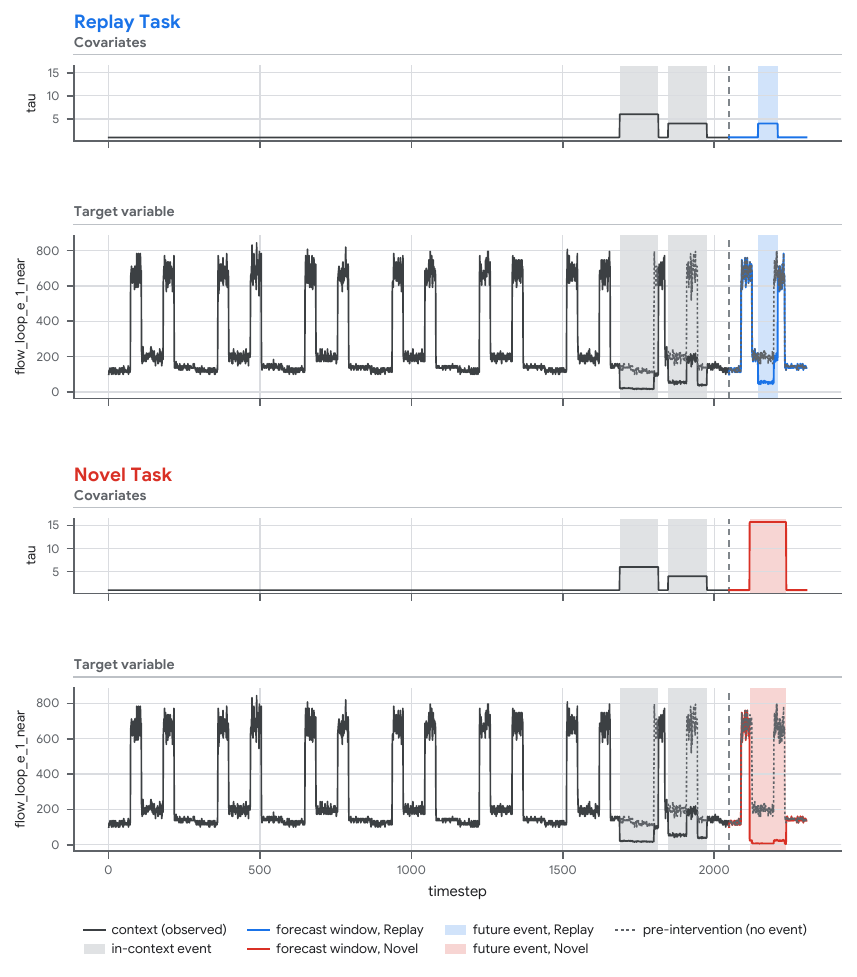}
  \caption{A Replay task and a Novel task over a shared history. The dotted pre-intervention target is shown for illustration only and is not part of the task.}
  \label{fig:ex-future-2}
\end{figure}

\begin{figure}[htbp]
  \centering
  \includegraphics[width=\linewidth]{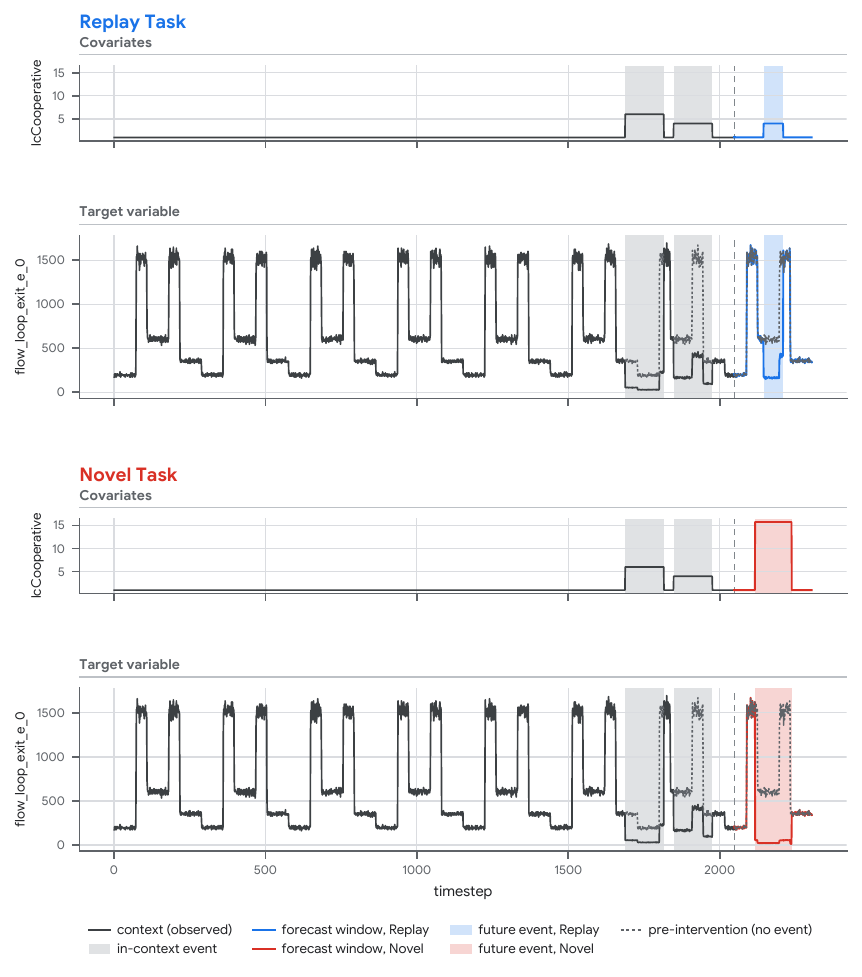}
  \caption{A Replay task and a Novel task over a shared history. The dotted pre-intervention target is shown for illustration only and is not part of the task.}
  \label{fig:ex-future-3}
\end{figure}

\FloatBarrier

\subsection{Event length}
\label{app:task-examples-eventlen}

Each figure varies the event length $\ell$ across columns, holding the target, covariate, context, and number of events fixed. Longer events hold the covariate in place for longer, so the target's response is seen over a longer window.

\begin{figure}[htbp]
  \centering
  \includegraphics[width=\linewidth]{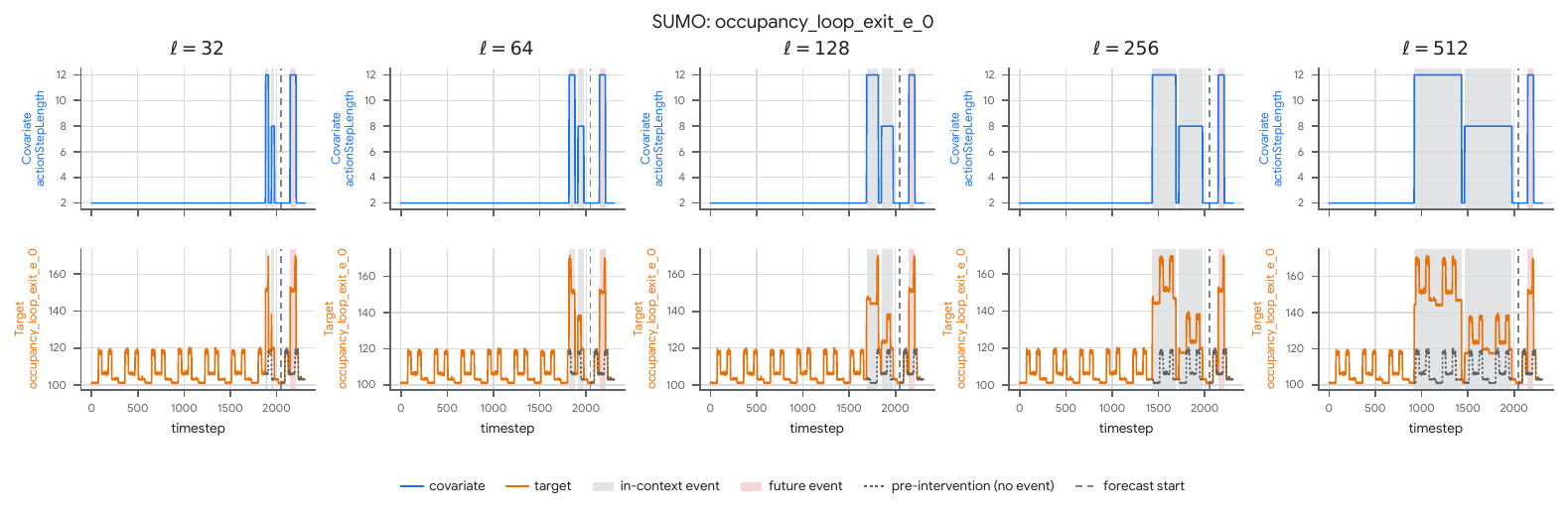}
  \caption{Tasks that differ only in event length $\ell = 32,64,128,256,512$. The dotted pre-intervention target is shown for illustration only and is not part of the task.}
  \label{fig:ex-eventlen-1}
\end{figure}

\begin{figure}[htbp]
  \centering
  \includegraphics[width=\linewidth]{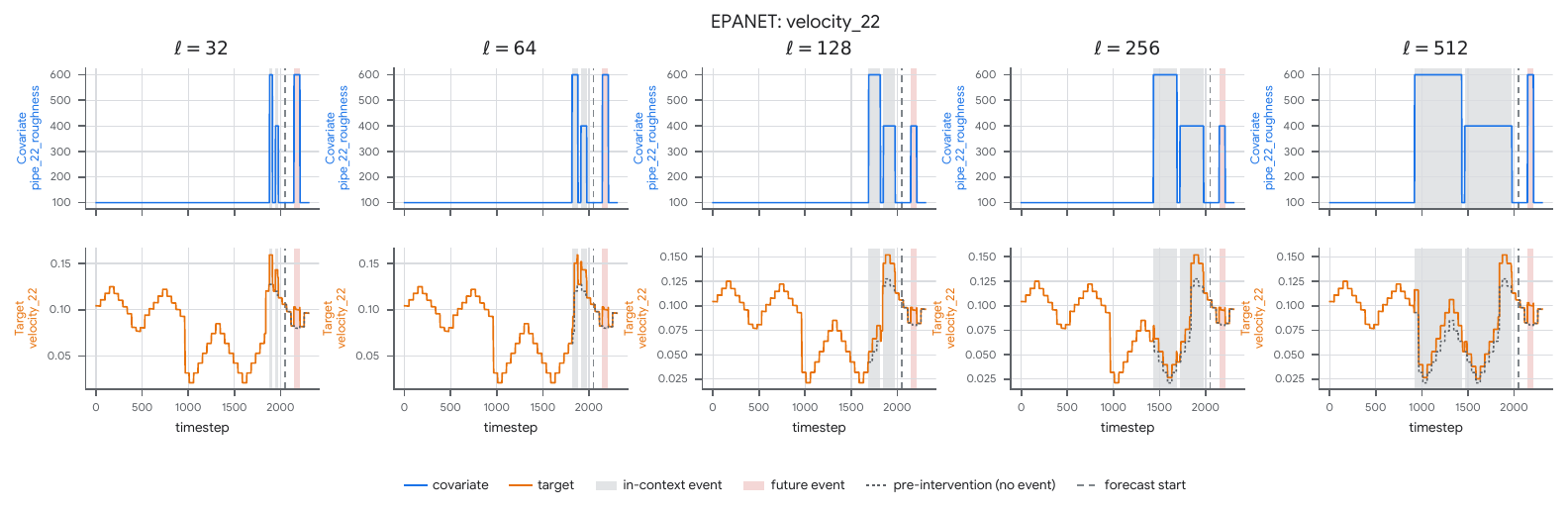}
  \caption{Tasks that differ only in event length $\ell = 32,64,128,256,512$. The dotted pre-intervention target is shown for illustration only and is not part of the task.}
  \label{fig:ex-eventlen-2}
\end{figure}

\begin{figure}[htbp]
  \centering
  \includegraphics[width=\linewidth]{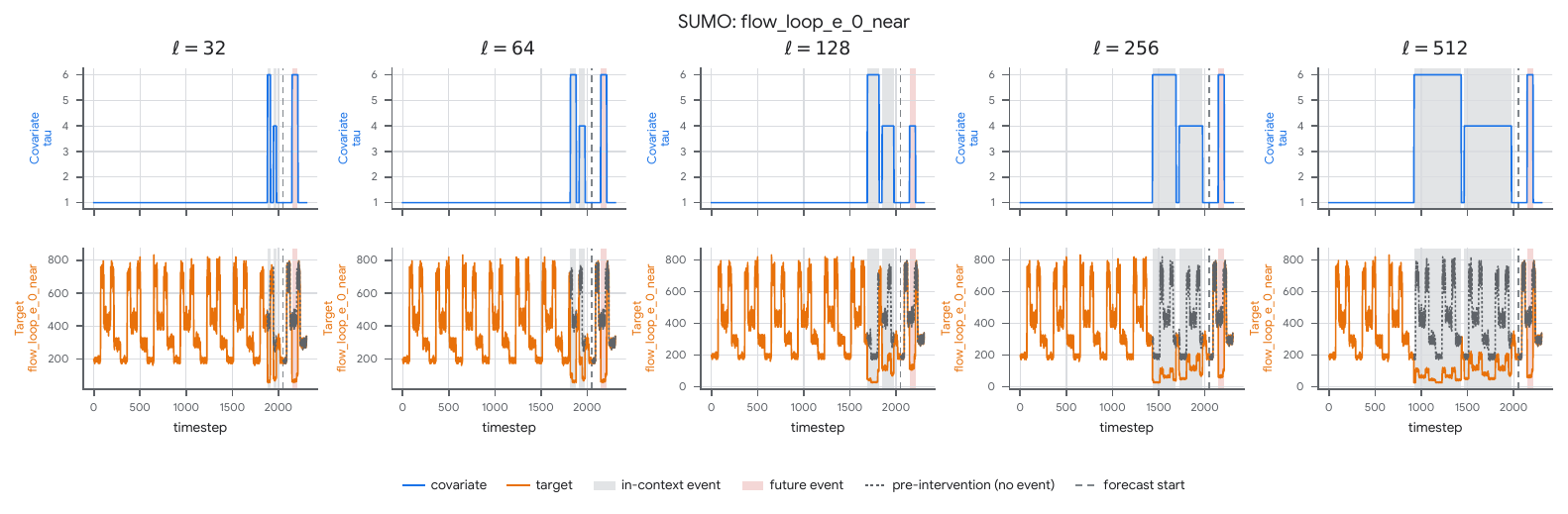}
  \caption{Tasks that differ only in event length $\ell = 32,64,128,256,512$. The dotted pre-intervention target is shown for illustration only and is not part of the task.}
  \label{fig:ex-eventlen-3}
\end{figure}

\begin{figure}[htbp]
  \centering
  \includegraphics[width=\linewidth]{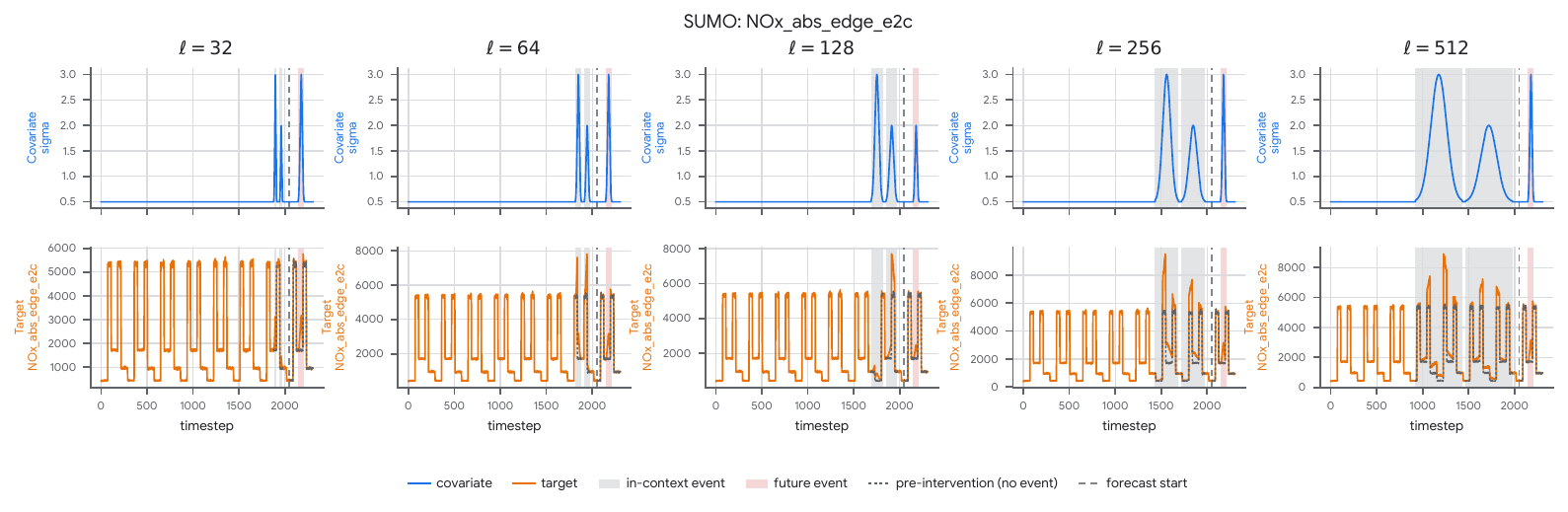}
  \caption{Tasks that differ only in event length $\ell = 32,64,128,256,512$. The dotted pre-intervention target is shown for illustration only and is not part of the task.}
  \label{fig:ex-eventlen-4}
\end{figure}

\FloatBarrier

\subsection{Number of events}
\label{app:task-examples-numevents}

Each figure varies the number of events $E$ across columns, holding every other axis fixed. More events set the covariate at more points across the history, giving more views of the covariate--target relationship.

\begin{figure}[htbp]
  \centering
  \includegraphics[width=\linewidth]{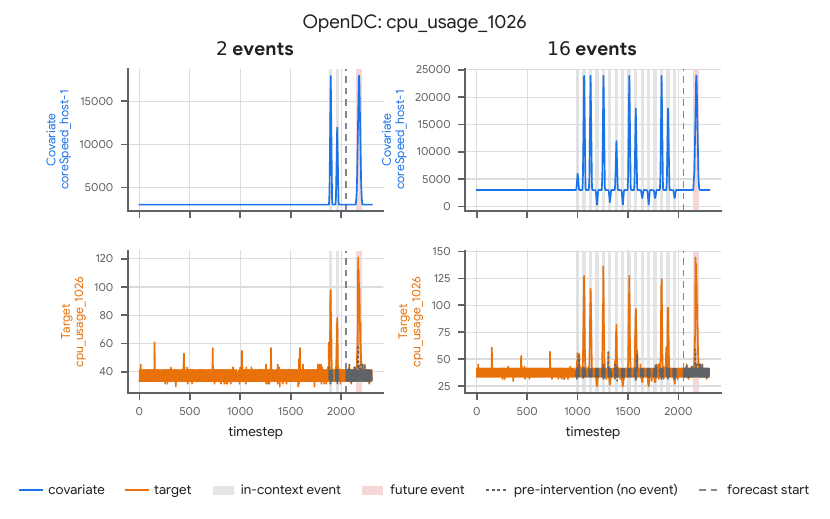}
  \caption{Tasks that differ only in the number of events (few versus many). The dotted pre-intervention target is shown for illustration only and is not part of the task.}
  \label{fig:ex-numevents-1}
\end{figure}

\begin{figure}[htbp]
  \centering
  \includegraphics[width=\linewidth]{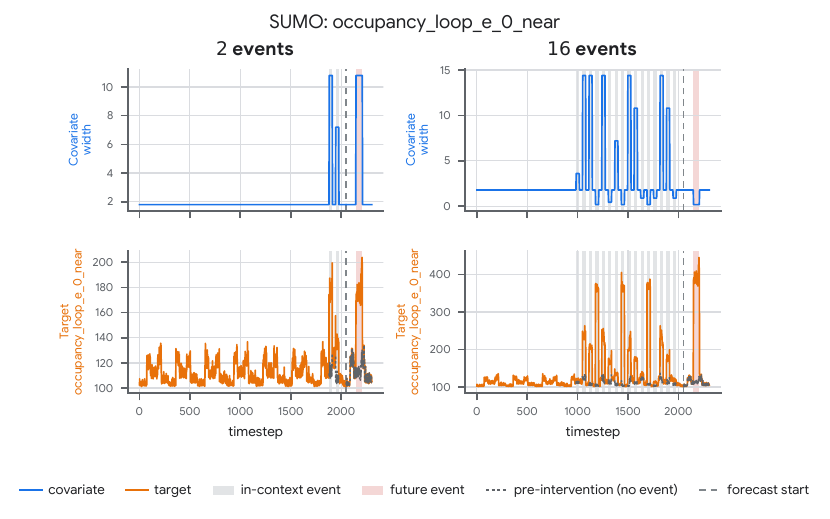}
  \caption{Tasks that differ only in the number of events (few versus many). The dotted pre-intervention target is shown for illustration only and is not part of the task.}
  \label{fig:ex-numevents-2}
\end{figure}

\begin{figure}[htbp]
  \centering
  \includegraphics[width=\linewidth]{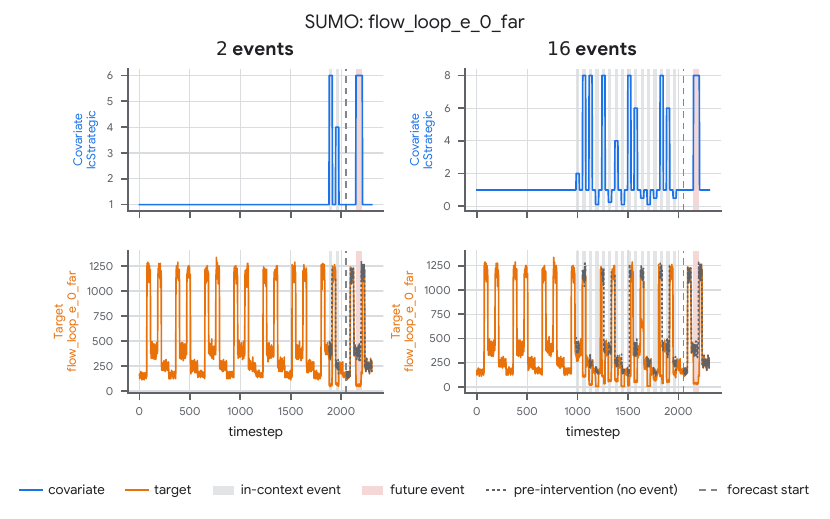}
  \caption{Tasks that differ only in the number of events (few versus many). The dotted pre-intervention target is shown for illustration only and is not part of the task.}
  \label{fig:ex-numevents-3}
\end{figure}

\FloatBarrier

\subsection{Context and prediction length}
\label{app:task-examples-context}

Each figure varies the context and prediction length across columns, holding every other axis fixed. Each column is labelled with its context length (CTX) and prediction length (PRED).

\begin{figure}[htbp]
  \centering
  \includegraphics[width=\linewidth]{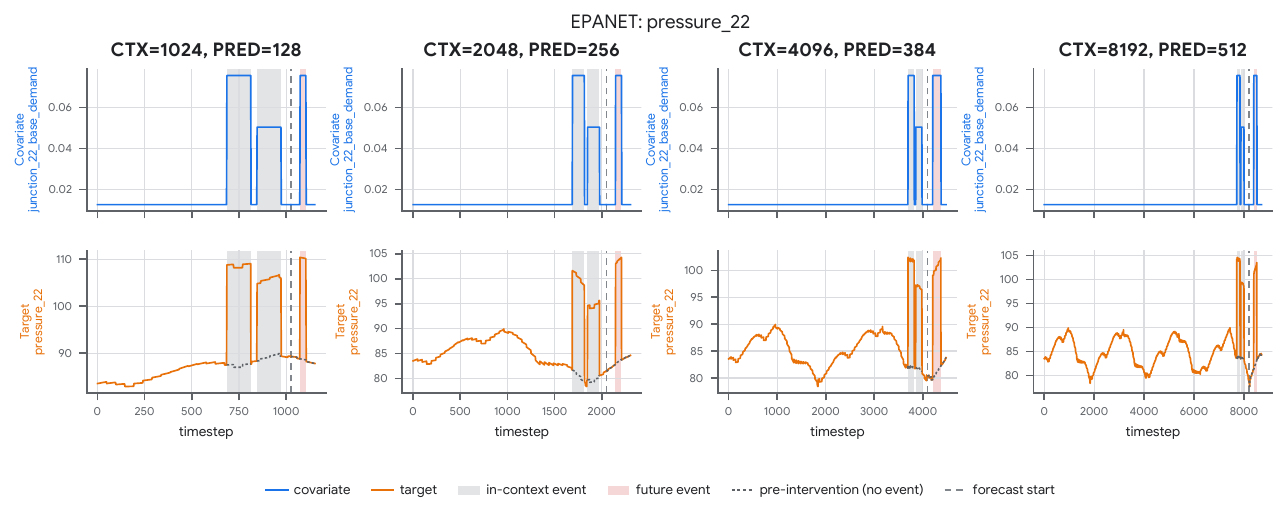}
  \caption{Tasks that differ only in context/prediction pair, $C_1$ through $C_4$. The dotted pre-intervention target is shown for illustration only and is not part of the task.}
  \label{fig:ex-context-1}
\end{figure}

\begin{figure}[htbp]
  \centering
  \includegraphics[width=\linewidth]{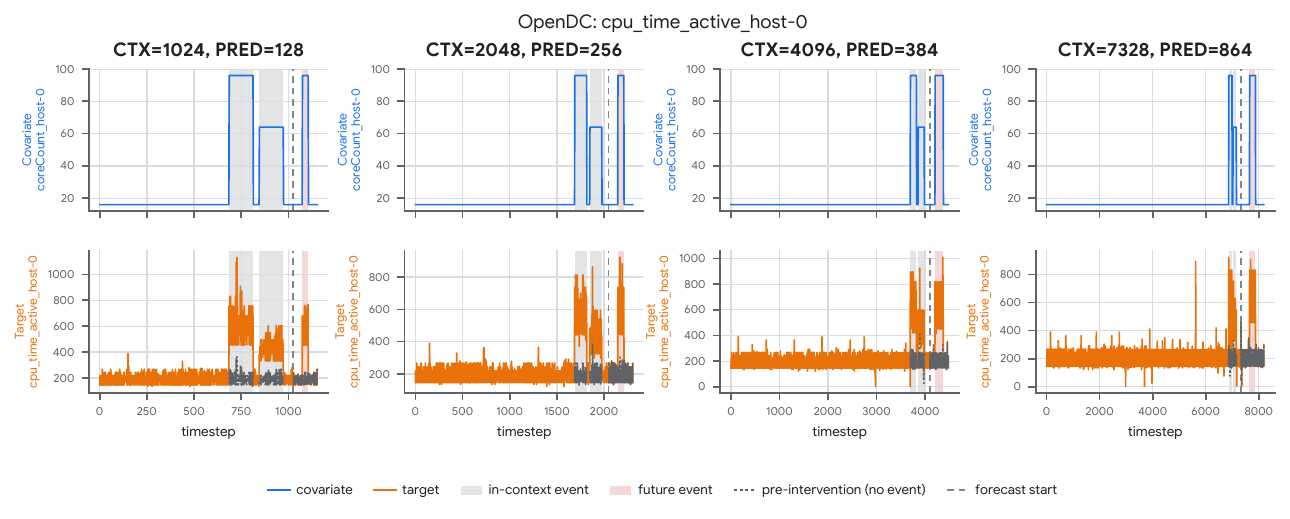}
  \caption{Tasks that differ only in context/prediction pair, $C_1$ through $C_4$. The dotted pre-intervention target is shown for illustration only and is not part of the task.}
  \label{fig:ex-context-2}
\end{figure}

\begin{figure}[htbp]
  \centering
  \includegraphics[width=\linewidth]{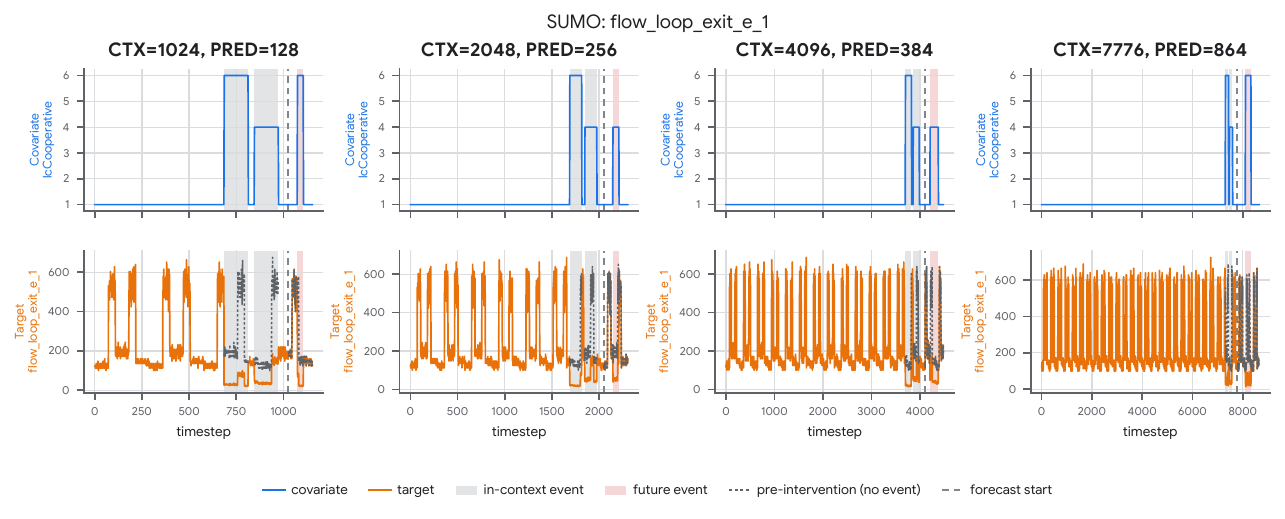}
  \caption{Tasks that differ only in context/prediction pair, $C_1$ through $C_4$. The dotted pre-intervention target is shown for illustration only and is not part of the task.}
  \label{fig:ex-context-3}
\end{figure}

\FloatBarrier

\subsection{Number of covariates}
\label{app:task-examples-covariates}

Each figure varies the number of covariates $K$ across columns, holding every other axis fixed. As $K$ grows, each event acts on more covariates at once, and the model must learn more relationships from the history. Here every covariate is plotted above the shared target.

\begin{figure}[htbp]
  \centering
  \includegraphics[width=\linewidth]{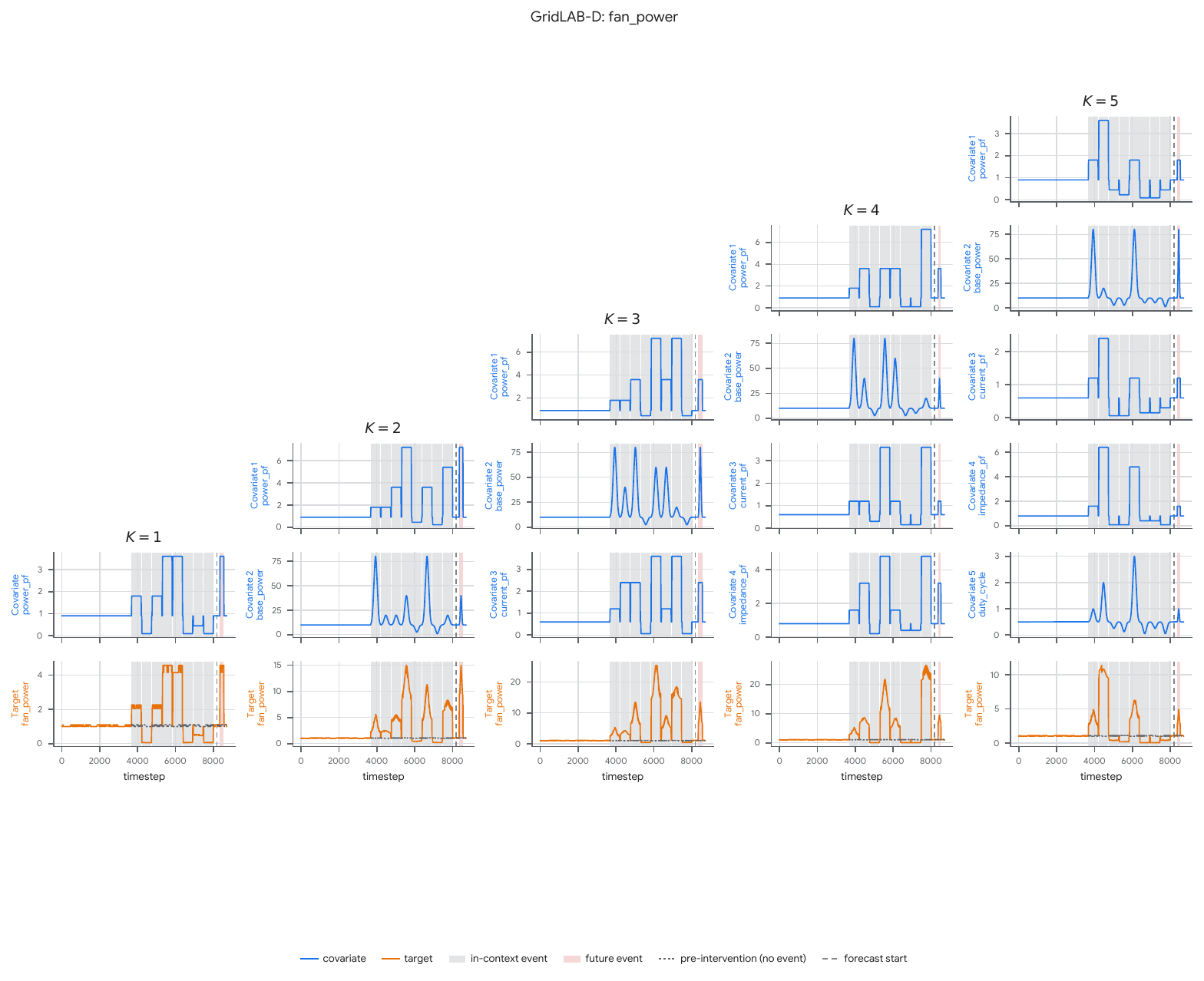}
  \caption{Tasks that differ only in the number of covariates, $K = 1$ through $5$. The dotted pre-intervention target is shown for illustration only and is not part of the task.}
  \label{fig:ex-covariates-1}
\end{figure}

\begin{figure}[htbp]
  \centering
  \includegraphics[width=\linewidth]{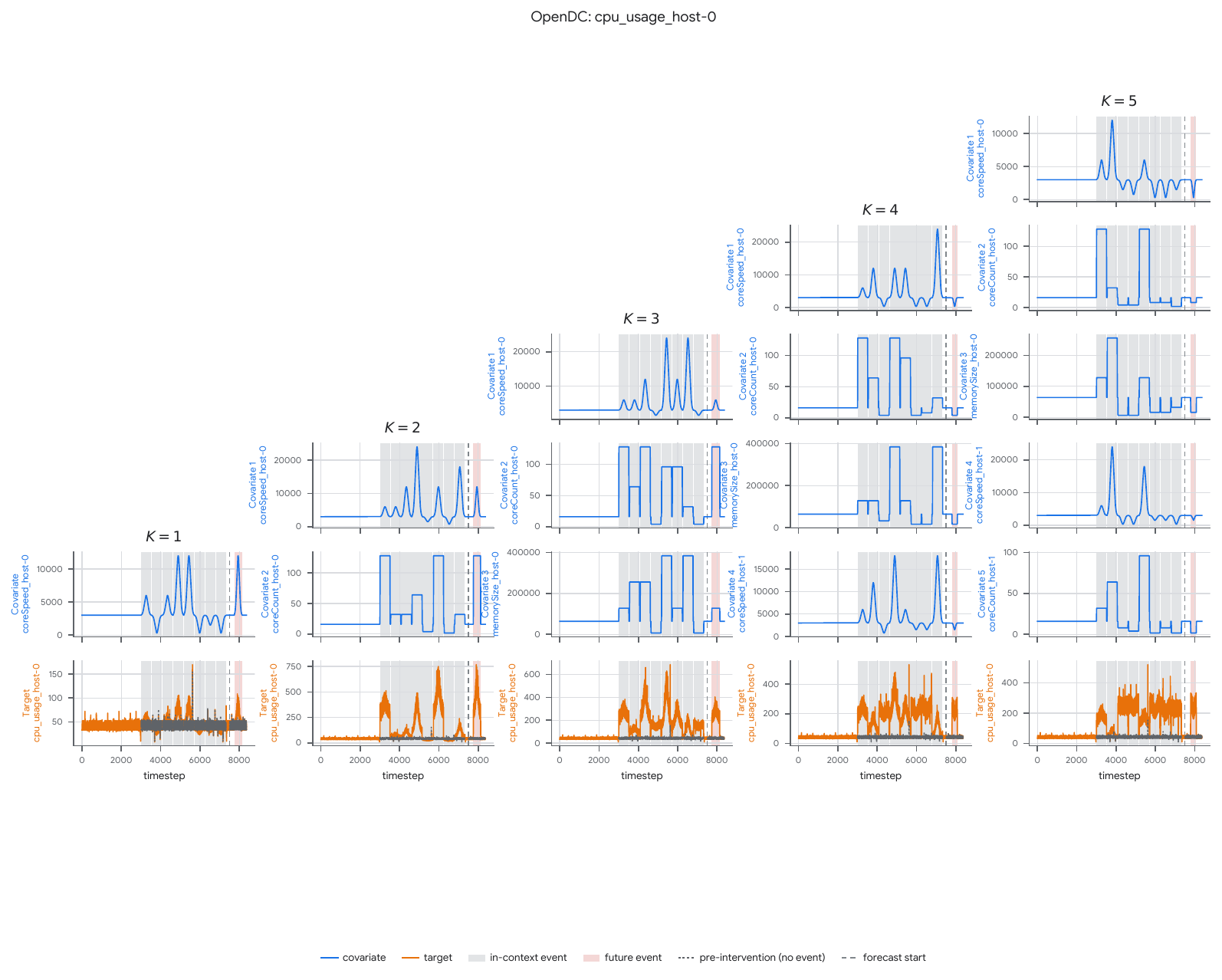}
  \caption{Tasks that differ only in the number of covariates, $K = 1$ through $5$. The dotted pre-intervention target is shown for illustration only and is not part of the task.}
  \label{fig:ex-covariates-2}
\end{figure}

\FloatBarrier